\documentclass[sigconf]{acmart}

\usepackage{amsmath}
\usepackage{bm}
\usepackage[mathscr]{eucal}
\usepackage{enumitem}
\usepackage[]{graphicx}
\usepackage{stfloats}

\usepackage{amssymb}

\usepackage{soul}
\soulregister\cite7 
\soulregister\ref7

\definecolor{visual_color}{HTML}{F668A7}
\definecolor{verbal_color}{HTML}{0EA4DC}
\definecolor{diagram_color}{HTML}{11B697}

\DeclareMathOperator*{\argmax}{argmax}
\DeclareMathOperator*{\argmin}{argmin}

\AtBeginDocument{%
  \providecommand\BibTeX{{%
    \normalfont B\kern-0.5em{\scshape i\kern-0.25em b}\kern-0.8em\TeX}
    }
}

\copyrightyear{2025} 
\acmYear{2025} 
\setcopyright{cc}
\setcctype{by}
\acmConference[CHI '25]{CHI Conference on Human Factors in Computing Systems}{April 26--May 01, 2025}{Yokohama, Japan}
\acmBooktitle{CHI Conference on Human Factors in Computing Systems (CHI '25), April 26--May 01, 2025, Yokohama, Japan}\acmDOI{10.1145/3706598.3714058}
\acmISBN{979-8-4007-1394-1/25/04}

\acmSubmissionID{4230}

\newcommand{\hlc}[2][yellow]{{\sethlcolor{#1}\hl{#2}}}
\definecolor{revision_color}{HTML}{FFFFFF}
\definecolor{todo_color}{HTML}{FF0000}
\newcommand{\rev}[1]{\hlc[revision_color]{#1}} 

\begin{document}



\title{Diagrammatization and Abduction to Improve AI Interpretability\\With Domain-Aligned Explanations for Medical Diagnosis}


\author{Brian Y. Lim}
\affiliation{%
  \department{Department of Computer Science}
  \institution{National University of Singapore}
  \city{Singapore}
  \country{Singapore}
}
\email{brianlim@comp.nus.edu.sg}
\authornote{Corresponding author}

\author{Joseph P. Cahaly}
\affiliation{%
  \institution{Massachusetts Institute of Technology}
  \city{Cambridge}
  \state{Massachusetts}
  \country{USA}
}
\email{jcahaly@mit.edu}

\author{Chester Y. F. Sng}
\affiliation{%
  \institution{National University of Singapore}
  \city{Singapore}
  \country{Singapore}
}
\email{chestersng@u.nus.edu}

\author{Adam Chew}
\affiliation{%
  \institution{National University of Singapore}
  \city{Singapore}
  \country{Singapore}
}
\email{yschew@u.nus.edu}

\renewcommand{\shortauthors}{Brian Y. Lim, et al.}

\begin{abstract} 
Many visualizations have been developed for explainable AI (XAI), but they often require further reasoning by users to interpret. 
Investigating XAI for high-stakes medical diagnosis, we propose improving domain alignment with diagrammatic and abductive reasoning to reduce the interpretability gap. 
We developed DiagramNet to predict cardiac diagnoses from heart auscultation, select the best-fitting hypothesis based on criteria evaluation, and explain with clinically-relevant murmur diagrams.
The ante-hoc interpretable model leverages domain-relevant ontology, representation, and reasoning process to increase trust in expert users.
In modeling studies, we found that DiagramNet not only provides faithful murmur shape explanations, but also has better performance than baseline models. 
We demonstrate the interpretability and trustworthiness of diagrammatic, abductive explanations in a qualitative user study with medical students, showing that clinically-relevant, diagrammatic explanations are preferred over technical saliency map explanations. 
This work contributes insights into providing domain-aligned explanations for user-centric XAI in complex domains.
\end{abstract}



\maketitle

\section{Introduction}

The need for AI accountability has spurred explainable AI (XAI) development~\cite{abdul2018trends, adadi2018peeking, arrieta2020explainable, guidotti2018survey, hohman2018visual}.
However, current approaches tend to use elementary, off-the-shelf visualizations, such as bar or line charts and heat maps, that assume users 
know how to properly interpret the explanations in context of the instances and decisions, and evaluate their plausibility.
Consequently, these are difficult to make sense of~\cite{kaur2022sensible}, too simplistic to provide effective feedback~\cite{poursabzi2021manipulating}, and require significant subsequent effort to interpret~\cite{dhanorkar2021needs}.

We argue that XAI should be \textbf{aligned} to domain \textit{knowledge}, \textit{representations}, and \textit{reasoning}~\cite{brachman2004knowledge} to be more accessible and interpretable to users for complex domains. 
It is clear that explanations need to be in terms of interpretable features~\cite{lim2011design, zytek2022need} or concepts~\cite{kim2018interpretability, koh2020concept}, but little has been done to ensure these terms are presented in domain-aligned representations that support established structures and reasoning methods to verify explanations and decisions.
One way to scaffold knowledge representations and reasoning explicitly is with diagrams, where users can perform diagrammatic reasoning or \textbf{diagrammatization} to interpret and evaluate explanations~\cite{peirce1976new, hoffmann2010diagrams,magnani2009abductive}.
Diagrams are used in many domains to explain complex phenomena. 
In physics, free body diagrams can explain how objects move due to forces, and new forces can be added to diagrams to better fit a complex observation.
In medicine, diagrams can be used to annotate physiological mechanisms of diseases on medical images, with different annotations indicating different diagnoses.
Diagrams are distinct from visualization since they are domain constrained~\cite{shimojima1999graphic}, provide a systematic approach to read and manipulate them~\cite{peirce1976new}, and are propositional to support inference making~\cite{larkin1987diagram}, thus enhancing interpretation.

Furthermore, when considering an explanation, we often judge its acceptability by comparing against multiple competing ones. 
Indeed, to make sense of observations, people employ abductive reasoning or \textbf{abduction}---inference to the best explanation---to generate hypotheses and note their effects, then evaluate them in context~\cite{popper2014conjectures}.
Yet, XAI techniques mostly provide single explanations based the model's deductive reasoning. Indubitably, contrastive explanations do provide explanations for why alternative outcomes were not chosen~\cite{lim2009and, miller2019explanation}, but these do not use domain-grounded hypotheses or evaluation criteria.
Along with Hoffman et al.~\cite{hoffman2017explaining, hoffman2020explaining, hoffman2022psychology}, Wang et al.~\cite{wang2019designing}, Medianovskyi et al.~\cite{medianovskyi2022explainable}, and Miller~\cite{miller2023explainable}, we argue that XAI should support abductive reasoning.

Finally, an additional requirement for XAI domain alignment is to align AI reasoning with human reasoning for \textbf{ante-hoc interpretability}~\cite{wang2019designing, zhang2022towards, matsuyama2023iris}, rather than rely on post-hoc model-agnostic explanations.
This is critical for high-stakes decisions to verify the actual reasoning process of the "glass box" AI model, rather than a proxy explanation of a "black box" model which may surreptitiously be using another reasoning method~\cite{rudin2019stop}.
Accordingly, this transparency can support users to trust and delegate their decisions since both the user and AI share the same reasoning.

\begin{figure*}[!t]
    \centering
    \includegraphics[width=13.8cm]{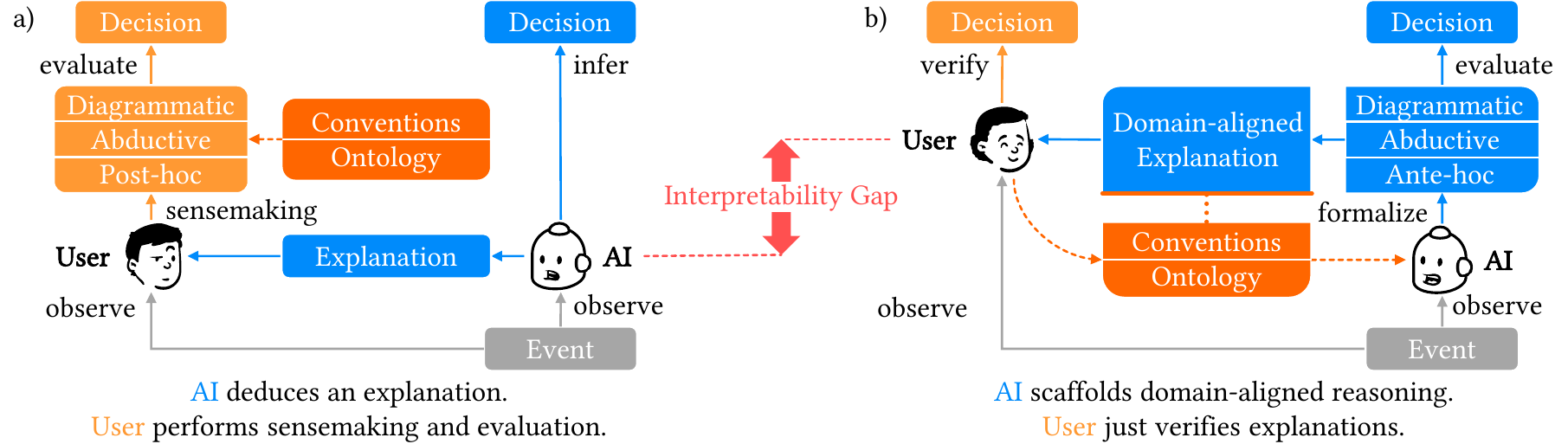}
    \vspace{-0.20cm}
    \caption{
    Reasoning processes between the user and AI.
    a) Current XAI: the user views an explanation based on deductive reasoning, and post-hoc has to make sense of it in context of the domain and evaluate its plausibility.
    This leaves an interpretability gap.
    b) Domain-aligned XAI: the ante-hoc interpretable AI encodes domain ontology and conventions to provide explanations diagrammatically to convey domain concepts and relational structures, and abductively to convey how it evaluates multiple hypotheses.
    The user simply verifies the AI explanation and prediction.
    }
    \Description{
    The image compares two reasoning processes between the user and AI. On the left, a linear flowchart shows the current state of explainable AI explanations, where the user generates and evaluates hypotheses, leading to an interpretability gap. This section includes icons representing data, user, and AI, connected by arrows indicating the flow of reasoning. Dotted arrows highlight the inference steps, and a dashed pink line represents the interpretability gap.
    On the right, an enhanced flowchart illustrates the proposed diagrammatization approach, where AI scaffolds abductive reasoning by enumerating, evaluating, and resolving hypotheses, conforming to domain conventions, and representing explanations visually or verbally. This section includes similar icons for data, user, and AI, with additional icons for diagrammatic explanations and domain conventions. Solid arrows indicate direct connections between steps, eliminating the interpretability gap. The process allows the user to validate the AI's explanation and prediction directly.
    In the left flowchart, the blocks are connected as follows:
    1. Data to User: An arrow from data to user indicates the user observes the data.
    2. User to Hypotheses: An arrow from user to hypotheses shows the user generates hypotheses.
    3. Hypotheses to Evaluation: An arrow from hypotheses to evaluation indicates the user evaluates the hypotheses.
    4. Evaluation to Explanation: An arrow from evaluation to explanation shows the user interprets the explanation.
    5. Explanation to Classification: An arrow from explanation to classification indicates the user reaches a classification or resolution.
    In the right flowchart, the blocks are connected as follows:
    1. Data to AI: An arrow from data to AI indicates the AI observes the data.
    2. AI to Diagrammatic Explanation: An arrow from AI to diagrammatic explanation shows the AI provides a diagrammatic explanation.
    3. Diagrammatic Explanation to Hypotheses: An arrow from diagrammatic explanation to hypotheses indicates the AI enumerates hypotheses.
    4. Hypotheses to Evaluation: An arrow from hypotheses to evaluation shows the AI evaluates the hypotheses.
    5. Evaluation to Domain Conventions: An arrow from evaluation to domain conventions indicates the AI conforms to domain conventions.
    6. Domain Conventions to Visual/Verbal Representation: An arrow from domain conventions to visual/verbal representation shows the AI represents explanations visually or verbally.
    7. Visual/Verbal Representation to User: An arrow from visual/verbal representation to user indicates the user validates the AI's explanation and prediction.
    8. User to Classification: An arrow from user to classification shows the user reaches a classification or resolution.
    The right flowchart also includes labels for three capabilities of diagrammatization:
    - i) Hypotheses is located between the diagrammatic explanation and hypotheses blocks.
    - ii) Conventions is located between the evaluation and domain conventions blocks.
    - iii) Visual/Verbal Representations is located between the domain conventions and visual/verbal representation blocks.
    }
    \label{fig:concept-interpretability-gap}
    \vspace{-0.20cm}
\end{figure*}

We studied a medical AI application, cardiac diagnosis on heart auscultation, that requires
i) \rev{annotative} diagrammatization, ii) selective abduction, and iii) ante-hoc interpretability.
Clinicians explain heart auscultation using murmur diagrams that encode representations of abnormal murmur sounds based on audio amplitude shapes and positions~\cite{judge2015heart} (see Figs. \ref{fig:concept-heart-murmur-diagrams}, \ref{fig:demo-explanations}, \ref{fig:demo-mvp}, \ref{fig:demo-as}).
When diagnosing, clinicians abductively infer the most likely explanatory diagnosis for the observed symptoms~\cite{brachman2004knowledge_ch13, patel2005thinking}.
Since medicine is high-stakes, the AI model needs to be intrinsically interpretable to ensure that clinicians have a faithful understanding of its decisions from generated explanations~\cite{rudin2019stop}.
Therefore, to support domain-aligned XAI, we propose DiagramNet, a modular ante-hoc interpretable model, to use \textit{diagrammatization} to present explanations as diagrams to scaffold diagrammatic reasoning, and \textit{abduction} to help users to assess how the model judged among hypotheses and selected the best explanation.
Our \textbf{contributions} are:
\begin{enumerate}
    \item[1)] \textit{Domain-aligned XAI} design framework for ante-hoc interpretability with diagrammatic and abductive reasoning.
    This can be implemented with various methods.
    \item[2)] \textit{DiagramNet} model to formalize diagrammatic constraints and provide diagram-based, abductive explanations by enumerating, evaluating, and resolving hypotheses to infer to the best explanation as the prediction label.
    \item[3)] \textit{Diagrammatic} explanations of cardiac diagnoses with clinically-relevant murmur diagrams. We formalized murmur shapes to predict them as explanations in DiagramNet.
    \item[4)] Evaluations with a real-world heart auscultation dataset~\cite{yaseen2018classification}.
    \begin{enumerate}
        \item[a)] \textit{Demonstration study} illustrating that diagrammatization can provide abductive explanations which follows domain conventions, and supplemental contrastive, counterfactual, and example-based explanations.
        \item[b)] \textit{Modeling study} showing that DiagramNet improves both prediction performance and explanation faithfulness compared to baseline and alternative models, and
        \item[c)] \textit{Qualitative user study} with medical domain experts finding that diagram-based explanations are more clinically sound, useful, and convincing than saliency map explanations.
    \end{enumerate}
    \item[5)] Discussion of the scope, generalization and contextualization of domain-aligned XAI.
\end{enumerate}

\section{Diagrammatic and abductive reasoning}

Current XAI use elementary visualizations that require users to do extensive sensemaking~\cite{kaur2020interpreting}. 
Fig. \ref{fig:concept-interpretability-gap} illustrates how domain-aligned explanations can close this interpretability gap.
We describe the human reasoning processes of diagrammatic and abductive reasoning to introduce how supporting them can improve XAI interpretability.

\subsection{Diagrammatization for domain-alignment} \label{sec:diagrammatization-design-space}

\begin{table*}[t]
\small
\centering
\vspace{-0.00cm}
\caption{
    Diagrammatization design space with dimensions to compare verbal and visual representations of XAI.
}
\vspace{-0.25cm}
\label{table:verbal-visual-diagram}
\begin{tabular}{lllllll}
\hline
 &
   &
   &
  \multicolumn{1}{l}{Level of states} &
  \multicolumn{1}{l}{Homomorphism} &
  \multicolumn{1}{l}{Expressivity} &
  \multicolumn{1}{l}{Inherent constraints} \\ 
\hline
				   
\addlinespace[0.05cm] 
\multicolumn{2}{l}{\color{visual_color}Visualization} &  &             &                       &             &               \\
                   & Attribution                   &  & Continuous  & Low (by data type)    & Some bounds (variables, importance) & None          \\
                   & Concept-based &  & Continuous  & Low (semantic)                & Bounded (latent variables)       & Taxonomical   \\
                   & Model-based   &  & Continuous  & Low (associative)              & Bounded (variables, relations)       & Topological   \\
                   & Schematic     &  & Continuous  & Medium (associative)             & Bounded (variables, relations)   & Topological   \\
                   & Annotative*    &  & Continuous  & High (associative, physical)  & Bounded (variables, relations, values)       & Topological, geometrical   \\
  
\addlinespace[0.02cm] 
\multicolumn{2}{l}{\color{verbal_color}Verbalization} &  &             &                       &             &               \\
                   & NL Generative                    &  & Categorical & Low (linguistic) & Unbounded   & None          \\
                   & Template-based                   &  & Categorical & Low (descriptive)     & Bounded (variables, relations)     & Taxonomical   \\
                   & Symbolic                         &  & Categorical & Low (mathematical)    & Bounded (variables, relations)     & Logical \\
				                      
\addlinespace[0.05cm]
\hline
\end{tabular}
\vspace{-0.25cm}
\end{table*}

\rev{Philosopher Charles S. Peirce defined diagrams as an encompassing reasoning framework for visual, symbolic, and verbal representations comprising: 
an \textbf{ontology} that defines the entities and their relations, 
\textbf{conventions} that prescribe how to \textit{interpret} diagrams and 
\textit{manipulate} diagrams to evaluate alternative explanations~\cite{peirce1976new}.}
As a literature review, we frame visual and verbal XAI methods in a diagrammatization design space to articulate their varying constraints that limit or facilitate reasoning in complex domains.

\subsubsection{Visualization}
Most XAI techniques leverage visualization to augment human cognition and understanding~\cite{card1999readings}.
\rev{While data visualization focuses on providing representations of data~\cite{munzner2014visualization}, visual diagrams are propositional to convey knowledge and relationships for inference making~\cite{larkin1987diagram, peirce1976new}.
We discuss various visual explanation representations with increasing domain constraints.}
\begin{enumerate}
    \item[a)] \textit{\color{visual_color}Attribution} 
    explanations use elementary visualizations to describe relationships between variables.
    Techniques include:
    \textit{Bar charts} 
    to show feature attributions~\cite{ribeiro2016should, kulesza2009fixing,lim2011design}, distribution point clouds~\cite{lundberg2017unified}, and violin plots~\cite{wang2021show}.
    \textit{Line graphs} 
    to show nonlinear relationships using partial dependence plots~\cite{krause2016interacting},
    modeled with generalized additive models~\cite{caruana2015intelligible,abdul2020cogam}, etc.
    \textit{Saliency maps}
    to show important regions as heatmaps on images~\cite{selvaraju2017grad,bach2015pixel,zhou2016learning} or highlights on text~\cite{wang2022interpretable}.
    \item[b)] \textit{\color{visual_color}Concept-based}
    explanations are descriptive with semantically defined concept vectors~\cite{kim2018interpretability, koh2020concept}, or relatable cues~\cite{zhang2022towards}.
    Such explanations could be verbal or visual.
    Listing concepts by names and indicating their importance is mostly verbal since their relation to the prediction is implicit.
    Interactive editing to see generated counterfactual outcomes is diagrammatic since it allows experimentation~\cite{cai2019human,kahng2018gan,zhang2021method}.
    \item[c)] \textit{\color{visual_color}Model-based}
    explanations visualize the abstract data structure of the prediction or explanation model typically as graph networks or rule structures.
    Techniques include:
    \textit{Neural network} activations~\cite{kahng2017cti}, canonical filters in CNNs~\cite{olah2017feature}, distilled networks~\cite{bau2017network,hohman2019s}, node-link diagrams of probabilistic or causal relations~\cite{koller2009probabilistic, pearl2009causality}, and decision trees~\cite{lim2009and, lim2019does, wu2018beyond}.
    These explanations are typically used by model developers with knowledge of data structures and machine learning.
    \item[d)] \textit{\color{visual_color}Schematic} 
    explanations use \rev{abstract} diagrams specific to each domain. Domain experts use them in education and practice, yet they are underutilized in XAI, perhaps, due to the focus on model debugging or simple application domains. 
    These include free-body, optics ray, and circuit diagrams in physics, \rev{and signaling pathway, food web, and protein interaction network diagrams in biology}.
    While they could be categorized as model-based, schematic diagrams are specifically constrained toward domain ontology and conventions.
    \rev{In this work, we define these constraints with domain-specific parameters that elementary visualizations do not impose.}
    \item[e)] \textit{\color{visual_color}Annotative}
    \rev{explanations extend domain-specific schematic diagrams to explicitly fit instances. 
    In mechanical engineering, this includes computational fluid dynamics and finite element analysis visualizations.}
    In medicine, this includes annotating on images (X-ray, ultrasonogram, dermoscopy, etc.) or on time series signals (electrocardiograms, phonocardiograms, etc.).
    \rev{They can be used to simulate outcomes based on hypotheses, and be physically mapped and overlaid onto a base representation of the instance as user-verifiable annotations. 
    In this work, we focus on annotative explanations to overlay murmur diagrams on temporal phonocardiograms.}
\end{enumerate}

\subsubsection{Verbalization}
Explanations can also be written (or spoken) verbally with logical syntax or natural language. 
\begin{enumerate}
    \item[a)] \textit{\color{verbal_color}Natural Language Generative (NLG)} 
    explanations emulate people through supervised learning of human rationalizations~\cite{ehsan2018rationalization, rajani2019explain, rosenthal2016verbalization}.
    Ehsan et al. define \textit{rationalization} as justifying a model's decision \textit{``based on how a human would think''}, but this does \textit{``not necessarily reveal the true decision making process''}~\cite{ehsan2019automated}.
    Instead, ante-hoc diagrammatization aligns the explanation to the AI's true reasoning. 
    \item[b)] \textit{\color{verbal_color}Template-based}
    explanations map symbolic expressions into text with fixed terms and sentence structures~\cite{abujabal2017quint}.
    \item[c)] \textit{\color{verbal_color}Symbolic} 
    explanations use mathematical notation to describe logical relationships~\cite{chandrasekaran2005makes}.
    \textit{Rules} are popular forms~\cite{letham2015interpretable,lakkaraju2016interpretable} and
    can complement counterfactual explanations~\cite{ribeiro2018anchors,wachter2017counterfactual}.
\end{enumerate}

\subsubsection{Diagrammatization design space}
Adapting dimensions for graphic-linguistic distinction~\cite{shimojima1999graphic}, we define a diagrammatization design space (Table \ref{table:verbal-visual-diagram}) 
for specifying how XAI representations affect interpreting domain-specific explanations.
\begin{enumerate}
    \item[A.] \textit{Level of states}
    describes whether the representions can be "analog" (categorical) or "digital" (continuous). 
    {\color{verbal_color}Verbalizations} are categorical,
    while {\color{visual_color}visualizations} can also be continuous.
    \item[D.] \textit{Homomorphism} 
    refers to how analogous the diagram is to the represented domain,
    conveying the alignment of \textit{relational structures}~\cite{keil2006explanation} between the explanation and observation.
    {\color{verbal_color}Verbal} representations have low homomophism, needing users to translate text to symbols and structures.
    {\color{visual_color}Attribution} visualizations may have formats irrelevant to the domain (e.g., spectrogram of heart sounds).
    {\color{visual_color}Model-based} and {\color{visual_color}schematic} visualizations associate abstract relations with low homomorphism.
    {\color{visual_color}Annotative} diagrams overlay on physical domain representations, so have high homomorphism.
    \item[E.] \textit{Expressivity} 
    refers to whether the representation constrains information expressiveness.
    {\color{verbal_color}Generative text} verbalization is unbounded, since any text could be predicted. 
    {\color{visual_color}Attribution} visualizations show values of variables independently.
    Other representations are bounded by the graphical, symbolic, or template formats.
    High expressivity is useful to show nuances for experts, but is overwhelming to non-experts.
    \item[F.] \textit{Inherent constraints.} 
    {\color{verbal_color}Generative text} can include any words, so have no inherent constraints, while {\color{verbal_color}template-based text} is bounded to the template taxonomy. 
    {\color{visual_color}Concept-based} visualizations are also constrained by the taxonomy of concepts.
    {\color{visual_color}Model-based} and {\color{visual_color}schematic} visualizations are constrained by topological structure (e.g., decision tree).
    {\color{visual_color}Annotative} diagrams are constrained topologically or geometrically.
\end{enumerate}
Thus, \textit{diagrammatization} is the reasoning process to represent domain ontology of concepts in relational structures, and to apply domain conventions to interpret and manipulate them to evaluate explanatory knowledge.
\textit{Diagrams} need to conform to these constraints, which we support with \rev{domain-specific} parametric functions (described later in Section \ref{subsection:formalization-murmur-shapes}).
\rev{Note that more expressive, less homomorphic, and less domain-constrained diagrams are prone to spurious explanations.}
\rev{Furthermore, some advanced visualization methods are parametric (e.g., edge bundling~\cite{holten2009force, hurter2012graph, lyu2019od}, dimensionality reduction~\cite{van2008visualizing, mcinnes2018umap}), but the parameters are for visual clarity rather than domain-based hypothesis reasoning.}

\rev{Although many visual and verbal explanations are Peircean diagrammatic, some are not and leave an interpretability gap.
Diagrammatic explanations need to be \textit{iconic} to represent concepts, \textit{relational} to associate the concepts, and \textit{inferential} to support reasoning on the concepts along the relations.
Explanations that only describe input feature values or prediction confidence merely provide declarative facts. Without relating these facts to the model outcome, they do not explicitly support diagrammatic reasoning. 
Similarly, \textit{example-based} explanations retrieve examples that are similar~\cite{koh2017understanding}, contrastive~\cite{cai2019effects}, counterfactual~\cite{wachter2017counterfactual} or adversarial~\cite{kim2016examples,woods2019adversarial}. However, without explicitly relating how the differences or similarities affect the prediction, they are not diagrammatic; users have to make sense of the comparison with the current instance themselves.
Nevertheless, attributing the relations will make these explanations more diagrammatic.
\textit{Scatter plot} visualizations of multivariate relationships~\cite{cavallo2018visual} and clusters~\cite{ahn2019fairsight} are also not diagrammatic explanations, since they do not relate why distance or grouping implies the prediction; though explicitly linking these would make them diagrammatic explanations.}

\subsection{Abductive reasoning with hypotheses} \label{subsection:abductive-reasoning}

\begin{figure*}[t!]
    \centering
    \includegraphics[width=13.5cm]{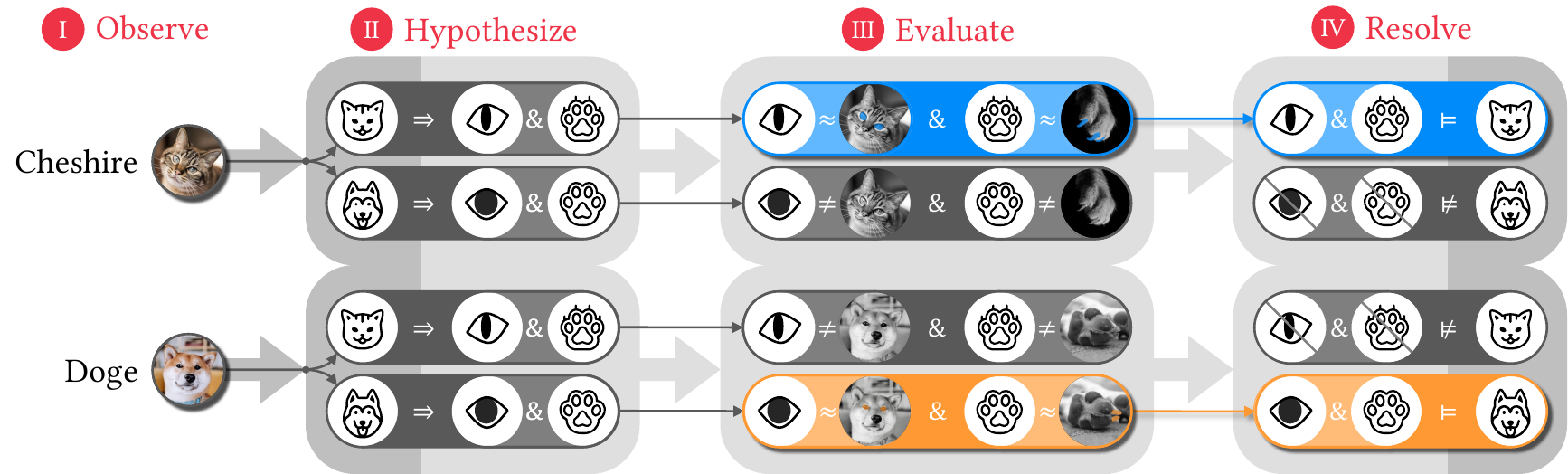}
    \vspace*{-0.25cm}
    \caption{
    Four steps of abductive reasoning demonstrated on a pedagogical scenario of inferring cats and dogs:
    I) observe the instance, 
    II) hypothesize causes and retrieve associated rules that determine the consequent evidence 
    (e.g., cat has vertical eye pupils and claws, dog has round pupils and nails), 
    III) evaluate the evidence on the observation then deductively backchain on each rule to determine the plausibility of its hypothesis
    (e.g., Cheshire's vertical pupils and claws imply it is a cat, but no round pupils or nails does not imply it is a dog), 
    and 
    IV) resolve the best explanation with highest plausibility, i.e., that Cheshire is a cat and Doge is a dog.
    Image credits: 
    ``{\color[HTML]{0060df}\underline{\href{https://thenounproject.com/icon/cat-1736115}{\smash{Cat}}}}'', 
    ``{\color[HTML]{0060df}\underline{\href{https://thenounproject.com/icon/dog-1609923}{\smash{Dog}}}}'', 
    ``{\color[HTML]{0060df}\underline{\href{https://thenounproject.com/icon/paw-1738438}{\smash{Paw}}}}'', and 
    ``{\color[HTML]{0060df}\underline{\href{https://thenounproject.com/icon/paw-1738440}{\smash{Paw}}}}'' 
    by {\color[HTML]{0060df}\underline{\href{https://thenounproject.com/maxim221}{\smash{Maxim Kulikov}}}}
    under the {\color[HTML]{0060df}\underline{\href{https://creativecommons.org/licenses/by/3.0/deed.en}{CC BY 3.0 license}}}.
    }
    \Description{
    A visual representation of the four steps of abductive reasoning applied to a pedagogical scenario involving two animals, labeled as ‘Cheshire’ and ‘Doge’. The process is depicted in four horizontally arranged panels, each corresponding to one step: observe, hypothesize, evaluate, and resolve.
    In the first panel (observe), there are two circular icons with magnifying glasses highlighting features of Cheshire and Doge. The second panel (hypothesize) shows two sets of three icons for each animal representing hypothesized causes based on observed features; for Cheshire, vertical eye pupils and claws are shown, while for Doge round pupils and nails are depicted.
    The third panel (evaluate) includes arrows pointing from the hypothesized causes back to the observed features along with checkmarks or crosses indicating whether the evidence supports the hypothesis. For Cheshire, both vertical pupils and claws have checkmarks; for Doge, round pupils have a cross while nails have a checkmark.
    The final panel (resolve) displays two circles around the icons that had checkmarks in the evaluation phase for each animal. This indicates that Cheshire’s features align with being a cat and Doge’s remaining feature aligns with being a dog.
    This figure illustrates how abductive reasoning can be used to infer correct classifications based on specific observable traits without repeating information already provided by the caption. It is relevant as it visually breaks down complex logical processes into understandable segments using familiar subjects like cats and dogs.
    }
    \label{fig:concept-abduction}
    \vspace{-0.20cm}
\end{figure*}

On observing an object or event, people engage in various reasoning processes. Peirce defined three types of inferential reasoning: induction, deduction, and abduction~\cite{peirce1903harvard}.
Inductive reasoning infers or learns general rules from features of multiple objects and events.
Deductive reasoning uses premises to infer labels or decisions based on observed evidence;
premises can be determined by logical proposition, probabilistic likelihood, learned rules, or cost-based criteria~\cite{leake1995abduction}.
Although abductive reasoning is used less frequently, it is central to diagnostic or investigative thinking.

Harman defined abduction as "inference to the best explanation" \cite{harman1965inference}; instead of inferring a label, this infers the underlying reason,
which could be causal or non-causal~\cite{williamson2016abductive}.
Popper described abduction as \textit{hypothetico-deductive} reasoning~\cite{popper2014conjectures}, which emphasizes that it involves forming and deductively testing hypotheses.
Like \cite{hoffman2017explaining, hoffman2020explaining, miller2023explainable, wang2019designing}, we argue that XAI should support user abductive reasoning over multiple hypotheses to justify the AI model's prediction.
We articulate the steps of the abduction reasoning process:
\begin{enumerate}
    \item[I.] \textit{Observe event}, noting relevant cues for further reasoning.    
    \item[II.] \textit{Hypothesize explanations} as potential causes of the observation that will manifest in specific consequent evidence.    
    \item[III.] \textit{Evaluate plausibility} by fitting each hypothesis to the observation and evaluating its corresponding evidence.
    \item[IV.] \textit{Resolve explanation} by selecting the \textit{simplest} hypothesis from any that have satisfied evidence. 
\end{enumerate}

Fig. \ref{fig:concept-abduction} illustrates the multi-step reasoning process of how people use abduction to retrieve and test hypotheses to understand their observations.
Fig. \ref{fig:concept-abduction-murmur} describes the abduction process for a more complex decision---cardiac diagnosis based on murmur diagrams.
With diagrams, users can perform abductive reasoning by experimenting with a different diagram for each hypothesis and evaluating their fit to the observation~\cite{hoffmann2010diagrams}.

\section{Domain application: Clinical background}

We investigate using abductive explanations of AI predictions in medicine.
Cardiovascular diseases cause an estimated 17.9 million worldwide deaths, accounting for 32\% of deaths in 2019~\cite{who2021cardiovascular}.
We aim to develop an early diagnosis AI system for heart disease 
to augment clinicians with deficient auscultation skills~\cite{alam2010cardiac}.
When predictions impact people's lives, it is critical to provide explanations for expert review.
Here, we describe the background to clarify how our AI explanations are clinically-relevant for practicing clinicians.

\begin{figure*}[t!]
    \centering
    \includegraphics[width=13.7cm]{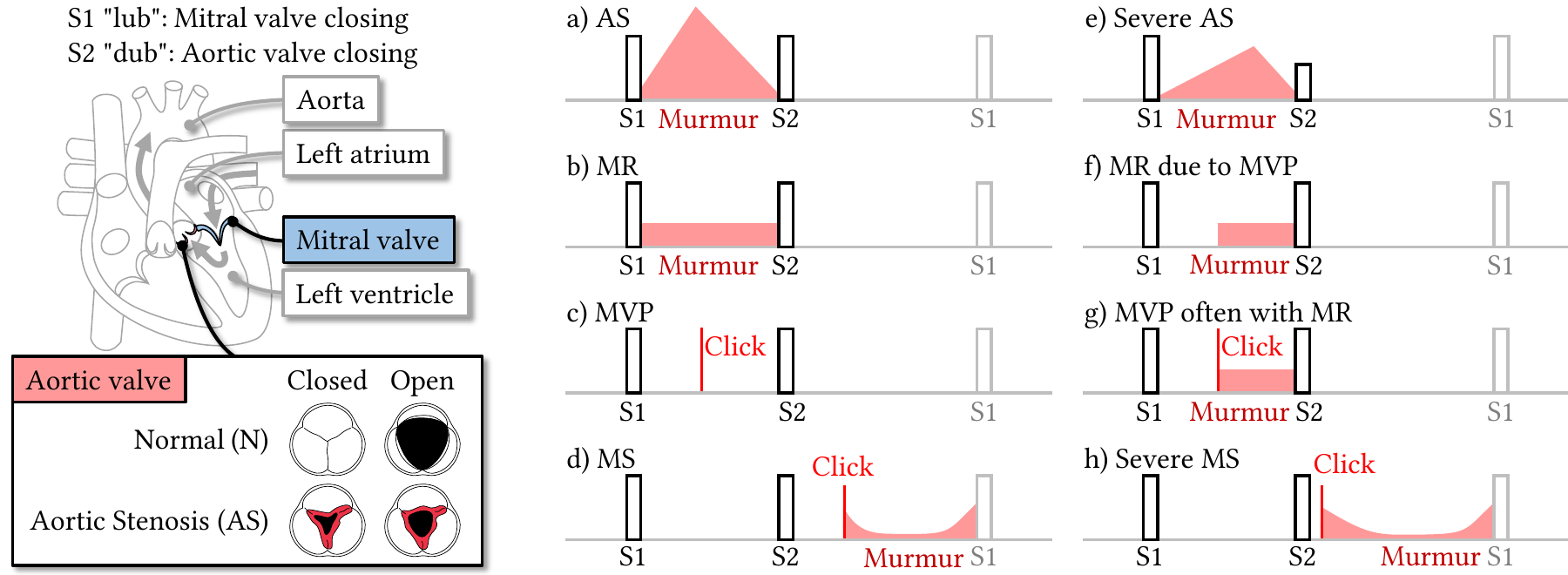}
    \vspace{-0.20cm}
    \caption{
    Upper-left:
    Anatomy and physiology of the heart to see how blood flows through the left atrium and ventricle via the aortic and mitral valves, which make lub-dub sounds on their closure. 
    Lower-left: 
    Aortic valve with normal or aortic stenosis pathology showing calcification (in red) that leads to stiffness, resulting the a crescendo-decresendo murmur sound.
    See \cite{lilly2012pathophysiology} for details on the physiological dynamics of other valvular diseases.
    Right:
    Example murmur diagrams showing typical murmurs for a) aortic stenosis (AS), b) mitral regurgitation (MR), c) mitral valve prolapse (MVP), d) mitral stenosis (MS), and their more severe variants (e--h)
    with slightly different shapes. 
    Black rectangles indicate normal "lub" (S1) and "dub" (S2) sounds. Red areas indicate abnormal murmur sounds.
    See Figs. \ref{fig:demo-explanations}--\ref{fig:demo-as} for diagrams on real PCGs.
    Image credits: 
    heart drawing adapted from {\color[HTML]{0060df}\underline{\href{https://commons.wikimedia.org/wiki/File:Diagram_of_the_human_heart_(valves_improved).svg}{\smash{Diagram of the human heart}}}} by {\color[HTML]{0060df}\underline{\href{https://en.wikipedia.org/wiki/de:User:Ungebeten}{\smash{Ungebeten}}}}, and
    valve drawings adapted from 
    {\color[HTML]{0060df}\underline{\href{https://commons.wikimedia.org/wiki/File:Aortic_valve_pathology_(CardioNetworks_ECHOpedia).svg}{\smash{Aortic valve pathology (CardioNetworks ECHOpedia)}}}}.
    All images under the {\color[HTML]{0060df}\underline{\href{https://creativecommons.org/licenses/by-sa/3.0/deed.en}{CC BY-SA 3.0 license}}}.
    }
    \Description{
    The image is a composite of diagrams illustrating the heart's anatomy, valve pathologies, and associated murmur sounds. Here is a detailed description of each part:
    \#\#\# Left Section
    **Upper-Left: Anatomy and Physiology of the Heart**
    - This diagram shows the heart's internal structure, focusing on the left atrium and ventricle. Arrows indicate the direction of blood flow through the aortic and mitral valves. The closure of these valves produces the "lub-dub" sounds heard during a heartbeat.
    **Lower-Left: Aortic Valve Pathology**
    - Two circular diagrams represent the aortic valve. The first shows a normal valve, while the second illustrates aortic stenosis with red areas indicating calcification. This calcification leads to stiffness in the valve, causing a crescendo-decrescendo murmur sound.
    \#\#\# Right Section: Murmur Diagrams
    **a) Aortic Stenosis (AS)**
    - This graph shows the typical murmur pattern for aortic stenosis. The murmur is represented by a red area between the normal "lub" (S1) and "dub" (S2) sounds, indicating a crescendo-decrescendo pattern.
    **b) Mitral Regurgitation (MR)**
    - This graph depicts the murmur pattern for mitral regurgitation. The red area extends throughout the systole, indicating a holosystolic murmur between S1 and S2.
    **c) Mitral Valve Prolapse (MVP)**
    - This graph shows the murmur pattern for mitral valve prolapse. A mid-systolic click followed by a late systolic murmur is represented by a red area after the click.
    **d) Mitral Stenosis (MS)**
    - This graph illustrates the murmur pattern for mitral stenosis. The red area represents a diastolic rumble following the "dub" (S2) sound.
    **e) Severe Aortic Stenosis (AS)**
    - This graph shows the murmur pattern for severe aortic stenosis. The red area is more pronounced, indicating a louder and longer crescendo-decrescendo murmur.
    **f) Severe Mitral Regurgitation (MR)**
    - This graph depicts the murmur pattern for severe mitral regurgitation. The red area is larger, indicating a more intense holosystolic murmur.
    **g) Severe Mitral Valve Prolapse (MVP)**
    - This graph shows the murmur pattern for severe mitral valve prolapse. The red area after the mid-systolic click is more extensive, indicating a louder late systolic murmur.
    **h) Severe Mitral Stenosis (MS)**
    - This graph illustrates the murmur pattern for severe mitral stenosis. The red area is larger, indicating a more pronounced diastolic rumble following the "dub" (S2) sound.
    Each part of the image provides a visual representation of different heart conditions and their associated murmur sounds, aiding in the understanding and diagnosis of valvular diseases.
    }
    \label{fig:concept-heart-murmur-diagrams}
    \vspace{-0.2cm}
\end{figure*}

\subsection{Heart auscultation}

Fig. \ref{fig:concept-heart-murmur-diagrams} (Upper-left) shows a partial heart cycle with blood flowing from the left atrium, into the left ventricle through the \textit{mitral valve}, and pumped out through the \textit{aortic valve}.
Valves prevent backward flow.
Their closing produces a "lub-dub" sound: 
the 1st heart sound (termed S1) "lub" is from the mitral valve, and the 2nd heart sound (S2) "dub" is from the aortic valve.
S1 and S2 demarcate the systolic (between S1 and S2) and diastolic phases of the heart cycle.
In heart auscultation, the clinician uses a stethoscope to listen for normal or abnormal sounds, and makes a \textit{first-line} cardiac diagnosis~\cite{judge2015heart, lilly2012pathophysiology}.
From this, the clinician can decide if follow-up tests are needed, e.g., echocardiogram, angiogram, but these are costly and invasive.
Abnormal heart sounds---murmurs---may indicate heart disease.
Clinicians make diagnoses by listening to changes in loudness.
These are represented in murmur diagrams~\cite{judge2015heart} (Fig. \ref{fig:concept-heart-murmur-diagrams}, Right).

\subsection{Diagrammatization with murmur diagrams}
With murmur diagrams, clinicians orientate themselves to the heart cycle by S1 and S2 locations, and note when any murmur occurs.
They pay attention to the shape of the murmur (one of possible shape types) and the heart phase in which it occurred.
Note that, in practice, clinicians mentally imagine these diagrams rather than explicitly draw them out.
We characterize murmur diagrams in the diagrammatization design space in context of the medical domain:
\begin{itemize}
    \item \textit{Representation ontology.}
    Key concepts are audio amplitude over time, normal "lub" (S1) and "dub" (S2) sounds, and abnormal murmur sounds. Murmurs can be systolic or diastolic, have shape categories with specific slopes (crescendo, decrescendo, uniform) and may include "clicks".
    \item \textit{Conventions for interpretation.} 
    Base: represent heart sounds with phonocardiograms (PCG) and draw amplitude. 
    Annotations: S1 and S2 positions are demarcated as tall rectangles, and murmur shapes are drawn with multi-part straight lines.
    \item \textit{Categorical and continuous level of states.}
    Each murmur shape must fit a categorical profile, but has continuous variation fit specific observations (e.g., slope steepness, time span length).
    \item \textit{Bounded expressivity.}
    Murmur diagrams show simplified amplitudes bounded to murmur shapes, but exclude
    other concepts (e.g., pitch, stethoscope position, sound radiation).
    \item \textit{High physical homomophism.}
    Murmur shapes can be overlaid on PCGs to fit to the amplitude, leveraging clinical training and familiarity on both PCGs and murmur diagrams.
    \item \textit{Geometrical inherent constraints.}
    Murmur shapes are geometrically constrained to be between S1-S2 or S2-S1, have positive, negative, or flat slopes, and be fitted to amplitude.
\end{itemize}

Hence, murmur diagrams are expressive, constrained, and conventional to convey explanations with murmur phase and shape from heart sounds for cardiac diagnosis.
Later, we describe our technical approach for the AI to scaffold abductive reasoning by evaluating hypotheses and presenting diagrammatic explanations.
By constraining to murmur diagram conventions and abductive diagnostic reasoning, we aim to increase clinician trust in the model.

\subsection{Abductive reasoning for cardiac diagnosis}\label{sec:murmurs}

Clinical diagnosis involves abductive inference to the most likely disease cause (best explanation) based on symptoms (evidence)~\cite{brachman2004knowledge_ch13, patel2005thinking}. 
We introduce the domain-specific concepts of cardiac valvular diseases that is reasoned on,
focusing on four prevalent diseases.

\begin{enumerate}

\item[1)] \textit{Aortic Stenosis (AS):}
the aortic valve leaflets are calcified and stiff 
(Fig. \ref{fig:concept-heart-murmur-diagrams}, Lower-left),
narrowing the valve opening (i.e., stenosis), 
causing a high-pitched noise that loudens as the valve opens and softens as it closes.
This systolic \textit{crescendo-descresendo} murmur is visualized as a diamond shape (Fig. \ref{fig:concept-heart-murmur-diagrams}a).
In severe AS, the shape apex is later and lower, due to delayed valve closure and weaker heart (Fig. \ref{fig:concept-heart-murmur-diagrams}e).

\item[2)] \textit{Mitral Regurgitation (MR):}
the mitral valve fails to fully close, allowing backward blood flow (i.e., regurgitation),
heard as a constant, high-pitched murmur during the systolic heart phase.
This is visualized as a \textit{uniform} low-amplitude sound (Fig. \ref{fig:concept-heart-murmur-diagrams}b).
Sometimes the valve opens mid-systole (Fig. \ref{fig:concept-heart-murmur-diagrams}f).

\item[3)] \textit{Mitral Valve Prolapse (MVP):}
the tendons keeping the mitral valve closed fails, causing the valve to pop open (prolapse), allowing blood to regurgitate.
This opening is heard as a mid-systolic "click", visualized as a vertical line (Fig. \ref{fig:concept-heart-murmur-diagrams}c).
Often the regurgitation is audible as a uniform, high-pitched murmur, which is MVP with MR (Fig. \ref{fig:concept-heart-murmur-diagrams}g).

\item[4)] \textit{Mitral Stenosis (MS):}
the mitral valve leaflets fuse (i.e., stenosis) due to rheumatic heart disease,
reducing blood flow during the \textit{diastolic} heart phase (Fig. \ref{fig:concept-heart-murmur-diagrams}d).
After the S2 "dub", 
the valve snaps open with a "click" sound, 
enabling blood flow, followed by a decrescendo as flow reduces, then a constant low-pitch "rumble", and a crescendo before the next S1.
Severe MS has an earlier "click" during diastole and longer murmur decrescendo (Fig. \ref{fig:concept-heart-murmur-diagrams}h).

\end{enumerate}

\begin{figure*}[t!]
    \centering
    \includegraphics[width=11.6cm]{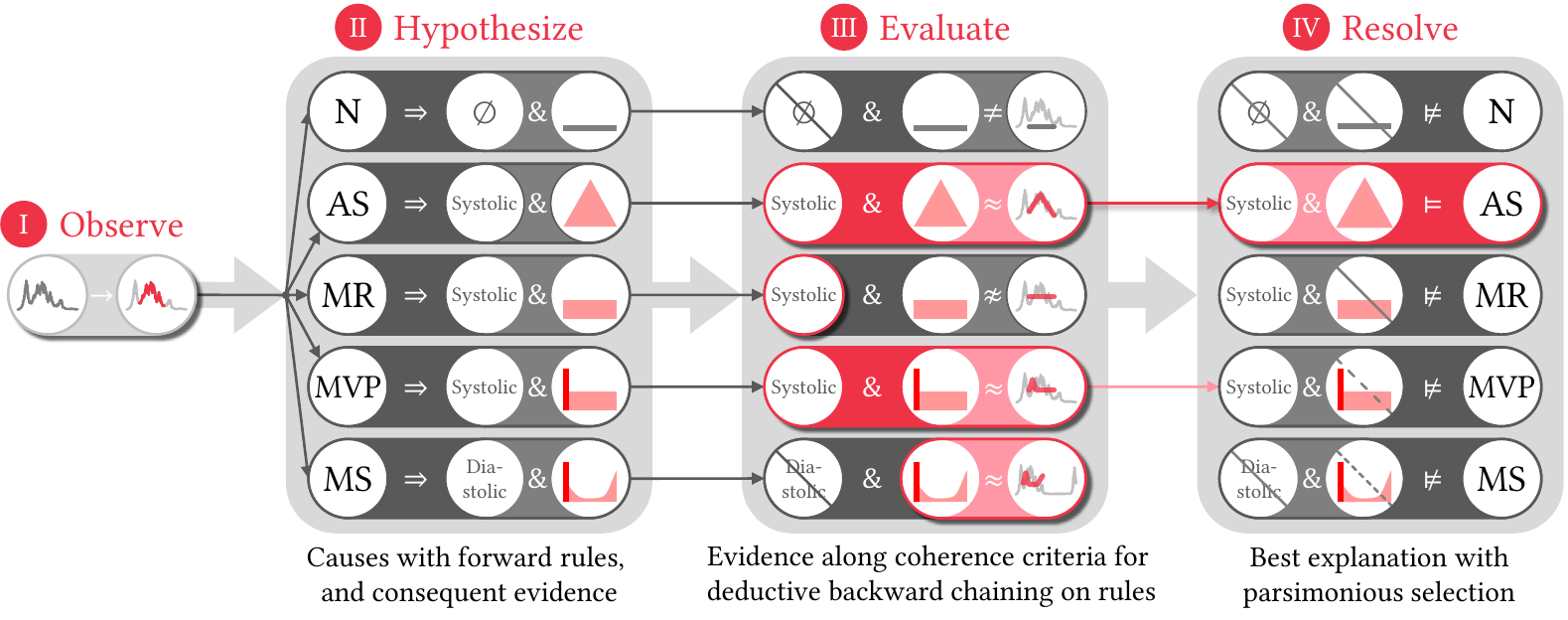}
    \vspace{-0.30cm}
    \caption{Abductive reasoning for cardiac diagnosis to
    I) observe an abnormal murmur,
    II) hypothesize possible diagnoses (N, AS, MR, MVP, MS) and retrieve corresponding rules that relate to evident symptoms of murmur heart phase and murmur shape, 
    III) evaluate the coherence of all symptoms with respect to the observation and deduce via backward chaining on each hypothesis rule to determine the plausibility of each hypothesis, and
    IV) resolve to select the plausible explanation with simplest murmur shape function to infer the best fit to the observation (AS in this case).
    Although the murmur shapes for MVP and MS could fit the murmur equally well as that for AS, these two shapes are increasingly complex, and thus less preferred based on the principle of parsimony. False premises are indicated with a cross negation line, and less preferred premises with dashed lines.
    }
    \Description{
    The image presents a flowchart illustrating the abductive reasoning process used in cardiac diagnosis. Here is a detailed description of the layout, with emphasis on the red blocks within each relevant step:
    \#\#\# Layout Description
    - **Step I: Observe**
    - Located at the top left, this section shows an abnormal murmur waveform with a red segment over the other normal waveform in grey.
    - **Step II: Hypothesize**
    - Positioned below the "Observe" section, this part lists possible diagnoses: Normal (N), Aortic Stenosis (AS), Mitral Regurgitation (MR), Mitral Valve Prolapse (MVP), and Mitral Stenosis (MS).
    - Each diagnosis is associated with forward rules and consequent evidence characterized by murmur heart phase and shape.
    - **Red Blocks**: Highlight the calcification in the aortic valve for Aortic Stenosis (AS), indicating the presence of a murmur. These blocks are crucial as they visually represent the abnormality leading to the murmur.
    - **Step III: Evaluate**
    - Found to the right of the "Hypothesize" section, this part evaluates the coherence of all symptoms with respect to the observation.
    - It uses backward chaining on each hypothesis rule to determine the plausibility of each hypothesis.
    - **Red Blocks**:
    - Mark false premises with a cross negation line over Normal (N) and Mitral Regurgitation (MR), indicating these diagnoses are not supported by the observed murmur.
    - Less preferred premises like Mitral Valve Prolapse (MVP) and Mitral Stenosis (MS) are indicated with dashed lines due to their complexity, showing they are less likely but still possible.
    - **Step IV: Resolve**
    - Located at the far right, this section concludes with selecting Aortic Stenosis as the plausible explanation.
    - It emphasizes the principle of parsimony, preferring the simplest murmur shape function that best fits the observation.
    - Untrue evidence indicated with diagonal solid or dashed lines.
    - Untrue conclusions indicated with not-entailed symbol.
    - **Red Blocks**: Highlight the selection of Aortic Stenosis (AS) as the most plausible diagnosis based on the simplest murmur shape function. This is visually emphasized to show the final decision in the diagnostic process.
    This layout and the use of red blocks help in visually distinguishing the critical elements of the diagnostic process, making it easier to follow the reasoning steps and understand the final diagnosis.
    }
    \label{fig:concept-abduction-murmur}
    \vspace{-0.25cm}
\end{figure*}

With the aforementioned medical background and know-how to interpret murmur diagrams, the clinician can diagnose using abductive reasoning as shown in Fig. \ref{fig:concept-abduction-murmur}.
On observing an abnormal murmur, the clinician hypothesizes plausible diseases, and evaluates based on their expected symptoms.
The murmur heart phase was evaluated to be Systolic, thus ruling out N and MS. 
Furthermore, the murmur shape was evaluated to fit that of AS, MVP or MS, but the latter two are overly-complex. 
Being parsimonious, the clinician would choose the simplest explanation and diagnose AS.

\subsection{Related work on medical XAI}
Several works have pursued XAI in medicine due to its critical need~\cite{lim2009assessing}.
Wang et al. showed how XAI can mitigate cognitive biases in medical diagnoses~\cite{wang2019designing}.
Cai et al. identified requirements for trust in medical AI, including to \textit{``compare and contrast AI schemas relative to known human decision-making schemas''}~\cite{cai2019hello}.
SMILEY was designed to find similar pathology cases by region, example and concept-based explanations~\cite{cai2019human}.
Lundberg et al. proposed tree-based explanations to address \textit{``model mismatch -- where the true relationships in data do not match the form of the model''}~\cite{lundberg2020local}.
Tjoa and Guan's review of medical XAI identified several challenges, including the lack of interpretability, explanation unfaithfulness, and need for data science training in medical education~\cite{tjoa2020survey}. 
In contrast, Vellido argued for the \textit{``need to integrate the medical experts in the design of data analysis interpretation strategies''}~\cite{vellido2020importance}.
Similarly, we use diagrammatization and abduction to imbue medical expertise into XAI.
Corti et al. found that clinicians desire explanations that are multimodal, interactive and actionable~\cite{corti2024moving}. 
We focus on another requirement: domain alignment for clinical relevance.

Regarding our focus on AI for cardiac disease,
much work has been on electrocardiogram (ECG)~\cite{siontis2021artificial}, and less on phonocardiograms (PCG).
Yet, the few works on PCG focus on classifying normal or abnormal sounds~\cite{rubin2017recognizing}, or segmenting time~\cite{dwivedi2018algorithms}. 
These lack clinical usefulness, since they do not provide a differential diagnosis of multiple plausible diagnoses.
Work on XAI for PCGs is even more sparse,
focusing on saliency maps on spectrograms~\cite{dissanayake2020robust, raza2022designing, ren2022deep},
which we show later are unconvincing to clinicians.

\section{Technical approach}

We developed an ante-hoc interpretable model for high-stakes cardiac diagnosis from phonocardiograms (PCG). 
Aligning to clinical practice, the model performs diagrammatization for murmur diagrams and abduction for diagnostic reasoning.
We describe our data preparation, diagram formalization, and proposed model.

\subsection{Heart auscultation data preparation}
\subsubsection{Dataset}
We trained models on the the dataset by Yaseen et al. \cite{yaseen2018classification} comprising 1000 audio recordings of heart cycles, each 1.15--3.99s long, sampled at 8 kHz.
There are 200 recordings of each diagnosis: N, AS, MR, MVP, and MS.

\subsubsection{Preprocessing}
To classify auscultations starting at any time point, we created instances based on sliding windows with length 1.0s (8000 samples) and stride 0.1s. 
The window length was chosen so that each instance contained only one heart cycle
to simplify murmur predictions.
In total, we had 14,672 instances, which was sufficiently large for deep learning achieving 86.0\% for a base CNN, and 95.7\% for our proposed model. 
We split the data into 50\% training and test sets, and
ensured all time windows for the same audio files were only in either set.
We then extracted the time series audio amplitude $\bm{a} = \mathscr{A}(\bm{x})$ to estimate murmur shapes later.

\subsubsection{Annotation}
The dataset only had diagnosis labels, so we manually annotated the murmur start $\tau_1$ and end $\tau_L$, respectively.
Using this segment, we fit a nonlinear function describing the correct murmur shape of each time series instance, described in Section \ref{subsection:formalization-murmur-shapes}.
The annotations were performed and verified in consultation with our cardiologist collaborators.
Since it was prohibitively expensive to recruit clinicians as annotators, we trained ourselves (computer scientists) to understand the domain concepts of auscultation and murmurs.
Two annotators checked the annotations for consistency with each diagnosis as described in Section \ref{sec:murmurs} using time series visualizations of all PCGs.

\subsection{Diagram ontology: Murmur shapes as piecewise linear functions} \label{subsection:formalization-murmur-shapes}

To constrain murmur diagrams with the ontology of murmur shapes,
we formulated murmur shapes as parametric functions $\mathscr{F}_y(t|\bm{\theta})$ over time $t$ with parameters $\bm{\theta} = (\bm{\tau}, \bm{\pi})$.
We modeled crescendo, decrescendo, and uniform slopes as straight lines, and the full shape as a \textit{piecewise linear function}. 
Other considered function families were problematic.
Taylor series would include clinically irrelevant, spurious mathematical artifacts. 
Fourier series, which spectrograms actually represent, would capture frequency information in murmurs, but not the amplitude shapes.

\begin{table*}[t]
\small
\centering
\caption{
    Formalization of murmur shapes as piecewise linear functions $\mathscr{F}_y({\color[HTML]{0070C0}t}|\bm{\theta})$ of murmur amplitude (red shape) changes over time ${\color[HTML]{0070C0}t}$.
    $[]$ represents the Iverson bracket, which is 1 if its internal expression is true and 0 otherwise.
}
\vspace{-0.20cm}
\label{table:math-murmur-functions}
\begin{tabular}{clrlccc}
\toprule
Diagnosis $y$ &
   &
  \multicolumn{1}{c}{Murmur Diagram} &
   &
  Heart Phase $\phi$ &
  Shape Function $\mathscr{F}_y({\color[HTML]{0070C0}t}|\bm{\theta})$ &
  Parameters $\bm{\theta}$ \\ \cmidrule{1-1} \cmidrule{3-3} \cmidrule{5-7} 
N &
   &  
   \raisebox{-0.4cm}{\includegraphics[width=3.5cm]{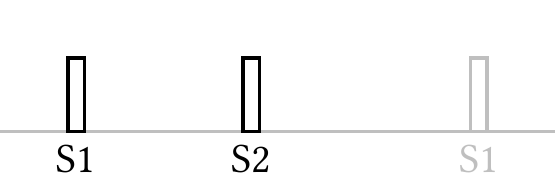}} &
   &
  {\color[HTML]{C0C0C0} \textit{n.a.}} &
  {\color[HTML]{C0C0C0} 0} &
  {\color[HTML]{C0C0C0} $\varnothing$} \\
AS &
   &
   \raisebox{-0.4cm}{\includegraphics[width=3.5cm]{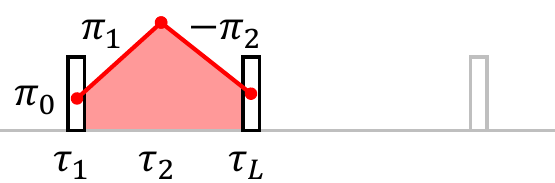}} &
   &
  Systolic &
  \begin{tabular}[c]{@{}l@{}}$[\tau_1 \leq {\color[HTML]{0070C0}t} < \tau_L](\pi_0 + \pi_1({\color[HTML]{0070C0}t} - \tau_1) $\\ $+ [\tau_2 \leq {\color[HTML]{0070C0}t}](-(\pi_1 + \pi_2)({\color[HTML]{0070C0}t} - \tau_2)))$\end{tabular} &
  \begin{tabular}[c]{@{}c@{}}$\tau_1, \tau_L,$\\ $\pi_0, \pi_1, \pi_2$\end{tabular} \\
MR &
   &
   \raisebox{-0.4cm}{\includegraphics[width=3.5cm]{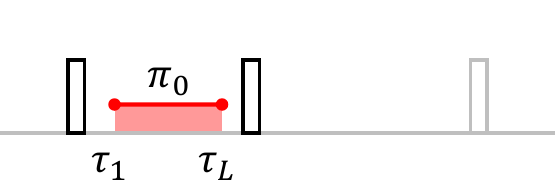}} &
   &
  Systolic &
  $[\tau_1 \leq {\color[HTML]{0070C0}t} < \tau_L] \pi_0$ &
  \begin{tabular}[c]{@{}c@{}}$\tau_1, \tau_L,$\\ $\pi_0$\end{tabular} \\
MVP &
   &
   \raisebox{-0.4cm}{\includegraphics[width=3.5cm]{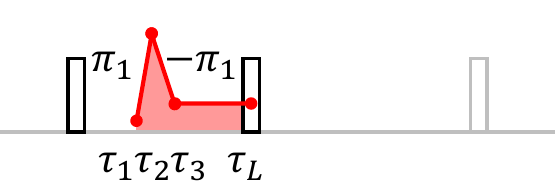}} &
   &
  Systolic &
  \begin{tabular}[c]{@{}l@{}}$[\tau_1 \leq {\color[HTML]{0070C0}t} < \tau_L](\pi_0 + \pi_1({\color[HTML]{0070C0}t} - \tau_1) $\\ $+ [\tau_2 \leq t](-2\pi_1({\color[HTML]{0070C0}t} - \tau_2)$\\ $+ [\tau_3 \leq {\color[HTML]{0070C0}t}](\pi_1({\color[HTML]{0070C0}t} - \tau_3))))$\end{tabular} &
  \begin{tabular}[c]{@{}c@{}}$\tau_1, \tau_2, \tau_3, \tau_L,$\\ $\pi_0, \pi_1$\end{tabular} \\
  \addlinespace[1.0mm]
MS &
   &
   \raisebox{-0.4cm}{\includegraphics[width=3.5cm]{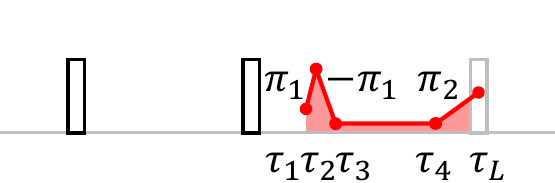}} &
   &
  Diastolic &
  \begin{tabular}[c]{@{}l@{}}$[\tau_1 \leq {\color[HTML]{0070C0}t} < \tau_L](\pi_0 + \pi_1({\color[HTML]{0070C0}t} - \tau_1) $\\ $+ [\tau_2 \leq {\color[HTML]{0070C0}t}](-2\pi_1({\color[HTML]{0070C0}t} - \tau_2)$\\ $+ [\tau_3 \leq {\color[HTML]{0070C0}t}](\pi_1({\color[HTML]{0070C0}t} - \tau_3)$\\ $+ [\tau_4 \leq {\color[HTML]{0070C0}t}](\pi_2({\color[HTML]{0070C0}t} - \tau_4)))))$\end{tabular} &
  \begin{tabular}[c]{@{}c@{}}$\tau_1, \tau_2, \tau_3, \tau_4, \tau_L,$\\ $\pi_0, \pi_1, \pi_2$\end{tabular} \\ 
  \bottomrule
\end{tabular}
\vspace{-0.20cm}
\end{table*}

All candidate murmur shapes share the murmur segment start $\tau_1$ and end $\tau_L$ time parameters, but can have varying number of time $\bm{\tau}$ and slope $\bm{\pi}$ parameters depending on the complexity of the shape.
Crescendos are modeled as lines with positive slope, decrescendos as lines with negative slope, and uniform with 0 slope.
Table \ref{table:math-murmur-functions} illustrates the murmur shapes mathematically with relevant parameters $\bm{\theta}$, and their shape function $\mathscr{F}_y(t|\bm{\theta})$ equations:

\begin{enumerate}
    \item[1)] \textit{Normal (N)}
    has no murmurs, so murmur segment start $\tau_1$ and end $\tau_L$ are undefined $\varnothing$, and $\mathscr{F}_N(t|\bm{\theta}) = 0$ by definition.
    \item[2)] \textit{Aortic stenosis (AS)}
    has a crescendo-decrescendo murmur starting at $\pi_0$ with positive slope $\pi_1$ from $\tau_1$ to $\tau_2$ and negative slope $-\pi_2$ from $\tau_2$ to $\tau_L$.
    \item[3)] \textit{Mitral regurgitation (MR)}
    has a uniform murmur between $\tau_1$ and $\tau_L$ at amplitude level $\pi_0$.
    \item[4)] \textit{Mitral valve prolapse (MVP)}
    murmurs start with a "click", modeled as a short spike with slopes $\pi_1$ and $-\pi_1$ from $\tau_1$ through $\tau_2$ to $\tau_3$,
    then uniform murmur spanning $\tau_3$ to $\tau_L$ with 0 slope, 
    which may have 0 amplitude if no MR.
    \item[5)] \textit{Mitral stenosis (MS)}
    has similar shape to MVP, but 
    this murmur happens in the diastolic heart phase, not systolic, and
    ends with a positive slope $\pi_2$ crescendo from $\tau_4$ to $\tau_L$.
\end{enumerate}

\subsection{DiagramNet: Diagrammatic network with abductive explanations of murmur shapes}\label{sec:DiagramNet}

\begin{figure*}[t]
    \centering
    \includegraphics[width=14.5cm]{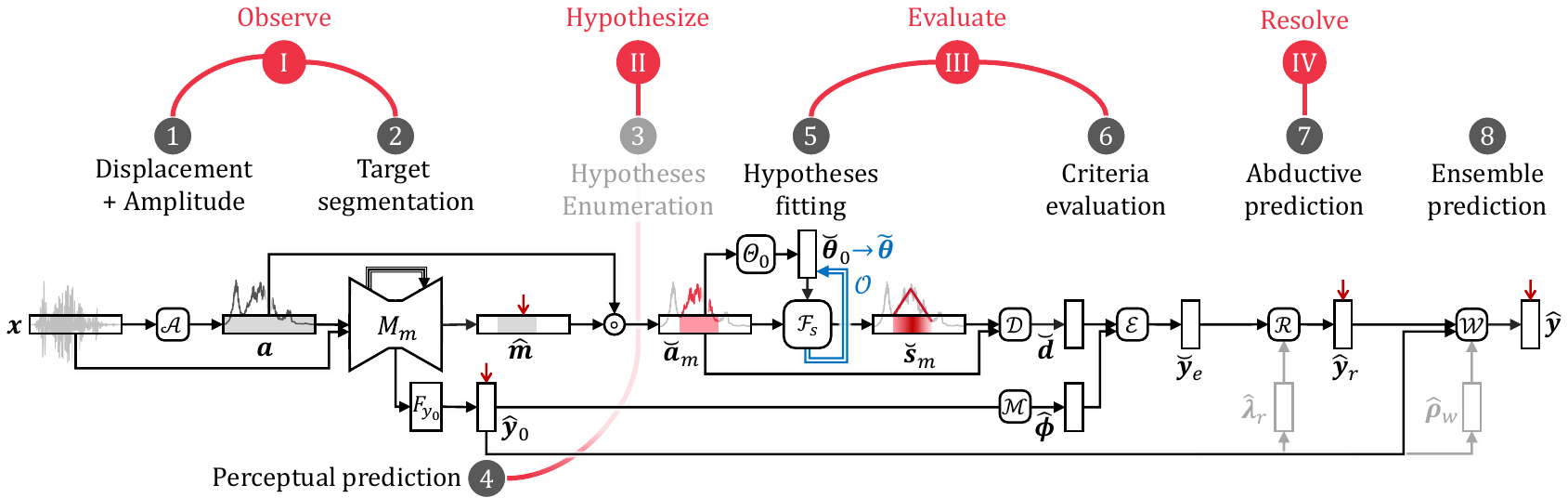}
    \vspace{-0.25cm}
    \caption{
    Modular architecture of DiagramNet with
    7 stages corresponding to the steps of abductive reasoning (I to IV) in Section \ref{subsection:abductive-reasoning}, and 
    the 8th stage for ensemble prediction to combine perceptual and abductive predictions.
    Black arrows indicate feedforward activations, the blue arrow indicates an iterative nonlinear optimization at inference time to estimate the final murmur shape parameters, and red downward arrows indicate which variables are trained with supervised labels. 
    Variables are annotated with \textbf{bold} for vectors or tensors, $\hat{\phantom{x}}$ for predictions from trained modules, $\Breve{\phantom{x}}$ for heuristically-calculated values, and $\tilde{\phantom{x}}$ for those optimized at inference time.
    $\circ$ is the Hardamard operator for element-wise multiplication.
    Narrow rectangles indicate an input or predicted variable.
    Other shapes indicate processes: trainable neural network blocks as rectangle $F_{y_0}$ or trapezoid $M_m$ (capital letters), non-trainable heuristic processes as rounded squares (script letters), and vector operators (circles).
    }
    \Description{
    The figure presents the modular architecture of DiagramNet. The architecture (shown as a flowchart) is divided into eight stages, each corresponding to specific steps in the reasoning process. Steps contain several variables shown as blocks connected via arrows through blocks for sub-models or functions. Stages 1-8 flow from left to right, and are grouped into Steps I to IV.
    - Stage I contains Step 1 Displacement + Amplitude and Step 2 Target segmentation.
    - Stage II contains Step 3 Hypothesis enumeration and Step 4 Perceptual prediction.
    - Stage III contains Step 5 Hypothesis fitting and Step 6 Criteria evaluation.
    - Stage IV includes Step 7 Abductive prediction.
    - Step 8 Ensemble prediction is not connected to Stages I-IV.
    }
    \label{fig:model-DiagramNet}
    \vspace{-0.15cm}
\end{figure*}

We introduce DiagramNet, a deep neural network meta-architecture inspired by the multi-step abductive reasoning process to infer the best explanation as the prediction (see Fig. \ref{fig:model-DiagramNet}). 
Unlike standard neural networks that tend to learn spurious and unintelligible neural activation, 
DiagramNet aligns to the clinical diagnosis domain by constraining with structural priors~\cite{hudson2018compositional} 
to implement the 4-step abduction process in 7 stages for perceptual and abductive predictions, and combine them for an ensemble prediction. 
It supports high-stakes decisions through its ante-hoc interpretability.

\subsubsection{Audio displacement and amplitude inputs}
Given the 1-sec (8000-sample) audio data as displacement $\bm{x}$, we extract the amplitude $\bm{a}$, concatenate them as a 2-channel 1D tensor $(\bm{x},\bm{a})$.
Although the convolutional layers of the CNN could learn frequency information from $\bm{x}$, explicitly computing $\bm{a}$ makes it easier for the model to learn patterns from amplitude.

\subsubsection{Murmur segmentation}
Next, we input $(\bm{x},\bm{a})$ into a U-Net~\cite{ronneberger2015unet} model $M_m$ to predict the time region\footnote{We used U-Net, popular for image segmentation~\cite{kohl2018probabilistic, ronneberger2015unet}, on time series by treating it as a 1D "image" to use convolutional filters on temporal motifs as spatial patterns.} of the murmur $\hat{\bm{m}}$, defined as a mask vector.
However, this suffers from \textit{over-segmentation} by inferring multiple regions of murmurs in a single instance, although there should only be one. 
As in \cite{farha2019ms, matsuyama2023iris}, we resolve this with a smoothing loss using the truncated mean squared error: 
$L_\mu = \frac{1}{T} \sum_t^T \max(\epsilon_t, \epsilon)$, where $\epsilon_t = (\log \hat{m}(t) - \log \hat{m}(t-1))^2$ is the squared of log differences and $\epsilon$ is the truncation hyperparameter.
This may still result in >1 segments, so we choose the longest one.

\subsubsection{Hypothesis enumeration}
Common to clinical diagnosis requiring rapid decisions, we focus on \textit{selective} abduction to enumerate hypothesis from a predetermined set~\cite{magnani2011abduction}. 
We enumerate hypotheses $y$ (N, AS, MR, MVP, MS) for classification in Stages 4 and 7, and retrieve corresponding murmur shape functions $\mathscr{F}_s^{(y)}(t|\bm{\theta})$ with yet-estimated parameters $\bm{\theta}$ to be evaluated in Stage 6.
Inferring on multiple hypotheses is canonical of abductive reasoning and goes beyond simple neural network classifiers that only use one hypothesis, i.e., one model with trained parameters.

\subsubsection{Perceptual prediction}
Using the embedding representation learned from murmur segmentation by
feeding it into fully-connected layers $F_{y_0}$, we predict an initial diagnosis $\hat{\bm{y}}_0$ based on perception.
This would be more accurate than a base CNN, since it benefits from the added multi-task learning to predict $\hat{\bm{m}}$ too.

\subsubsection{Hypotheses fitting}

Before evaluating each hypothesis, we need to fit its shape function to the time series observation. 
First, we extract the murmur amplitude by masking the amplitude $\bm{a}$ within the murmur segment $\hat{\bm{m}}$ from $\tau_1=\hat{m}_1$ to $\tau_L=\hat{m}_L$, i.e., $\Breve{\bm{a}}_m = \bm{a} \circ \hat{\bm{m}}$, where $\circ$ is the Hadamard element-wise multiplication.
Next, We initialize the shape parameter values using heuristics based on typical characteristics $\Breve{\bm{\theta}}_0 = \Theta_0(\Breve{\bm{a}}_m)$ described as follows.

\begin{enumerate}
    \item[1)] \textit{Normal (N).}
    No parameters, as no murmur expected.
    \item[2)] \textit{Aortic stenosis (AS).}
    We estimate the apex of the crescendo-decresendo to occur at the time $\tau_2 = \argmax_t(a)$ of highest amplitude $\max(a)$.
    $\pi_0$ is just the amplitude at $\tau_1$, $\pi_1$ is the slope from the murmur start to apex, and $\pi_2 \approx \pi_1$.
    \item[3)] \textit{Mitral regurgitation (MR).}
    The shape is a flat line at the average amplitude of the murmur segment $\bar{a}_m = \sum_{\tau_1 < t < \tau_L} a(t)$.
    \item[4)] \textit{Mitral valve prolapse (MVP).}
    We estimate the apex at $\tau_2$ similarly as for AS, $\tau_3$ to occur at twice the distance from $\tau_1$ to $\tau_2$.
    $\pi_0$ and $\pi_1$ are calculated the same way as for AS.
    \item[5)] \textit{Mitral stenosis (MS).}
    Estimating time parameters is poor using heuristics, so we use a data-driven approach by using the median of time differences from the training dataset. These are calculated for $\tau_2$ and $\tau_4$ relative to murmur start $\tau_1$ and end $\tau_L$.
    $\tau_3$, $\pi_0$ and $\pi_1$ are calculated the same way as for MVP.
\end{enumerate}

Starting with initial parameter values $\Breve{\bm{\theta}}_0$, we optimize $\mathscr{O}$ the murmur shapes for all diagnoses $\Breve{\bm{s}}_m = \mathscr{F}_s(t|\Breve{\bm{\theta}}_0)$
using the Limited-memory Broyden-Fletcher-Goldfarb-Shanno (L-BFGS) algorithm \cite{nocedal2006numerical} to minimize the shape fit mean square error (MSE), i.e., $\tilde{\bm{\theta}} = \argmin_{\bm{\theta}} \left\| \mathscr{F}_t(t|\bm{\theta}) - \Breve{\bm{a}}_m \right\|_2^2$. 
Like activation maximization~\cite{nguyen2016synthesizing} and CLIP~\cite{radford2021learning}, but unlike standard supervised learning that optimizes model parameters at training, shape functions are optimized \textit{at inference} per instance.
See Table \ref{table:math-initial-params} for summary.

\begin{table*}[t]
\small
\centering
\caption{
    Heuristics to initialize murmur parameters for each plausible diagnosis $y$ based on predicted murmur segment start $\hat{m}_1$ and end $\hat{m}_L$, and data statistics (for MS).
    $a$ is all amplitudes in the murmur, $\bar{a}_m$ is the average amplitude, $a(t)$ is the amplitude at time $t$, $\Delta\tau_{12} = \tau_2 - \tau_1$, $\Delta\tau_{4L} = \tau_L - \tau_4$, and $\mu_{0.5}(\Delta\tau)$ is the median of training instances for $\Delta\tau$.
    Parameters undefined $\varnothing$ for $y = \text{N}$.
}
\label{table:math-initial-params}
\vspace{-0.15cm}
\begin{tabular}{rlccccclccc}
\toprule
 &
   &
  \multicolumn{5}{c}{Initial Time Parameters} &
   &
  \multicolumn{3}{c}{Initial Slope Parameters} \\ \cline{3-7} \cline{9-11} 
$y$ &
   &
  $\tau_1$ &
  $\tau_2$ &
  $\tau_3$ &
  $\tau_4$ &
  $\tau_L$ &
   &
  $\pi_0$ &
  $\pi_1$ &
  $\pi_2$ \\ \cmidrule{1-1} \cmidrule{3-7} \cmidrule{9-11} 
AS &
   &
  \textit{$\hat{m}_1$} &
  $\argmax_t (a)$ &
  {\color[HTML]{C0C0C0} $\varnothing$} &
  {\color[HTML]{C0C0C0} $\varnothing$} &
  \textit{$\hat{m}_L$} &
   &
  $a(\tau_1)$ &
  $\frac{a(\tau_2)-\pi_0}{\Delta\tau_{12}}$ &
  $\pi_1$ \\
    \addlinespace[0.1cm]
MR &
   &
  \textit{$\hat{m}_1$} &
  {\color[HTML]{C0C0C0} $\varnothing$} &
  {\color[HTML]{C0C0C0} $\varnothing$} &
  {\color[HTML]{C0C0C0} $\varnothing$} &
  \textit{$\hat{m}_L$} &
   &
  $\bar{a}_m$ &
  {\color[HTML]{C0C0C0} $\varnothing$} &
  {\color[HTML]{C0C0C0} $\varnothing$} \\
    \addlinespace[0.1cm]
MVP &
   &
  \textit{$\hat{m}_1$} &
  $\argmax_t (a)$ &
  $\tau_1 + 2\Delta\tau_{12}$ &
  {\color[HTML]{C0C0C0} $\varnothing$} &
  \textit{$\hat{m}_L$} &
   &
  $a(\tau_1)$ &
  $\frac{a(\tau_2)-\pi_0}{\Delta\tau_{12}}$ &
  {\color[HTML]{C0C0C0} $\varnothing$} \\
    \addlinespace[0.1cm]
MS &
   &
  \textit{$\hat{m}_1$} &
  $\tau_1 + \mu_{0.5}(\Delta\tau_{12})$ &
  $\tau_1 + 2\Delta\tau_{12}$ &
  $\tau_L - \mu_{0.5}(\Delta\tau_{4L})$ &
  \textit{$\hat{m}_L$} &
   &
  $a(\tau_1)$ &
  $\frac{a(\tau_2)-\pi_0}{\Delta\tau_{12}}$ &
  $\frac{a(\tau_L)-a(\tau_4)}{\Delta\tau_{4L}}$ \\ 
  \bottomrule
\end{tabular}
\end{table*}

\subsubsection{Criteria evaluation}

Now, we determine the evidence for each hypothesis. Similar to other abduction-based approaches, we evaluate each hypothesis with cost-based and probabilistic coherence criteria~\cite{leake1995abduction, morvan2008simulation} by calculating cost scores to assess how well its hypothesized evidence fits the observation. 
For heart auscultation, we use murmur shape goodness-of-fit as a cost-based criterion and murmur heart phase prediction likelihood as probabilistic criterion.

We compute the murmur shape goodness-of-fit as the mean square error (MSE) distance between the murmur amplitude $\Breve{\bm{a}}_m$ and optimized murmur shape for each diagnosis $\Breve{\bm{s}}_m^{(y)}$, i.e., $\Breve{\bm{d}} = \mathscr{D}(\Breve{\bm{a}}_m, \Breve{\bm{s}}_m) \coloneqq \left\| \Breve{\bm{a}}_m - \Breve{\bm{s}}_m \right\|_2^2$.
We invert\footnote{We chose inverse ($1/d$), which leads to better model performance than reverse ($1-d$).} this distance metric to get the \textit{murmur shape criterion score} of each hypothesis, i.e., $\Breve{\bm{\gamma}}_d = \Breve{\bm{d}}^{\circ -1}$, where ${\scriptstyle{\square}}^{\circ -1}$ is the Hadamard inverse operation for elementwise inverse. 
Next, we heuristically determine\footnote{Another method to estimate the murmur heart phase is to segment the S1 and S2 locations and determine whether the murmur is absent ($\varnothing$), between S1-S2 (Systolic), or between S2-S1 (Diastolic)~\cite{renna2019deep}.} the likelihood of murmur heart phase based on the perceptual prediction $\hat{\bm{y}}_0$ by transformation $\hat{\bm{\phi}} = \mathscr{M}(\hat{\bm{y}}_0) \coloneqq M_{y \phi}^\top \hat{\bm{y}}_0$, where $M_{y\phi} = \left( \bm{e}_0, \bm{e}_1, \bm{e}_1, \bm{e}_1, \bm{e}_2 \right)^\top$ is a matrix to map diagnosis to heart phase, and $\bm{e}_0$, $\bm{e}_1$, $\bm{e}_2$ are standard basis vectors.
We compute the \textit{murmur heart phase criterion score} of each hypothesis by mapping diseases with the same phase to the same likelihood, i.e., $\hat{\bm{y}}_\phi = M_{y \phi}^\top \hat{\bm{\phi}}$.

We combine these criteria to determine the overall plausibility of each hypothesis over all evidence by conjoining the rules for murmur heart phase and shape, i.e., $\hat{\phi}^{(y)} \wedge \Breve{s}_m^{(y)} \vDash \Breve{y}_e$. We simulate this logical conjunction in the neural network as the multiplication of the two likelihood density vectors $\hat{\bm{y}}_\phi$ and $\Breve{\bm{\gamma}}_d$ with normalization, 
\[
\hat{\phi} \wedge \Breve{s}_m \equiv \mathscr{E}(\hat{\bm{\phi}}, \Breve{\bm{d}}) \coloneqq \bm{\sigma}(\hat{\bm{y}}_\phi \circ \Breve{\bm{\gamma}}_d) = \Breve{\bm{y}}_e,
\]
where $\circ$ is the Hadamard operator for element-wise multiplication, and $\bm{\sigma}$ is the softmax function to normalize the output to 0--1 like a probability.
This likelihood conjunction avoids the contradiction of inconsistent murmur shape and heart phase. For example, if $\hat{y}_\phi^\mathrm{Sys}$ is highest and $\Breve{d}^\mathrm{MVP}$ is lowest, then $\Breve{y}_e$ would be inferred as MVP; alternatively, if $\hat{y}_\phi^\mathrm{Dias}$ is lowest, then $\Breve{y}_e$ would not be inferred as MS, even if $\Breve{d}^\mathrm{MS}$ is very low. 
$\Breve{\bm{y}}_e$ computes the likelihood of each diagnosis only using heuristic criteria evaluation and not supervised learning. 
However, the best scoring item in $\Breve{\bm{y}}_e$ may not be the "best" explanation since it may be overcomplicated. We resolve this with parsimonious selection in the next stage.

\subsubsection{Abductive prediction}

When choosing among equally satisfying hypotheses, abductive reasoning would select the simplest~\cite{leake1995abduction}. 
For heart auscultation, we note that more expressive shape functions can subsume simpler ones, i.e., $\Breve{s}_m^\mathrm{MR} \subseteq \Breve{s}_m^\mathrm{AS} \subseteq \Breve{s}_m^\mathrm{MVP} \subseteq \Breve{s}_m^\mathrm{MS}$, e.g., Fig. \ref{fig:demo-mvp}d shows MS overfitting for MVP. 
We enforce simplicity by regularizing hypotheses by their complexity. However, unlike regularization in machine learning that tune weight hyperparameters based on validation datasets, we predict the regularization weight using supervised learning on each observation. The methods are not equivalent since the former is inductive inference on a dataset and the latter abductive inference on an instance.

We define parameters $\hat{\bm{\lambda}}_r$ for regularization penalties for all hypotheses, where a larger value represents a higher penalty for more complex murmur shapes, specifically, $1 \le \hat{\lambda}_r^\mathrm{MR} \le \hat{\lambda}_r^\mathrm{AS} \le \hat{\lambda}_r^\mathrm{MVP} \le \hat{\lambda}_r^\mathrm{MS}$. 
We predict $\hat{\bm{\lambda}}_r$ using a feedforward layer with the perceptual diagnosis $\hat{\bm{y}}_0$ as input, so they depend on the observed murmur.
On average, $\hat{\lambda}_r^\mathrm{MR} = 1.6$, $\hat{\lambda}_r^\mathrm{AS} = 14.3$, $\hat{\lambda}_r^\mathrm{MVP} = \hat{\lambda}_r^\mathrm{MS} = 33.7$. Notice how $\hat{\lambda}_r^\mathrm{MVP} = \hat{\lambda}_r^\mathrm{MS}$ because they apply at different heart phases, obviating the need to distinguish between the shapes. 
Hence, we resolve a parsimonious abductive prediction by dividing the evaluated prediction $\Breve{\bm{y}}_e$ with the penalties and normalizing, i.e.,
\[
\hat{\bm{y}}_r = \mathscr{R}(\Breve{\bm{y}}_e) \coloneqq \bm{\sigma} \left( \Breve{\bm{y}}_e \circ \hat{\bm{\lambda}}_r^{\circ -1} \right).
\]
While this involves supervised learning to minimize the categorical cross entropy loss $L(\bm{y}, \hat{\bm{y}}_r)$, it is only to estimate the monotonic regularization penalty for each instance, rather than fully training a black box classifier.

\subsubsection{Ensemble prediction}

Conforming to the hypotheses evaluation, abductive prediction $\hat{\bm{y}}_r$ trades-off performance for interpretability. 
To improve performance, we combine $\hat{\bm{y}}_r$ with perceptual prediction $\hat{\bm{y}}_0$ in a weighted sum,
\[
\hat{\bm{y}} = \mathscr{W}(\hat{\bm{y}}_r, \Breve{\bm{y}}_e) \coloneqq \bm{\ell_1} \left( (1 - \hat{\bm{\rho}}_w^\top) \hat{\bm{y}}_0 + \hat{\bm{\rho}}_w^\top \hat{\bm{y}}_r \right),
\]
where $\hat{\bm{\rho}}_w$ is a vector of weight fractions for each hypothesis predicted using a feedforward layer with the perceptual diagnosis $\hat{\bm{y}}_0$ as input, and $\bm{\ell_1}$ is the L1 norm to min-max normalize across classes. On average, $\hat{\bm{\rho}}_w = \left(.78, .80, .79, .79, .79 \right)^\top$, indicating that 79\% of the prediction is due to abduction $\hat{\bm{y}}_r$ instead of perception $\hat{\bm{y}}_0$.

In summary, the multi-stage technical approach (Stages 1-7) follows the steps of selective abduction process:
\begin{enumerate}
    \item[I.] \textit{Observe event} 
    by audio displacement to interpret its amplitude (Stage 1), and segment the murmur location (2). 
    \item[II.] \textit{Hypothesize} 
    by enumerating diagnoses, retrieving corresponding murmur shape functions (3), and perceiving an initial diagnosis from the hypotheses (4).
    \item[III.] \textit{Evaluate plausibility} 
    by fitting all murmur shapes to the observation (5), and
    deductively evaluating coherence criteria murmur shape goodness-of-fit and murmur heart phase mapping (6).
    \item[IV.] \textit{Resolve explanation} by penalizing complex murmur shapes to select the simplest hypothesis (7).
\end{enumerate}
The abductive prediction from Step IV (Stage 7) is then combined with the perceptual prediction from Step II (Stage 4) to make a final combined ensemble prediction (Stage 8).
See example diagram explanations in Figs. \ref{fig:demo-explanations},  \ref{fig:demo-mvp},  \ref{fig:demo-as}.

\subsection{Related work on abductive AI interference}

There is a long history of abductive reasoning in AI, 
starting with Pople~\cite{pople1973mechanization}, 
followed by much work in the 1980s and 1990s (e.g.,~\cite{josephson1996abductive, mooney2000integrating}). 
Works on intelligent tutoring using abductive reasoning to align with human understanding (e.g.,~\cite{fraser1989errors, makatchev2004abductive, makatchev2004b_abductive}) share similar objectives as our work. 
For medical diagnosis, abductive framing has been used to abstract features from time series data~\cite{teijeiro2018abductive, teijeiro2016heartbeat}.
While abduction has been studied in formal methods of machine reasoning for AI~\cite{bottou2014machine}, its use in machine learning and deep neural networks (DNN) is limited. 
Causal networks model a graph of causal chains from causes to evidence~\cite{pearl2009causality} and can perform abductive inference by calculating the posterior probabilities of causes given the evidence. 
Knowledge-base reasoning, case base reasoning and predicate logic systems model deductive rules from causes to effects, and can perform abductive reasoning by backward chaining along the rules~\cite{leake1995abduction, kakas1992abductive}, though predicate rules and premise propositions are manually specified~\cite{kakas1992abductive}. 
Recent work has also sought to support probabilistic abductive explanations~\cite{izza2023computing}.
Since simpler explanations are more plausible, some methods enforce simplicity criteria~\cite{leake1995abduction, morvan2008simulation}. 
For phenomena with known physical equations, agent-based modeling can simulate the consequences of different hypotheses by iterating variable values within a pre-specified range in the hypothesis space, and select the best hypothesis that fits the evidence based on cost-based criteria~\cite{morvan2008simulation}. 
While these formal methods support explainable AI by implementing abductive reasoning, they are less capable of predicting on unstructured data, such as images and audio, which is commonly modeled with DNNs.
   
Recently, some DNNs have implemented abductive reasoning by adding abductive modules. 
Dai et al. appended a DNN with an abductive logic programming (ALP) component to finetune predictions with pseudo-labels~\cite{dai2019bridging}. The ALP component is solved separately, thus the model cannot be trained end-to-end. 
For video understanding, after manual annotation of event labels and causal relationships, DNNs have been trained to perform abductive reasoning with event sequence pairs from event graphs~\cite{du2021learning, li2023multi, liang2022visual}, thus learning causal networks incrementally in an end-to-end manner. 
These methods reason over homogeneous categorical entities (events), while our approach handles heterogeneous concepts; our hypotheses are diagnosis labels and evidence include parametric time series shape functions, so they cannot be fully represented only as a causal graph. 
Although the predicted event causes are likely true, the generalized rules from causes to consequents are not transparent to users. 
Furthermore, without knowledge constraints, the general hypothesis rules learned are not transparent to users, and may be spurious and misaligned with domain experts. This is particularly dangerous for clinical applications. 
Finallly, though common in abductive reasoning systems, hand-coding the hypotheses as in our approach limits its scalability. This could be mitigated by implementing neuro-symbolic AI methods that can more simply encode domain-specific or user-driven logical rules in neural networks~\cite{riegel2020logical}.

\section{Evaluations}

We evaluated DiagramNet in multiple stages:
1) a demonstration study showing the interpretability of domain-aligned explanations;
2) a quantitative modeling study comparing DiagramNet against baseline CNN models and other reasonable approaches; and
3) a qualitative user study with medical students investigating the usefulness of domain-aligned explanations compared to more common, but overly-technical saliency map explanations.

\subsection{Demonstration study}
We demonstrate the usefulness of DiagramNet for abductive and supplemental explanations types~\cite{lim2009assessing, liao2020questioning, lim2011design}.

\begin{figure*}[t!]
    \centering
    \includegraphics[width=9.2cm]{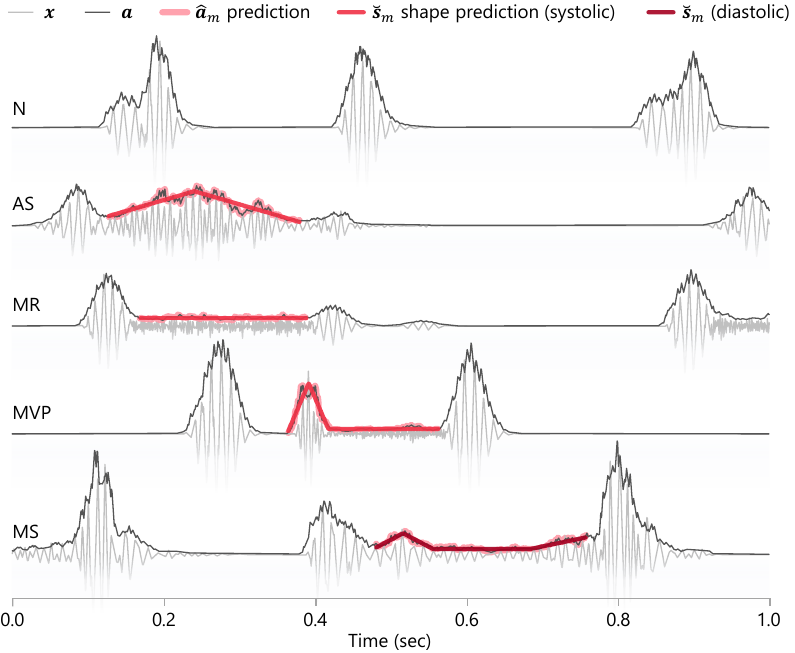}
    \vspace*{-0.25cm}
    \caption{
    Murmur diagrams of abductive explanations for phonocardiograms (PCGs) of different cardiac diagnosis predictions, showing the best-fitting murmur shapes and murmur heart phase (systolic in red or diastolic in dark red).
    We provide interactive demos to explore shape functions for: 
    {\color[HTML]{0060df}\underline{\href{http://www.desmos.com/calculator/gjvrllldr0}{AS}}}, 
    {\color[HTML]{0060df}\underline{\href{http://www.desmos.com/calculator/jmdhs9gsci}{MR}}}, 
    {\color[HTML]{0060df}\underline{\href{http://www.desmos.com/calculator/raaap9gyqb}{MVP}}}, and 
    {\color[HTML]{0060df}\underline{\href{http://www.desmos.com/calculator/d559nqxzng}{MS}}}.
    }
    \Description{
    This figure displays 5 line graphs in vertical order of phonocardiogram waveforms for different cardiac conditions: Normal (N), Aortic Stenosis (AS), Mitral Regurgitation (MR), Mitral Valve Prolapse (MVP), and Mitral Stenosis (MS). Each waveform is plotted over time (0 to 1.0 seconds). Highlighted areas indicate predicted murmur shapes and phases, with systolic murmurs in red and diastolic murmurs in dark red. These highlights align murmurs with heart phases, aiding in diagnosis prediction based on murmur characteristics within the cardiac cycle.
    }
    \label{fig:demo-explanations}
\end{figure*}

\begin{figure*}[t!]
    \includegraphics[width=12.2cm]{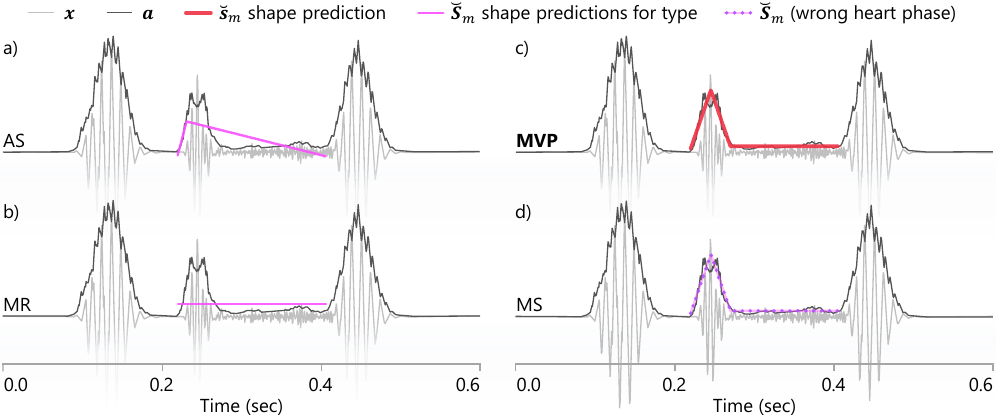}
    \vspace*{-0.25cm}
    \caption{
    Murmur diagrams of contrastive explanations for a PCG with MVP prediction, showing alternative hypothesized murmur shapes.
    Murmur shapes for AS and MR do not fit the murmur, but both MVP and MS shapes do,
    because the MS function overfits to MVP data.
    The MS murmur should be during the diastolic heart phase, not systolic, so this was not inferred.
    }
    \Description{
    The figure shows four line graphs, labeled (a) to (d), comparing hypothesized murmur shapes for different heart conditions against a phonocardiogram (PCG) with mitral valve prolapse (MVP) diagnosis. Each graph displays two superimposed waveforms over time in seconds: one representing the actual PCG data and the other representing a shape prediction for a specific condition. The x-axis ranges from 0.0 to 0.6 seconds.
    Graphs (a) and (b) show murmur shapes (purple lines) for Aortic Stenosis (AS) and Mitral Regurgitation (MR), respectively, which do not match the MVP murmur shape. Graphs (c) and (d) display murmur shapes (red, and purple-dashed lines, respectively) for MVP and Mitral Stenosis (MS), respectively, where both fit the MVP data waveform; however, MS is incorrectly shown during the systolic phase instead of diastolic.
    }
    \label{fig:demo-mvp}
\end{figure*}

\begin{figure*}[t!]
    \includegraphics[width=12.2cm]{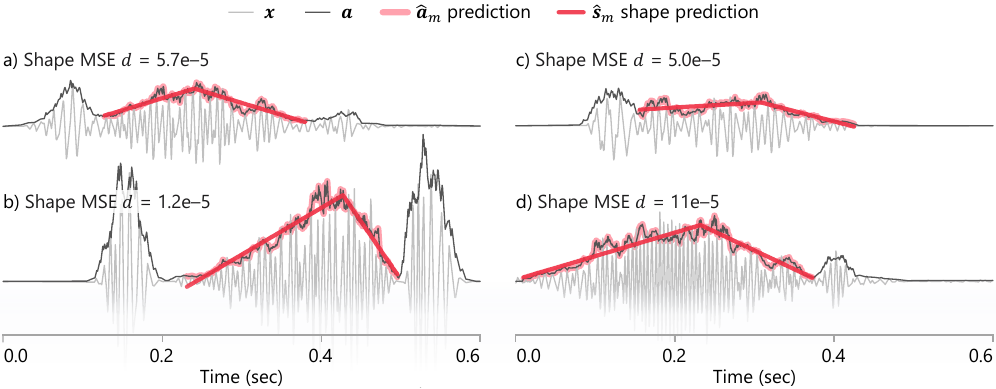}
    \vspace*{-0.25cm}
    \caption{
    Murmur diagrams of example-based explanations of similar AS cases to
    compare how a specific case looks similar to others with the same diagnosis.
    For example, (c) is missing the S2 sound, but is similar to (a) with a very weak S2 sound too.
    }
    \Description{
    The figure of line graph PCGs shows four cases (a, b, c, d) with similar aortic stenosis (AS) diagnoses. Case (a) has a very weak S2 sound, while case (c) is missing the S2 sound entirely. Case (b) serves as a reference with a normal S2 sound. Case (d) also has a weak S2 sound and is missing the S1 sound. The murmur regions are highlighted in red, and the predicted murmur shape drawn with a red line each. The Shape MSE d values are 5.7e-5 for case (a), 1.2e-5 for case (b), 5.0e-5 for case (c), and 11e-5 for case (d).
    }
    \label{fig:demo-as}
\end{figure*}

\begin{enumerate}
    \item[a)] \textit{Abductive explanations} to select the best-fitting hypotheses for each prediction.
    Fig. \ref{fig:demo-explanations} shows the best explanation for instances of each diagnosis type.
    Users can see the predicted murmur segment by the coverage of the red shape, and how the shape fits the amplitude time series optimally.
    
    \item[b)] \textit{Contrastive explanations}~\cite{miller2019explanation} to describe the evidence for alternative model outcomes.
    Fig. \ref{fig:demo-mvp} shows how alternative hypotheses for a case with MVP were not selected due to poor murmur shape fit or wrong heart phase.
    See Appendix Table \ref{table:demo-mvp-predictions} for shape parameters and fit MSE.
    
    \item[c)] \textit{Counterfactual explanations}~\cite{wachter2017counterfactual,cai2019human,zhang2022towards} to propose feature changes to predict another outcome.
    These can be derived from the contrastive explanations to show how the murmur amplitude could be slightly different to be predicted as due to another diagnosis. 
    See Appendix Fig. \ref{fig:demo-mvp-counterfactual} for examples.
    
    \item[d)] \textit{Example-based explanations} to compare similar or counterfactual examples~\cite{cai2019effects}.
    Fig. \ref{fig:demo-as} demonstrates several AS cases with different crescendo-decrescendo murmurs. 
\end{enumerate}

\subsection{Modeling study}\label{sec:modeling-study}

Since DiagramNet $M$ conforms to domain hypotheses, 
we expect it to have better prediction performance and explanation faithfulness than other non-knowledge-based models.
We describe the models compared, evaluation metrics, and results of our modeling study.

\subsubsection{Comparison models}
We compared DiagramNet against simpler models with a subset of its modules ($M_0$, $M_m$) for an ablation study, and alternative models that predict cardiac diagnosis and murmur shapes.
Architectures described in Appendix \ref{subsubsection:comparison-models}.
\begin{enumerate}
    \item[1)] $M_0(\bm{x},\bm{a}) = \hat{\bm{y}}_0$, base CNN model trained on displacement $\bm{x}$ and amplitude $\bm{a}$ to predict diagnosis.
    \item[2)] $M_\nu(\bm{\nu}) = \hat{\bm{y}}_0$, base CNN model trained on spectrogram $\bm{s}$ to predict diagnosis. This baseline is used in our user study.
    \item[3)] $M_{\tau}(\bm{x},\bm{a}) = (\hat{\bm{y}}_0,\hat{\bm{\tau}})$, multi-task model to predict diagnosis, and murmur segment start and end times. This is trained with supervised learning from $y$ labels and $\bm{\tau}=(\tau_1,\tau_L)$ annotations. This does not consider spatial information.
    \item[4)] $M_{\theta}(\bm{x},\bm{a}) = (\hat{\bm{y}}_0,\hat{\bm{\theta}})$, multi-task model to predict diagnosis, and murmur parameters. This is trained with $y$ labels and $\bm{\theta}=(\bm{\tau},\bm{\pi})$ annotations. This neglects spatial information.
    \item[5)] $M_{m}(\bm{x},\bm{a}) = (\hat{\bm{y}}_0,\hat{\bm{m}})$, encoder-decoder model to predict diagnosis, and murmur segment. Like $M_{\tau}$, this identifies the murmur start and end times, but by using U-Net \cite{ronneberger2015unet} to locate the murmur spatially with transpose-CNN layers.
    \item[6)] $M_{a_m}(\bm{x},\bm{a}) = (\hat{\bm{y}}_0,\hat{\bm{a}}_m)$, encoder-decoder model to predict diagnosis, and murmur amplitude. 
    This model can "see" the murmur and attempt to reconstruct it.
    Unlike diagrammatic explanations that adhere to parametric constraints, the amplitude predictions here are unconstrained.
    \item[7)] $M(\bm{x},\bm{a}) = (\hat{\bm{y}}_0, \hat{\bm{\phi}},\Breve{\bm{s}}_m, \Breve{\bm{y}}_e,\hat{\bm{y}}_r,\hat{\bm{y}})$, DiagramNet to infer murmur heart phase $\hat{\bm{\phi}}$, murmur shapes for all hypotheses $\Breve{\bm{s}}_m$, and perceptual $\hat{\bm{y}}_0$, evaluation-based $\Breve{\bm{y}}_e$, parsimoniously-resolved abductive $\hat{\bm{y}}_r$, and final $\hat{\bm{y}}$ diagnoses.
\end{enumerate}

\subsubsection{Evaluation metrics}
We compared the models using various measures of 
\textit{prediction performance} (accuracy, sensitivity, specificity) and 
\textit{explanation faithfulness} (murmur overlap, murmur parameters estimation errors). 
These were evaluated on a dataset of 7,262 1-sec instances.
For each instance, we calculated:
\begin{itemize}
    \item \textit{Prediction correctness} ($\uparrow$ better) 
    aggregated as \textit{accuracy}, and as \textit{sensitivity} and \textit{specificity}, commonly used in medicine.
    \item \textit{Murmur segment Dice coefficient} ($\uparrow$ better) 
    measures the overlap between predicted $\hat{\bm{m}}$ and actual $\bm{m}$ murmur segments, 
    i.e., $s_\tau = 2(\bm{m} \cdot \hat{\bm{m}}) / (|\bm{m}|^2 + |\hat{\bm{m}}|^2)$. 
    For $M_\tau$ and $M_\theta$ that only predict parameters, we computed $\hat{\bm{m}} = [\tau_1 < \bm{t} < \tau_L]$. 
    \item \textit{Murmur segment parameter MSE} ($\downarrow$ better) 
    indicates how well the model predicted the start $\tau_1$ and end $\tau_L$ time parameters of the murmur, i.e., $\varepsilon_\tau = ||\tau_1-\hat{\tau}_1||_2^2 + ||\tau_L-\hat{\tau}_L||_2^2$. 
    
    \item \textit{Murmur shape parameters MSE} ($\downarrow$ better) 
    indicates how well the model predicted the murmur shape function parameters, i.e., $\varepsilon_\theta = ||\bm{\theta} - \hat{\bm{\theta}}_y||_2^2$, where $\bm{\theta}$ and $\hat{\bm{\theta}}_y$ are the actual and predicted parameters for the correct diagnosis $y$, respectively.
    \item \textit{Murmur shape fit MSE} ($\downarrow$ better) 
    indicates how well the inferred murmur shape $\Breve{\bm{s}}_m$ (or reconstructed murmur amplitude $\hat{\bm{a}}_m$) fits the ground truth murmur amplitude $\bm{a}_m$, i.e., $\varepsilon_a = ||\bm{a}_m - \Breve{\bm{s}}_m||_2^2$ (or $\varepsilon_a = ||\bm{a}_m - \hat{\bm{a}}_m||_2^2$).
\end{itemize}

\subsubsection{Results}
Fig. \ref{fig:results-model-performances} shows the performance of all 7 models for four evaluation metrics. See Appendix Table \ref{table:results-performance} for numeric details.
For base CNN models, predicting on the spectrogram ($M_\nu$) improved performance only very slightly over predicting on amplitude ($M_0$), suggesting that CNNs can already model frequency information with its convolution filters.
Multi-task models ($M_\tau$, $M_\theta$) sacrificed diagnosis prediction accuracy to predict murmur parameters, yet still had high errors, and very inaccurate segment predictions.
Encoder-decoder models ($M_m$, $M_{a_m}$) performed better by more accurately predicting diagnoses than base CNN models, could reasonably locate segment regions, and had moderately low shape parameter and fit estimation errors.
This suggests that merely treating parameters as stochastic variables to predict is less reliable than explicitly modeling spatial and geometric information.

DiagramNet was the best performing with highest accuracy for diagnosis prediction $\hat{\bm{y}}$, and lowest error for shape parameters $\tilde{\bm{\theta}}$ and fit $\Breve{\bm{s}}_m$ estimation.
Interestingly, despite murmur shape prediction $\Breve{\bm{s}}_m$ being less expressive than murmur amplitude prediction $\hat{\bm{a}}_m$, since it predicts straight lines, its fit is still better (lower MSE). 
Due to training with backprop from $\hat{\bm{m}}$ and  $\hat{\bm{y}}$, even its perceptual diagnosis $\hat{\bm{y}}_0$ was better than that of other models.
However, evaluation-based $\Breve{\bm{y}}_e$ and parsimoniously-resolved abductive $\hat{\bm{y}}_r$ predictions were weaker. 
Adhering to the rule-like conjunction for $\Breve{\bm{y}}_e$ enforces evaluation to be based on backward chaining of hypothesis rules. This supports interpretable abductive reasoning to satisfy the coherence criterion, but this trades-off accuracy. 
In contrast, $\hat{\bm{y}}_r$ significantly improved accuracy by penalizing overcomplicated murmur shapes. Furthermore, $\hat{\bm{y}}_r$ is interpretable, unlike $\hat{\bm{y}}_0$, and achieves comparable performance. 
Combining the perceptual and abductive predictions as a weighted sum ($\sim$79\% based on $\hat{\bm{y}}_r$) provided a partially interpretable ensemble prediction $\hat{\bm{y}}$ with highest accuracy; this improved both its prediction performance and interpretability~\cite{rudin2019stop}. 
Applications could report $\hat{\bm{y}}_r$ or $\hat{\bm{y}}$ depending on interpretability and performance requirements.

\begin{figure*}[t!]
    \centering
    \includegraphics[width=16.65cm]{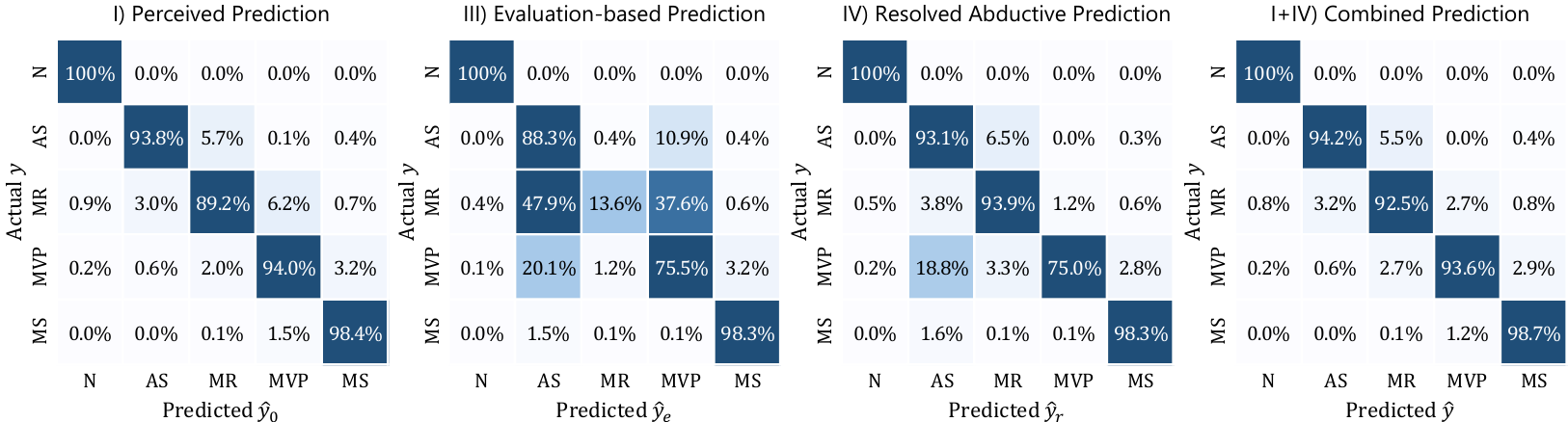}
    \vspace{-0.25cm}
    \caption{
    Confusion matrices of DiagramNet predictions at various steps based on: 
    I) spatial (temporal) perception, 
    III) rule-like evaluation of conjunctive evidence, 
    IV) parsimonious abductive resolution, and 
    I+IV) combined perceptual and abductive reasoning.
    See Fig. \ref{fig:results-confusion-matrix-cnn} in Appendix to compare against the baseline CNN model.
    }
    \Description{
    Confusion Matrices for Different Prediction Models. The figure shows confusion matrices for Perceived Prediction, Evaluation-based Prediction, Resolved Abductive Prediction, and Combined Prediction models. The matrices compare actual labels (N, AS, MR, MVP, MS) across rows with corresponding predicted labels across columns. High accuracies are observed for normal (N) and mitral stenosis (MS) across all models, while the performance varies for aortic stenosis (AS), mitral regurgitation (MR), and mitral valve prolapse (MVP). Each matrix uses a heatmap with color gradients to indicate accuracy levels. All four confusion matrices mostly show high accuracy for the matrix diagonal, with Combined Prediction being the strongest, and Evaluation-based and Resolved Abudction Predictions weakest.
    }
    \label{fig:results-confusion-matrices}
\end{figure*}

\begin{figure*}[t!]
    \centering
    \includegraphics[width=11.4cm]{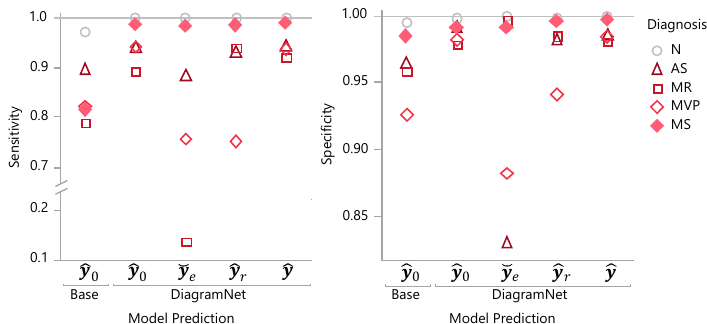}
    \vspace{-0.25cm}
    \caption{
    Clinical performance of DiagramNet compared to the base CNN model for various diagnoses.
    The perceptual $\hat{\bm{y}}_0$, abductive $\hat{\bm{y}}_r$, and final $\hat{\bm{y}}$ predictions of DiagramNet are higher than that of the baseline model $\hat{\bm{y}}_0$ for each diagnosis.
    The evaluation-based prediction $\Breve{\bm{y}}_e$ has lower accuracy for some diagnoses, due to the lower dimensionality of deducing only with shape goodness-of-fit distances instead of CNN embeddings.
    Hypothesis regularization to penalize overcomplicated hypotheses helps to improve accuracy with the resolved abductive prediction $\hat{\bm{y}}_r$.
    }
    \Description{
    Figure comparing sensitivity and specificity of different model predictions for various diagnoses.    
    The figure consists of two scatter plots. The left plot shows sensitivity values from 0.1 to 1.0, and the right plot shows specificity values from 0.85 to 1.0. Both plots compare five model predictions: Base (
    y_0), DiagramNet (y_0, y_e, y_r, y). Diagnoses are represented by symbols: circle (N), triangle (AS), square (MR), diamond (MVP), and solid-diamond (MS).
    }
    \label{fig:results-sensitivity-specificity}
\end{figure*}

Next, we examined the diagnostic performance for each cardiac disease.
Fig. \ref{fig:results-confusion-matrices} shows the confusion matrices for the four prediction stages of DiagramNet, and Appendix Fig. \ref{fig:results-confusion-matrix-cnn} shows that for the base CNN model.
Base CNN often confused different diseases, such as between MVP and MS due to their similar murmur shapes.
When predicting on evidence evaluation with $\Breve{\bm{y}}_e$, DiagramNet did confuse between systolic murmurs (AS, MR, MVP), but could accurately distinguish between MVP and MS due to their different murmur heart phases $\hat{\phi}_0$. 
The confusion between systolic murmurs in $\Breve{\bm{y}}_e$ was due to the overfitting of more complex murmur shapes (e.g., MR overfit by AS and MVP, AS overfit by MVP), but this was mitigated with the hypothesis regularization in $\hat{\bm{y}}_r$.
The combined diagnosis prediction $\hat{\bm{y}}$ ameliorated weaknesses in the perceptual and abductive predictions to produce a very clean confusion matrix.
Finally, Fig. \ref{fig:results-sensitivity-specificity} shows that DiagramNet has higher final sensitivity and specificity for all diagnoses compared to the base CNN.

\begin{figure*}[t!]
    \centering
    \includegraphics[width=11.85cm]{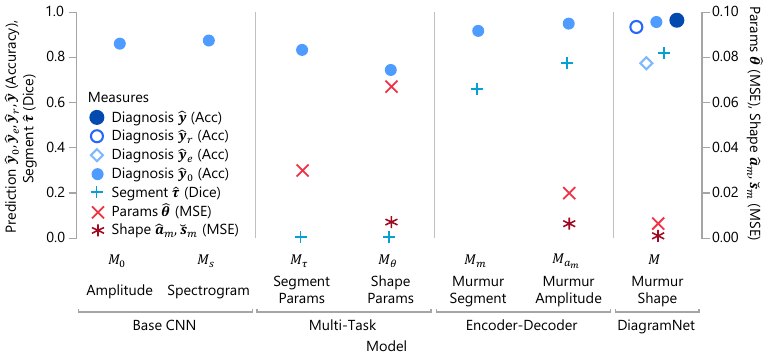}
    \vspace{-0.35cm}
    \caption{
    Results from the modeling study comparing DiagramNet with baseline and alternative models.
    Model performance is measured with perceptual $\hat{\bm{y}}_0$, rule-like evaluation-based $\Breve{\bm{y}}_e$, parsimoniously-resolved abductive $\hat{\bm{y}}_r$, and ensemble $\hat{\bm{y}}$ diagnosis accuracy.
    Explanation faithfulness is measured by murmur segment $\hat{\bm{\tau}}$ Dice coefficient, murmur shape parameters $\hat{\bm{\theta}}$, $\tilde{\bm{\theta}}$ MSE, murmur shape $\Breve{\bm{s}}_m$ fit MSE, and reconstructed murmur shape amplitude $\hat{\bm{a}}_m$ MSE.
    For blue metrics, higher is better; for red metrics, lower is better.
    DiagramNet has highest prediction and segmentation accuracy, and lowest estimation error for parameter values and shape fits (all good).
    In Appendix \ref{sec:appendix-performance-layout}, see Table \ref{table:results-performance} for numeric details, and Fig. \ref{fig:results-model-performances-expanded} for expanded layout.
    }
    \Description{
    The figure of a dot plot comparing the performance of different models on various measures. The figure displays performance metrics for models including Base CNN, Multi-Task, Encoder-Decoder, and DiagramNet on diagnosis accuracy, segment Dice score, parameter mean squared error (MSE), and shape MSE. Each model's performance is represented by symbols and colors: blue symbols for diagnosis accuracy of different models (solid dark blue for DiagramNet combined prediction, outline blue circle for DiagramNet parsimonously-resolved abductive prediction, outline blue diamond for DiagramNet evaluation-based prediction, and small light blue circle for base perceptual prediction), blue plus signs for segment Dice score, red crosses for parameter MSE, and red asterisks for shape MSE. The y-axis represents prediction accuracy and segment Dice score, while the secondary y-axis on the right represents parameter and shape MSE. The x-axis lists models: Base CNN with Amplitude and Spectrogram variants, Multi-Task model with Segment Parameters and Shape Parameters variants, Encoder-Decoder models with Murmur Segment and Murmur Amplitude segments, and DiagramNet with Murmur Shape.
    }
    \label{fig:results-model-performances}
\end{figure*}
    
\subsection{Qualitative user study}

We evaluated the usefulness of domain-aligned explanations with a qualitative user study.
We recruited medical students as domain experts, due to their training on auscultation to diagnose heart murmurs.
We did not conduct a summative evaluation due to limited recruitment.
Our key research questions were: How do clinicians ...
\begin{enumerate}
    \item[RQ1.] Diagnose heart auscultation \textit{without AI}?
    \item[RQ2.] Accept AI-based diagnosis \textit{without XAI}?
    \item[RQ3.] Interpret AI-based diagnosis \textit{with XAI} that is\\domain-aligned \textit{diagrammatic} or \textit{saliency}-based?
\end{enumerate}

\subsubsection{Experiment conditions and apparatus}

\begin{figure*}[t!]
    \centering
    \includegraphics[width=13.5cm]{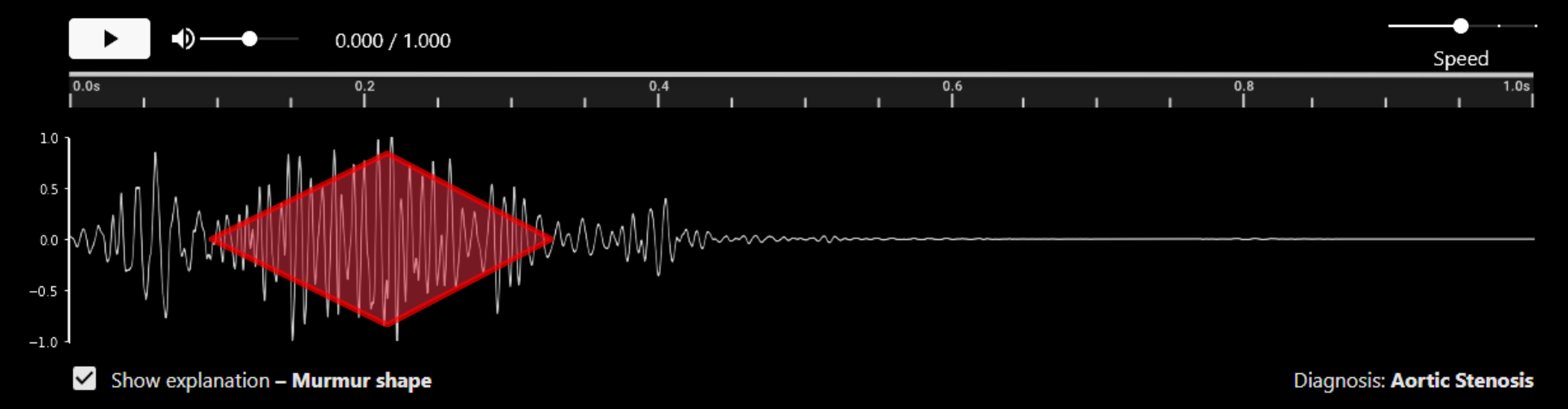}
    \vspace*{-0.20cm}
    \caption{
    User interface for the AI diagnosis system with Murmur-diagram XAI used in the user study.
    The user can play the heart sound at various volumes and speeds, and view the phonocardiogram (PCG).
    On clicking to view the explanation, the murmur diagram of the predicted diagnosis is overlaid on the PCG, showing the recognized murmur region and shape, as a shaded red area.
    In this case, the model fit a crescendo-decresendo shape in the murmur region to explain its prediction for AS.
    }
    \Description{
    Figure shows the screenshot of the user interface, showing a phonocardiogram in one row with highlighted murmur shape indicating aortic stenosis. The phonocardiogram is of a time series waveform over a 1-second interval with an amplitude range of -1.0 to 1.0. A red diamond shape highlights the murmur, peaking around 0.2 seconds and tapering off by 0.6 seconds. The diagnosis is indicated as aortic stenosis. The user interface includes a play button, volume slider, speed slider, and checkbox to show or hide the explanation.
    }
    \label{fig:userstudy-shape}
\end{figure*}
    
\begin{figure*}[t!]
    \centering
    \includegraphics[width=13.5cm]{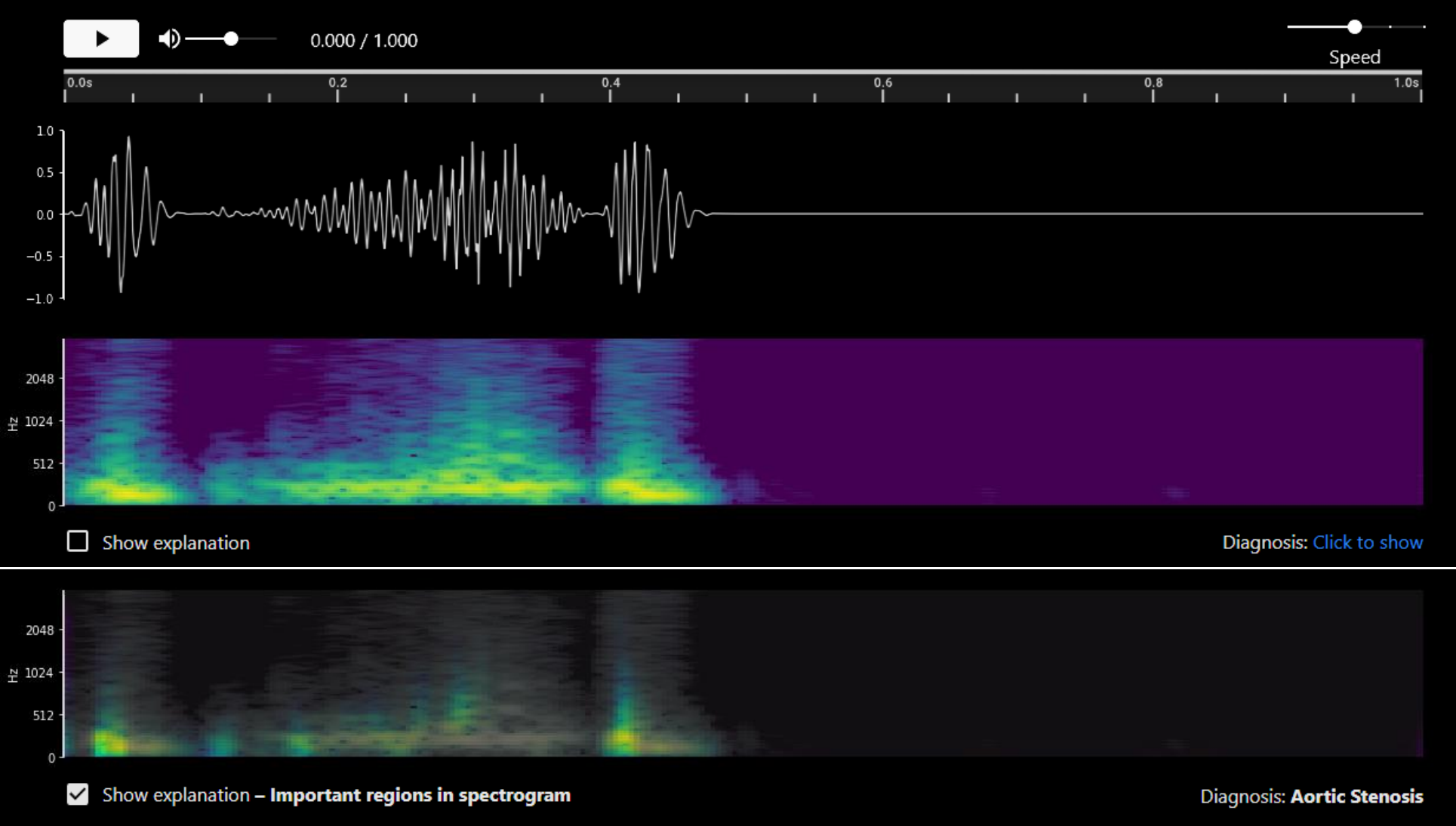}
    \vspace*{-0.20cm}
    \caption{
    User interface with Spectrogram-saliency XAI used in the user study.
    One can view the PCG (top) and spectrogram (middle) of the heart sound.
    For the spectrogram, we used the Viridis color map, where yellow-green indicates higher amplitude for the frequency at the time shown, and dark purple indicates lower amplitude.
    After initial diagnosis, one can click to view the predicted diagnosis, and click to view the explanation.
    For this UI, the explanation is a saliency map showing the important regions in the spectrogram (bottom).
    More important regions are left colored, while less important ones are more transparent.
    In this case, the model predicts that the diagnosis is AS, because the low frequencies during S1 and S2 were most important, followed by sporadic time regions in the murmur and one region near the apex with higher frequencies.
    }
    \Description{
    Figure shows two screenshots of the user interface, with the first one showing a phonocardiogram of the time series waveform over a 1-second interval in top row and a spectrogram of frequency over time in the bottom row. The spectrogram shows brighter pixel colors for louder frequencies at each time. That spectrogram is the base without showing explanations. The second screenshot shows just the spectrogram with the saliency map explanation, such that less important pixels are transparent (appearing dark on the black background), and important pixels more visible. Salient regions include the S1 lub, S2 dub, and systolic murmur times, particularly at moments with higher audio amplitude.
    }
    \label{fig:userstudy-spectrogram}
\end{figure*}
    
\begin{figure*}[t!]
    \centering
    \includegraphics[width=13.5cm]{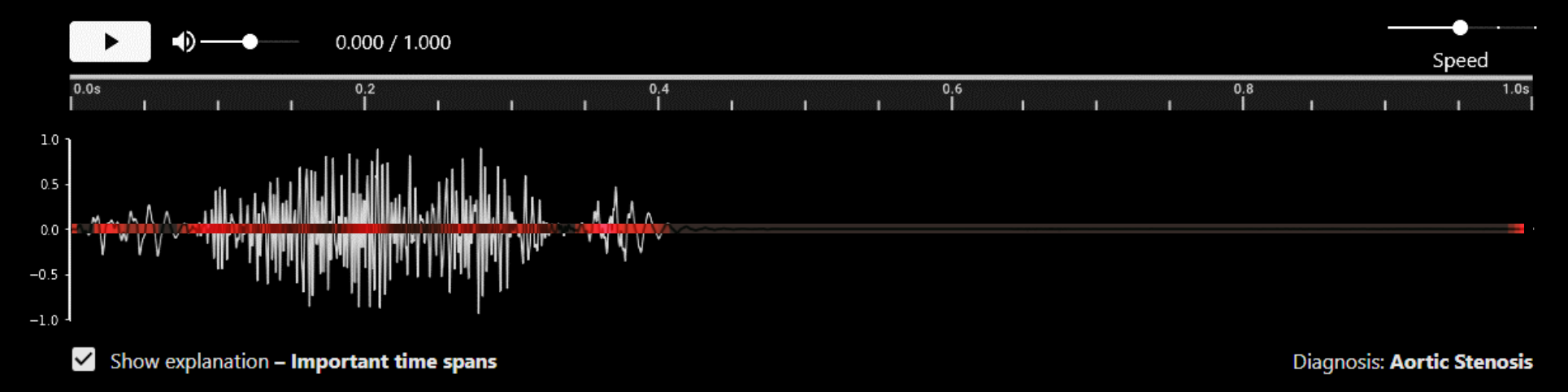}
    \vspace*{-0.20cm}
    \caption{
    User interface with Time-saliency XAI used in the user study.
    On clicking to view the explanation, a 1D saliency map is shown to indicate important time regions for the prediction. Redder indicates more important.
    In this case, the model thinks that the start of S1, S2, and some sporadic regions in the murmur (perhaps, including the apex) were important for predicting aortic stenosis (AS).
    Strangely, the model also thinks that the end of the PCG is important.
    }
    \Description{
    Figure shows a screenshot of the user interface, showing a phonocardiogram in one row. The phonocardiogram is of a time series waveform overlaid with the Time-saliency map shown as a short horizontal bar that spans the time from 0 to 1 second, indicating red for important times, and transparent for less important.
    }
    \label{fig:userstudy-time}
\end{figure*}

Participants used different user interfaces (UI) of our cardiac diagnosis system based on XAI condition.
All UI were implemented on a black background to increase the visibility of the saliency maps.
In addition to the non-explainable baseline, there were three explainable UI variants:
\begin{enumerate}
    \item[0)] \textit{Baseline} 
    to play the heart sound audio and view the phonocardiogram (PCG). After providing an initial diagnosis, the participant can click to reveal the AI's predicted diagnosis.
    \item[1)] \textit{Murmur-diagram XAI}
    explaining diagnosis prediction by overlaying a red murmur shape on the predicted murmur region (Fig. \ref{fig:userstudy-shape}).
    This aligns with clinical training, so we expect it to be the most useful and trusted explanation type.
    \item[2)] \textit{Spectrogram-saliency XAI} 
    showing the PCG and spectrogram with a saliency map overlaid as a transparency mask (Fig. \ref{fig:userstudy-spectrogram}). 
    We expect saliency maps to be less trusted due to non-use in clinical practice, spectrograms being overly-technical, and saliency maps being potentially spurious.
    \item[3)] \textit{Time-saliency XAI} 
    showing the PCG with a 1D saliency map overlaid along time (Fig. \ref{fig:userstudy-time}). 
    This simpler saliency map does not require users to know of spectrograms. 
\end{enumerate}

\subsubsection{Experiment method and procedure}
Several cases are presented to each participant, where we observed how he/she interacted with the UI and described his/her thoughts using the think aloud protocol, and performed a structured interview.
For simplicity, we only included cases where the AI made correct diagnoses, and measured trust of the AI. 
We verified that participants had decent headphones to carefully hear the heart sounds.
We obtained ethics approval from our institution before commencing the study.
For each participant, the procedure was:
\begin{enumerate}
    \item[1)] \textit{Introduction} on the experiment, and given a primer on cardiac diagnoses for N, AS, MVP, MR, and MS. We confirm the participant is familiar with these diagnoses.
    \item[2)] \textit{Consent} to participate and record their voice and interactions.
    \item[3)] \textit{Tutorial} on the UI variants including how to interpret their explanations.
    Since spectrograms are rather technical, we took care to teach how to interpret them, check for understanding later during think aloud, and clarify as needed.
    \item[4)] Three UI sessions, each randomly assigned to an XAI type, for up to two patient case trials, where
    \begin{itemize}
        \item[$\circ$] \textit{Case} is randomly chosen with a specific diagnosis (N, AS, MVP, MR, or MS).
        As clinicians also use patient information (e.g., age, symptoms, medical history) when making diagnoses, we provide it on request.
    \end{itemize}
    \begin{enumerate}
        \item[i)] \textit{Initial diagnosis} is elicited to learn their decision and rationale based on the PCG. Participants using Spectrogram-saliency also see the spectrogram at this stage.
        \item[ii)] \textit{AI diagnosis} is revealed, and the participant is asked whether he/she agreed or disagreed, and why.
        \item[iii)] \textit{XAI explanation} is shown by condition. The participant interprets the explanation, describes helpful and unhelpful aspects, and provides suggestions for improvement.
    \end{enumerate}
    \item[5)] \textit{Ranking} of XAI types by convincingness and explain why.
    \item[6)] \textit{Debrief and conclusion.} We thank the participant for their time and feedback, and conclude with compensation.
\end{enumerate}

\subsection{User study findings}
We recruited 7 medical students via snowball sampling. 
They were 5 females, 2 males, ages 20-23, and
in year 4 or 5 of their MBBS undergraduate degree.
None had prior experience with XAI.
It was difficult to recruit more due to their busy schedules.
The study took 30 minutes, and participation compensated with a \$10 gift card.

Participants completed 40 cases collectively.
They were good at diagnosing independently (80\% correct),
mostly agreed with the AI (90\% agreement) before seeing explanations (i.e., independent of XAI method).
They trusted murmur diagram explanations more than saliency explanations (see Table \ref{table:user-findings}).
Note that the results are only formative and not conclusive due to the small sample size and high variance.
Since participants were thinking aloud and discussing with the experimenter, it was not meaningful to evaluate task times.
We thematically analyzed usage and utterances, and identified several key themes with regards to XAI usage.

\subsubsection{Diagrammatic explanations were most domain-aligned, helpful and trusted.}
After diagnosing multiple cases across all XAI types, all but one participant ranked Murmur-diagram XAI to be the most convincing.
Commenting on an AS case, P2 mentioned that \textit{``when I see crescendo-decrescendo, I trust the AI diagnosis is correct, compared to if the [Time-saliency] interface shows straight line ... I can understand this without having to think ... When I see this [shape] outline given by the AI, then I realised that the [amplitude] waveform does depict crescendo-decrescendo better, but before I saw the red thing I would have looked at this and see that the volume is very level, I blocked out the fact that this is an up-slope/down-slope but I now clearly see its there.''}
P4 noticed the that \textit{``there is a linear murmur [between S1 and S2], which makes it pansystolic''} and agreed that the explanation \textit{``aligns with my understanding''}.
P6 remarked how the murmur shapes \textit{``aligns to what I learnt in school''}.
The exception was P3, who felt that \textit{``systolic murmurs are better differentiated by the spectrograms''}, and that some shape-based explanations \textit{``did not really conform to usual shapes''}, referring to an MS case where he \textit{``was expecting more of a rectangular [shape]''} but saw a decrescendo-uniform-crescendo shape instead.

\subsubsection{Unfamiliar and unconventional explanations were less interpretable and trustworthy}
Participants were unfamiliar with spectrograms initially, though many eventually understood them. 
P1 \textit{``didn't really look at the frequency much''} and felt that \textit{``it kind of is different from what we will use in real life''}.
P5 \textit{``personally am not used to a spectrogram... so kind of... no I don't think it makes me trust the AI because I don't understand it myself''}.
Conversely, P7 \textit{``agreed with [the XAI], lower pitch meaning that it's the sounds of regular valve closure, and murmur is higher pitch than regular S1 and S2''}. She felt the \textit{``underlying theory of the spectrogram explanation makes sense to me, but [I've] never seen or learnt the spectrogram explanation. However, quite intuitive after brief introduction''}.
Surprisingly, participants found the simpler Time-saliency XAI less unintuitive.
Noting the salient regions, P5 thought \textit{``that [its decision] was fair, since the AI pays attention to the S1, S2, and the "click".''}
However, some participants struggled to interpret it; P7 felt that \textit{``this is more confusing to me, therefore I trust the AI less ... I didn't understand what it was saying just by reading important time spans.''}

\begin{table}[t!]
\small
\centering
\caption{User study diagnosis performance and trust from viewing various explanations. 
Correctness is measured pre-XAI.
Participants most trusted diagram explanations.}
\label{table:user-findings}
\vspace{-0.20cm}
\begin{tabular}{rccc}
\toprule
                     & Murmur  & \multicolumn{2}{c}{Saliency} \\ \cline{3-4} 
\multicolumn{1}{l}{} & Diagram & Spectrogram      & Time      \\ \hline
Total trials         & 14      & 13               & 13        \\ \hline
Pre-AI correct       & 8       & 11               & 12        \\
Post-AI correct      & 11      & 11               & 13        \\ \hline
Post-XAI trust of AI             & \textbf{12}      & 5                & 8         \\ 
\bottomrule
\end{tabular}
\vspace{-0.25cm}
\end{table}

\subsubsection{Benefits of rich technical XAI with relatable contextualization}
Some participants wanted detailed explanations.
P5 thought that \textit{``the spectrogram makes me trust the AI more [than Time-saliency]; it is really very intricate in differentiating between high or low pitch ... spectrogram can be quite unconventional but it has the most answer out of all, most details out of all 3 of them, but the extra details are useful.''}
P4 liked the experimenter-provided \textit{``verbal explanation of spectrograms, since the more information can be obtained to increase the user's interpretation ability... if someone can give a clearer interpretation, the frequency information will be more valuable than the time-span one.''} 
Hence, explanations need to be relatable~\cite{zhang2022towards}.

\subsubsection{Need for supplementary information}
Some participants used information beyond what the model processed.
P2 remarked that, with a real patient, \textit{``I would be confirming [the diagnosis] based on the position of stethoscope, anti-apex, is the apex the loudest part I'm hearing this sound. That's the first thing, time with pulse check if really pan-systolic, [since] other murmurs can be pan-diastolic.''}
On examining an MS case, P3 diagnosed it as either AS or MS and wanted a differential diagnosis (with additional tests or observations) to eliminate which would be less likely. 

\subsubsection{Limitations and future work}
We did not collect diagnostic decisions post-XAI due to the potential confounder of participants just copying the AI prediction due to overtrust, i.e., automation bias~\cite{dzindolet2003role}. Hence, it is unclear if participants may change their decisions similarly with murmur diagrams or saliency maps regardless of trust.
We had limited our user trials to test AI predictions that were correct to avoid participants speculating messily on spurious AI explanations. Had the AI made incorrect predictions, perhaps the domain-relevant explanations may make the error more obvious and harm user trust on the AI~\cite{lim2011investigating}, compared to the less familiar explanations. Future work could study these confounders.

Our user study evaluated domain-aligned explanations against popular saliency maps~\cite{dissanayake2020robust, ren2022deep}.
While we compared segment-based ($M_m$) vs. shape-based ($M$) models in our modeling study (Section \ref{sec:modeling-study}), we did not compare their explanations in the user study. Future work could more finely evaluate the interpretability gap of users making sense of murmur segments on their own.
The weaknesses of saliency maps~\cite{boggust2022shared} make them a weak baseline, despite their popularity.
Instead, concept-based explanations~\cite{kim2018interpretability, koh2020concept, zhang2022towards} (e.g., verbally explaining "crescendo-decrescendo murmur" for "AS") could be an interesting baseline.
This would investigate the usefulness of diagrams with domain constraints on data, rather than just providing associated terms. 
Furthermore, we did not conduct a controlled study to specifically evaluate various aspects of domain-aligned XAI: abductive/deductive, domain-specific/domain-free, annotative/schematic.
Hence, future work could investigate these specific variables to determine which aspect is more essential for interpretable XAI.
Finally, we had evaluated with clinical experts with domain knowledge to read murmur diagrams. 
Yet, the domain-aligned explanations could be useful to train lay users to interpret and trust the AI. Future work can investigate this to help with the adoption of AI-based remote auscultation for patients~\cite{lee2022fully}.

\section{Discussion}
We discuss the scope, generalization and contextualization of domain-aligned XAI.

\subsection{Scope of domain-aligned XAI}

\subsubsection{Supporting human cognition and user domain knowledge}
Despite significant innovations, XAI techniques neglect user domain knowledge, leaving an interpretability gap.
This goes beyond supporting human-centric XAI at the cognitive level by tailoring explanations to support specific reasoning processes~\cite{wang2019designing, matsuyama2023iris, lyu2023if}, cognitive load limitations~\cite{abdul2020cogam, bo2024incremental, lage2019evaluation}, uncertainty aversion~\cite{wang2021show}, preferences~\cite{lage2018human, ross2017right, erion2021improving}, or relatability~\cite{zhang2022towards}.
This goes beyond social factors~\cite{ehsan2021expanding, liao2020questioning, veale2018fairness}, or fitting contextual situations~\cite{lim2009assessing}.
\textit{Domain-aligned XAI} supports user reasoning along domain norms by explicitly encoding and representing domain ontology and conventions for interpretation and evaluation.
While our implementation for auscultation-based cardiac diagnosis combined the requirements of diagrammatization, abduction and ante-hoc interpretability, other applications do not need to tightly couple them, e.g., if the domain is simple enough to not need complex diagrams, or users do not typically perform abductive reasoning.

\subsubsection{Domain-aligned XAI for high-stakes, complex domains}
Developing concrete explanations for complex domains requires significant effort in formulation and evaluation.
It requires deep involvement with domain experts to inform about the domain ontology and conventions, and hypothesis evaluation criteria (see Section \ref{subsection:discussion-generalized-approach} for steps).
We had studied one application---heart auscultation;
future work can validate on other domains (see Section \ref{subsection:discussion-generalizing-other-apps}).
However, we do not recommend all requirements for simple domains, such as bird classification or music recommendation, since formalizing the domain ontology is too costly compared to making decision errors, diagrams are not typical in the domain, or users are not domain experts to require domain-specific explanations.

\subsubsection{Abductive explanations for confirmatory analysis}
Providing abductive explanations assumes users want to generate and evaluate hypotheses, and requires hypotheses to be mathematically formulated with defined evaluation criteria.
This is unsuitable for 
1) \textit{exploratory analysis} without hypotheses, such as when data scientists debug models by looking for spurious effects rather than explicitly hypothesizing bugs.
Open-ended representations like feature attribution or saliency map would be more suitable here.
It is also unsuitable for
2) \textit{unbounded representations} like natural language explanations that acquire open-ended text from people without expectations on a finite set of explanations, so the number of hypotheses may be unbounded.
However, categorizing text responses into a discrete taxonomy would simplify identifying key hypotheses, and this could be suitable for abductive explanations.

\subsubsection{Limitations of modeling and evaluating domain-aligned explanations with domain experts}
We had conducted a small qualitative study due to challenges in recruiting domain experts. This is a perennial challenge when recruiting busy professionals with rare expertise, and will intensify as we develop more useful, domain-relevant explanations.
Nevertheless, we identified strengths and some weaknesses in our approach; a larger, summative study would have limited value despite high cost.
To model diagrammatic XAI, we formalized murmur shapes with amplitude, but omitted other concepts such as pitch, position, and radiation.
Yet, our model performance and convincingness are already superior.
Incorporating these features is left to future work for a comprehensive solution.

\subsubsection{Risk of cognitive biases on AI over-reliance}
The increased trust of using domain-aligned explanations could lead to over-reliance in AI~\cite{kaur2020interpreting} due to various cognitive biases.
\textit{Automation bias~\cite{lyell2017automation}:} If the user trusts the model due to its human-like reasoning-aligned, ante-hoc interpretable design, the user may neglect to inspect future decisions. Evaluative AI could mitigate this to show explanations only and not predictions to let the user form their opinions first~\cite{miller2023explainable}.
\textit{Complexity bias~\cite{johnson2019simplicity}:} Since diagrammatic explanations may show more details than simple visualizations, users may consider them more credible even if the AI prediction is wrong.
\textit{Confirmation bias~\cite{oswald2004confirmation}:} The user could fixate on the evidences evaluated in the hypothesis-based explanations, neglect other latent concepts and agree with the AI explanation. Predetermined hypotheses can also hinder considering other diagnoses, but this also happens with classification models. Encouraging forward reasoning to start with observing the raw input rather than AI explanation or prediction can mitigate this~\cite{wang2019designing}.
Also, supporting users to verify intermediate reasoning steps, such as identifying wrong segmentation, can help them to identify errors and limit over-reliance.

\subsection{Generalizing domain-aligned explanations}

\begin{table*}[t!]
    \setlength{\tabcolsep}{3pt}
    \small
    \centering
    \caption{
        Generalizing applications 
        --- a) ECG diagnosis, b) stock price prediction, c) skin cancer detection, d) bank loan approval ---
        i) with various diagrams
        ii) of different data types, and
        iii) base representation to
        iv) extract the diagram representation,
        v) with various evidence,
        vi) to justify predictions.
        Image credits: 
        ECG adapted from {\color[HTML]{0060df}\underline{\href{https://commons.wikimedia.org/wiki/File:Atrial_flutter34.svg}{\smash{Atrial\_flutter34}}}} by {\color[HTML]{0060df}\underline{\href{https://commons.wikimedia.org/wiki/User:Jmh649}{\smash{James Heilman, MD}}}} under the {\color[HTML]{0060df}\underline{\href{https://creativecommons.org/licenses/by-sa/3.0/deed.en}{CC BY-SA 3.0 license}}}, 
        stock price data from {\color[HTML]{0060df}\underline{\href{https://commons.wikimedia.org/wiki/File:Gold_Price_(1968-2008).gif}{\smash{Gold Price (1968-2008)}}}} by {\color[HTML]{0060df}\underline{\href{https://commons.wikimedia.org/wiki/User:Emilfaro}{\smash{Emilfaro}}}}, 
        {\color[HTML]{0060df}\underline{\href{https://visualsonline.cancer.gov/details.cfm?imageid=9186}{\smash{Melanoma}}}} from the National Cancer Institute.
    }
    \vspace{-0.15cm}
    \label{table:discussion-alt-diagrams}
    \begin{tabular}{llcccc}
    \toprule
       &              & a) ECG cardiac diagnosis    & b) Stock price prediction & c) Skin cancer detection & d) Bank loan approval \\ 
    \midrule
    i. &
      Diagram &
      \includegraphics[width=2.8cm]{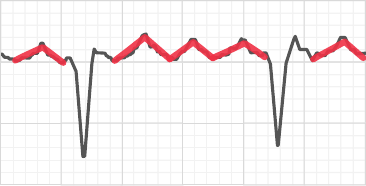} &
      \includegraphics[width=2.8cm]{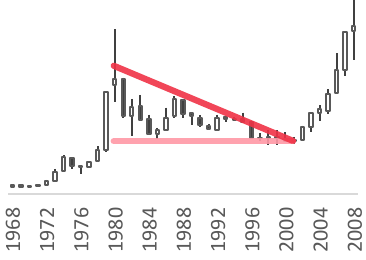} &
      \includegraphics[width=2.8cm]{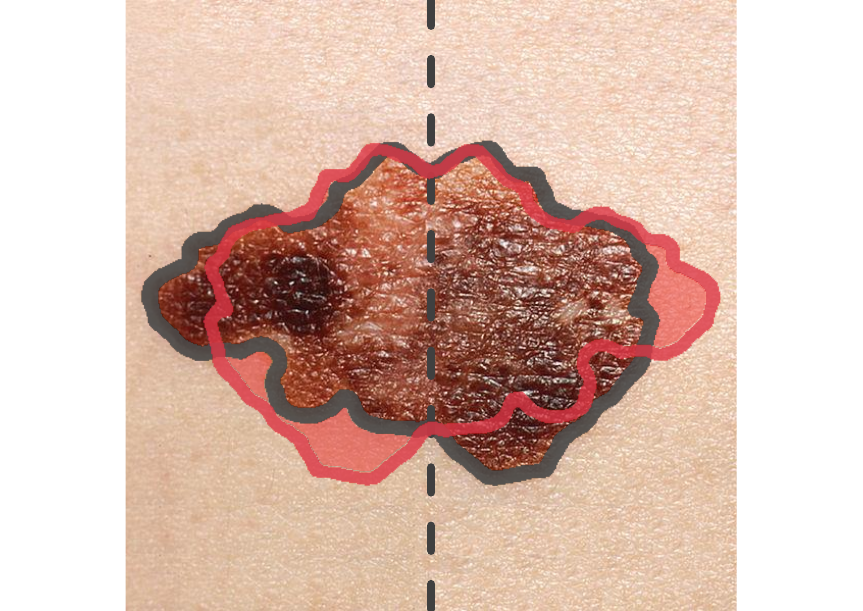} &
      \includegraphics[width=2.8cm]{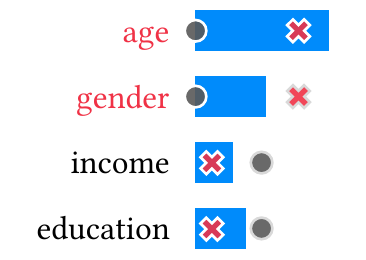} \\
    ii. & Data type    & Time series (over msec) & Time series (over years)  & Image  & Numeric vector                    \\ \addlinespace[0.13cm]
    iii. & Base representation    & Electrocardiogram (ECG) & Candlestick chart         & Photograph  & Feature attribution bars               \\
    iv. & Diagram line & ECG trace signal        & Key prices over time      & Lesion outline  & Hypothetical attribution           \\ \addlinespace[0.13cm]
    v. & Explanatory evidence  & Sawtooth wave           & Descending triangle       & Asymmetrical outline  &  {\color[HTML]{F03246}$\times$} Sensitive features used     \\
    vi. & Prediction (cause)   & Atrial flutter          & Breakdown imminent        & Malignant tumor  & Has discrimination          \\ 
    \bottomrule
    \end{tabular}
    \vspace{-0.20cm}
\end{table*}

\subsubsection{Approach to prepare domain-aligned explanations} \label{subsection:discussion-generalized-approach}
We describe the general approach for other applications:
\begin{enumerate}
    \item[1)] \textit{Diagram identification.} 
    \rev{Through user-centered design, identify domain expert stakeholder users~\cite{langer2021we}, and perform interviews, contextual inquiry~\cite{beyer1999contextual}, or stage-based participatory design~\cite{eiband2018bringing, ye2020user} to understand practices of task reasoning and explanation.
    Elicit mental models of diagrams through diagramming tasks or concept mapping~\cite{hoffman2023measures}, with diagramming tools~\cite{ma2020domain}.
    Peruse relevant documentation, trade literature, or textbooks to deepen understanding of the diagrams.}
    
    \item[2)] \textit{Diagrammatic representations.}
    Characterize the \textbf{diagram} along the diagrammatization dimensions (Section \ref{sec:diagrammatization-design-space}) to inform how to parameterize diagrammatic explanations.
    
    \item[3)] \textit{Ontology.}
    Identify interpretable \textbf{concepts} and
    model \textbf{relations} between them with heuristics to compute them (interpretable), or trained models to infer them (black box).
    
    \item[4)] \textit{Conventions.}
    Identify how each concept (e.g., murmur shape) is \textbf{manifested} in diagrams, so that they can be represented appropriately (e.g., see Section \ref{subsection:discussion-diagram-formalization}). 
    Restricting the functional form of the concepts limits spuriousness, and parameterizing them supports domain-bounded expressivity.
    
    \item[5)] \textit{Abductive reasoning.} 
    Identify alternative \textbf{hypotheses} and define \textbf{rules} relating to evidences. 
    Identify \textbf{evaluation criteria} to compare hypotheses by their fit to the observation.
    
    \item[6)] \textit{Ante-hoc interpretability.} Implement model \textbf{reasoning} using a "glassbox" approach, e.g., rules, Bayesian network, or modular neural network. For lower-stakes decisions, model-agnostic explanation could be implemented instead.
\end{enumerate}

\subsubsection{Hypothesis formalization} \label{subsection:discussion-hypothesis-formalization}
We discuss how to formalize hypotheses for abductive explanations that relate evidence concepts to decisions.
For domain-aligned XAI, these relationships should be constrained to match domain knowledge, and
need to be explicitly defined and formalized.
For decision tasks of symbolic reasoning (vs. perceptual pattern matching), this can be specified with predicate logic~\cite{kakas1992abductive} or graphical model~\cite{pearl2009causality}, appended to a neural network~\cite{dai2019bridging, liang2022visual}, encoded with neuro-symbolic methods~\cite{riegel2020logical}, or designed as modules in a neural network like with DiagramNet.

\subsubsection{Diagrammatic concept formalization} \label{subsection:discussion-diagram-formalization}
In this work, we use parametric functions to model domain concepts (murmur shapes) as line diagrams.
We performed iterative design to identify functions that align to domain ontology, what manipulations are permissible for the diagram, and what parameters could control those changes. 
Unlike standard machine learning that fits one model to a dataset, we fit one model per concept (murmur shape type).

We have investigated diagram parameterization for murmur diagrams, which are 2D, visual, line-based, from a clear line signal (amplitude).
We believe this can apply to other diagrams with low dimensions (2D, 3D), visual, clear extracted signals.
If decisions depend on many variables, it would be hard to visually conceive the high dimensions for hypothesis-driven diagramming.
Some applications require segmentation to extract lines (e.g., outline of skin lesion, facial wrinkles for action unit recognition), yet accuracy may be poor. This can lead to erroneous line fits on noisy signals. 

\subsubsection{Generalizing DiagramNet to other applications with line diagrams} \label{subsection:discussion-generalizing-other-apps}
DiagramNet is generalizable through its modules.
Steps I, III-IV of the selective abduction process can be performed automatically, while
Step II requires hand-coding of hypotheses.
We discuss adapting DiagramNet to four other applications.

\textit{Electrocardiograms (ECG)} are another clinical diagram for cardiac diagnosis.
Clinicians diagnose atrial flutter by inferring a "sawtooth" pattern (Table \ref{table:discussion-alt-diagrams}a). 
The base representation is the ECG signal $\bm{x}$, which we segment, then fit the sawtooth pattern as piecewise lines.

\textit{Candlestick charts} represent low, opening, closing, and high stock prices for each time period.
Analysts find patterns like ``broadening top'', ``descending triangle'', and ``rising wedge'' to anticipate changes~\cite{bulkowski2021encyclopedia} (Table \ref{table:discussion-alt-diagrams}b).
To explain an imminent breakdown, a ``descending triangle'' annotation could be fit to a segment $(\tau_1,\tau_L)$ with two lines $(\bm{x}_1,\bm{x}_2)$,
which we fit as 10\%-tile $x_1(t) = P(x_{\text{low}},10)$, and hypotenuse $x_2(t) = -wt + x_0$, where $w$ and $x_0$ are parameters.

\textit{Photographs of skin lesions} with ABCDE criteria are an image-diagram method to diagnose skin cancer.
For \textit{A}symmetry (Table \ref{table:discussion-alt-diagrams}c):
1) extract the lesion outline via edge detection~\cite{ziou1998edge},
2) reflect the outline across a bisecting axis, and  
3) compute the non-overlapping area $a$.
Hypothesis evaluation criterion $a > \alpha$, where $\alpha$ is a learned threshold, would indicate asymmetry for malignant diagnosis.

\textit{Bank loan approval} is a less technical application using demographic features.
Consider a case of a denied older female applicant, explained via feature importance~\cite{bach2015pixel, ribeiro2016should} (Table \ref{table:discussion-alt-diagrams}d).
To investigate discrimination, the user can hypothesize a trend line of red {\color[HTML]{F03246}$\times$} markers with higher importance for sensitive features age and gender, and an alternative hypothesis of not using sensitive features (grey {\color[HTML]{555555}$\bullet$} markers). 
Since the red trend fits better, discrimination is explained.
This demonstrates a simpler diagrammatic explanation with user-driven hypotheses, post-hoc model-agnostic explanations, an off-the-shelf visualization, on less well-defined features.

\subsection{Relation to other XAI and AI approaches}

\subsubsection{Comparing diagrams to visualization}
Current XAI visualizations use simple charts and heatmaps that may not be customary in the target domain and require contextualization. 
With diagrammatization, we encourage XAI developers to consider how domain users explain with their own conventions to develop more relevant visualizations.
\rev{Our annotative-schematic diagrammatic explanations enable users to verify explanations against the instance under inference;
visual explanations tend to be abstract or based on data-structures, requiring the user to map between the visualization and real representations.
Moreover, we showed that encoding domain constraints ante-hoc in the AI model improves model performance,
but post-hoc visualization explanations do not change the original AI prediction, and hence cannot improve the AI performance.}

\subsubsection{Comparing diagrammatization to verbal rationalization}
Instead of explaining visually with diagrams, one can also do so verbally, e.g., explaining an AS diagnosis as because
\textit{``the aortic valve leaflets are abnormally stiff due to calcification, causing the valve to have difficulty opening and closing, producing a crescendo-decrescendo murmur sound''}.
\rev{Unlike annotative diagrams, this NLG text explanation is generic (not instance-specific), has lower homomorphism (cannot be verified  physically against observation), and has unbounded expressiveness without inherent constraints (higher risk of spuriousness).
Improving its domain-alignment requires modeling knowledge bases and causal networks to relate symptoms to diagnoses.
As a \textit{rationalization}~\cite{ehsan2018rationalization}, it also makes assumptions beyond PCG observation, since there was no direct evidence of the calcification at the aortic valve, which needs an additional echocardiogram to observe.}
Since we focused on XAI for clinicians who already know how to extrapolate from disease to murmur shape, we omitted providing the low-level, concept-based, rationalization explanation in this work.
Nevertheless, future work can combine diagrams and knowledge-guided rationalization for multimodal explanations.

\subsubsection{Comparing diagrammatic abduction to maximum a posteriori (MAP) estimation} \label{subsection:discussion-max-a-posteriori}
Our approach shares some similarity to MAP, with subtle differences.
\textit{Hypothesis generation:} both abduction and MAP use predetermined hypotheses or classes.
\textit{Evaluation criteria:} abduction uses criteria to evaluate evidence toward hypotheses which can be domain-specific or subjective, whereas MAP uses a well-defined probabilistic framework.
\textit{Prior knowledge:} MAP models prior beliefs as probability distributions, while abduction incorporates tacit knowledge relationally using rules or functions.
\textit{Representation:} our approach with murmur diagrams use homomorphic murmur shapes on PCGs, while MAP represents hypotheses abstractly.
%
%
Although abductive reasoning is not necessarily probabilistic~\cite{leake1995abduction}, for future work, some stages in DiagramNet could be reimplemented using MAP to be theoretically grounded on probability and accommodate uncertainty estimation.

\subsubsection{Comparing domain-aligned explanations to feature engineering} \label{subsection:discussion-feature-engineering}
\rev{To align with domain ontology, our method ante-hoc infers interpretable features (murmur segments, shapes, and phases).}
While feature engineering can also derive domain-specific features, the features are simply input into black box models for the models to self-learn how to use the features.
\rev{Yet, these features may be spuriously leveraged, and \textit{how} they are functionally used is uncontrolled.}
Our approach with DiagramNet imposes constraints to clarify the \textit{relational structure} between concepts 
to ensure alignment with domain ontology of relations, and domain conventions of reasoning.
%
Model training is also \textit{end-to-end}, allowing co-training of multiple tasks (murmur segmentation, shape optimization, diagnosis prediction), improving performance of all tasks.

\section{Conclusion}
We presented a framework for domain-aligned XAI with diagrammatic and abductive reasoning in an ante-hoc interpretable model to reduce the interpretability gap.
Focusing on cardiac diagnosis from heart sounds, we developed DiagramNet that formalized murmur diagrams as parametric functions and diagnostic reasoning with modular stages aligned with the 4-step selective abduction process.
Demonstrations showed that it can provide abductive, contrastive, counterfactual, and example explanations.
Modeling evaluations found that DiagramNet not only had more faithful explanations but also better performance than several baseline models.
In a qualitative user study, clinicians preferred domain-aligned, diagram-based explanations over saliency map explanations on spectrograms.
This work gives insights into 
improving AI interpretability with domain-aligned XAI to richly encode domain ontology and conventions into diagrammatic, abductive explanations.

%
\begin{acks}
We thank our collaborators Drs Toon-Wei Lim and Yinghao Lim, and our participants for their clinical expertise, and
Gucheng Wang and Zijie Yang for early technical discussions.
We thank the reviewers for their insightful comments and feedback, which helped us to significantly improve the work.
This research is supported by 
the Ministry of Education, Singapore (Award No: T2EP20121-0040)
the National Research Foundation, Singapore and Infocomm Media Development Authority under its Trust Tech Funding Initiative (Award No: DTC-RGC-09), 
the NUS iHealthtech Smart Sensors \& AI Grant,
and a Google Research Scholar Award.
Joe Cahaly was supported by the MIT MISTI Program.
Any opinions, findings and conclusions or recommendations expressed in this material are those of the author(s) and do not reflect the views of 
the Ministry of Education, Singapore,
National Research Foundation, Singapore, Infocomm Media Development Authority, 
Google, and iHealthtech.
\end{acks}

\bibliographystyle{ACM-Reference-Format}
\bibliography{manuscript}


\begin{thebibliography}{152}


\ifx \showCODEN    \undefined \def \showCODEN     #1{\unskip}     \fi
\ifx \showDOI      \undefined \def \showDOI       #1{#1}\fi
\ifx \showISBNx    \undefined \def \showISBNx     #1{\unskip}     \fi
\ifx \showISBNxiii \undefined \def \showISBNxiii  #1{\unskip}     \fi
\ifx \showISSN     \undefined \def \showISSN      #1{\unskip}     \fi
\ifx \showLCCN     \undefined \def \showLCCN      #1{\unskip}     \fi
\ifx \shownote     \undefined \def \shownote      #1{#1}          \fi
\ifx \showarticletitle \undefined \def \showarticletitle #1{#1}   \fi
\ifx \showURL      \undefined \def \showURL       {\relax}        \fi
\providecommand\bibfield[2]{#2}
\providecommand\bibinfo[2]{#2}
\providecommand\natexlab[1]{#1}
\providecommand\showeprint[2][]{arXiv:#2}

\bibitem[Abdul et~al\mbox{.}(2018)]%
        {abdul2018trends}
\bibfield{author}{\bibinfo{person}{Ashraf Abdul}, \bibinfo{person}{Jo
  Vermeulen}, \bibinfo{person}{Danding Wang}, \bibinfo{person}{Brian~Y Lim},
  {and} \bibinfo{person}{Mohan Kankanhalli}.} \bibinfo{year}{2018}\natexlab{}.
\newblock \showarticletitle{Trends and trajectories for explainable,
  accountable and intelligible systems: An hci research agenda}. In
  \bibinfo{booktitle}{\emph{Proceedings of the 2018 CHI conference on human
  factors in computing systems}}. \bibinfo{pages}{1--18}.
\newblock


\bibitem[Abdul et~al\mbox{.}(2020)]%
        {abdul2020cogam}
\bibfield{author}{\bibinfo{person}{Ashraf Abdul}, \bibinfo{person}{Christian
  von~der Weth}, \bibinfo{person}{Mohan Kankanhalli}, {and}
  \bibinfo{person}{Brian~Y Lim}.} \bibinfo{year}{2020}\natexlab{}.
\newblock \showarticletitle{COGAM: measuring and moderating cognitive load in
  machine learning model explanations}. In
  \bibinfo{booktitle}{\emph{Proceedings of the 2020 CHI Conference on Human
  Factors in Computing Systems}}. \bibinfo{pages}{1--14}.
\newblock


\bibitem[Abujabal et~al\mbox{.}(2017)]%
        {abujabal2017quint}
\bibfield{author}{\bibinfo{person}{Abdalghani Abujabal},
  \bibinfo{person}{Rishiraj~Saha Roy}, \bibinfo{person}{Mohamed Yahya}, {and}
  \bibinfo{person}{Gerhard Weikum}.} \bibinfo{year}{2017}\natexlab{}.
\newblock \showarticletitle{Quint: Interpretable question answering over
  knowledge bases}. In \bibinfo{booktitle}{\emph{Proceedings of the 2017
  Conference on Empirical Methods in Natural Language Processing: System
  Demonstrations}}. \bibinfo{pages}{61--66}.
\newblock


\bibitem[Adadi and Berrada(2018)]%
        {adadi2018peeking}
\bibfield{author}{\bibinfo{person}{Amina Adadi} {and} \bibinfo{person}{Mohammed
  Berrada}.} \bibinfo{year}{2018}\natexlab{}.
\newblock \showarticletitle{Peeking inside the black-box: a survey on
  explainable artificial intelligence (XAI)}.
\newblock \bibinfo{journal}{\emph{IEEE access}}  \bibinfo{volume}{6}
  (\bibinfo{year}{2018}).
\newblock


\bibitem[Ahn and Lin(2019)]%
        {ahn2019fairsight}
\bibfield{author}{\bibinfo{person}{Yongsu Ahn} {and} \bibinfo{person}{Yu-Ru
  Lin}.} \bibinfo{year}{2019}\natexlab{}.
\newblock \showarticletitle{Fairsight: Visual analytics for fairness in
  decision making}.
\newblock \bibinfo{journal}{\emph{IEEE transactions on visualization and
  computer graphics}} \bibinfo{volume}{26}, \bibinfo{number}{1}
  (\bibinfo{year}{2019}), \bibinfo{pages}{1086--1095}.
\newblock


\bibitem[Alam et~al\mbox{.}(2010)]%
        {alam2010cardiac}
\bibfield{author}{\bibinfo{person}{Uazman Alam}, \bibinfo{person}{Omar Asghar},
  \bibinfo{person}{Sohail~Q Khan}, \bibinfo{person}{Sajad Hayat}, {and}
  \bibinfo{person}{Rayaz~A Malik}.} \bibinfo{year}{2010}\natexlab{}.
\newblock \showarticletitle{Cardiac auscultation: an essential clinical skill
  in decline}.
\newblock \bibinfo{journal}{\emph{British Journal of Cardiology}}
  \bibinfo{volume}{17}, \bibinfo{number}{1} (\bibinfo{year}{2010}),
  \bibinfo{pages}{8}.
\newblock


\bibitem[Arrieta et~al\mbox{.}(2020)]%
        {arrieta2020explainable}
\bibfield{author}{\bibinfo{person}{Alejandro~Barredo Arrieta},
  \bibinfo{person}{Natalia D{\'\i}az-Rodr{\'\i}guez}, \bibinfo{person}{Javier
  Del~Ser}, \bibinfo{person}{Adrien Bennetot}, \bibinfo{person}{Siham Tabik},
  \bibinfo{person}{Alberto Barbado}, \bibinfo{person}{Salvador Garc{\'\i}a},
  \bibinfo{person}{Sergio Gil-L{\'o}pez}, \bibinfo{person}{Daniel Molina},
  \bibinfo{person}{Richard Benjamins}, {et~al\mbox{.}}}
  \bibinfo{year}{2020}\natexlab{}.
\newblock \showarticletitle{Explainable Artificial Intelligence (XAI):
  Concepts, taxonomies, opportunities and challenges toward responsible AI}.
\newblock \bibinfo{journal}{\emph{Information Fusion}}  \bibinfo{volume}{58}
  (\bibinfo{year}{2020}), \bibinfo{pages}{82--115}.
\newblock


\bibitem[Bach et~al\mbox{.}(2015)]%
        {bach2015pixel}
\bibfield{author}{\bibinfo{person}{Sebastian Bach}, \bibinfo{person}{Alexander
  Binder}, \bibinfo{person}{Gr{\'e}goire Montavon}, \bibinfo{person}{Frederick
  Klauschen}, \bibinfo{person}{Klaus-Robert M{\"u}ller}, {and}
  \bibinfo{person}{Wojciech Samek}.} \bibinfo{year}{2015}\natexlab{}.
\newblock \showarticletitle{On pixel-wise explanations for non-linear
  classifier decisions by layer-wise relevance propagation}.
\newblock \bibinfo{journal}{\emph{PloS one}} \bibinfo{volume}{10},
  \bibinfo{number}{7} (\bibinfo{year}{2015}), \bibinfo{pages}{e0130140}.
\newblock


\bibitem[Bau et~al\mbox{.}(2017)]%
        {bau2017network}
\bibfield{author}{\bibinfo{person}{David Bau}, \bibinfo{person}{Bolei Zhou},
  \bibinfo{person}{Aditya Khosla}, \bibinfo{person}{Aude Oliva}, {and}
  \bibinfo{person}{Antonio Torralba}.} \bibinfo{year}{2017}\natexlab{}.
\newblock \showarticletitle{Network dissection: Quantifying interpretability of
  deep visual representations}. In \bibinfo{booktitle}{\emph{Proceedings of the
  IEEE conference on computer vision and pattern recognition}}.
\newblock


\bibitem[Beyer and Holtzblatt(1999)]%
        {beyer1999contextual}
\bibfield{author}{\bibinfo{person}{Hugh Beyer} {and} \bibinfo{person}{Karen
  Holtzblatt}.} \bibinfo{year}{1999}\natexlab{}.
\newblock \showarticletitle{Contextual design}.
\newblock \bibinfo{journal}{\emph{interactions}} \bibinfo{volume}{6},
  \bibinfo{number}{1} (\bibinfo{year}{1999}), \bibinfo{pages}{32--42}.
\newblock


\bibitem[Bo et~al\mbox{.}(2024)]%
        {bo2024incremental}
\bibfield{author}{\bibinfo{person}{Jessica~Y Bo}, \bibinfo{person}{Pan Hao},
  {and} \bibinfo{person}{Brian~Y Lim}.} \bibinfo{year}{2024}\natexlab{}.
\newblock \showarticletitle{Incremental XAI: Memorable Understanding of AI with
  Incremental Explanations}. In \bibinfo{booktitle}{\emph{Proceedings of the
  2024 CHI Conference on Human Factors in Computing Systems}}.
  \bibinfo{pages}{1--17}.
\newblock


\bibitem[Boggust et~al\mbox{.}(2022)]%
        {boggust2022shared}
\bibfield{author}{\bibinfo{person}{Angie Boggust}, \bibinfo{person}{Benjamin
  Hoover}, \bibinfo{person}{Arvind Satyanarayan}, {and}
  \bibinfo{person}{Hendrik Strobelt}.} \bibinfo{year}{2022}\natexlab{}.
\newblock \showarticletitle{Shared interest: Measuring human-ai alignment to
  identify recurring patterns in model behavior}. In
  \bibinfo{booktitle}{\emph{Proceedings of the 2022 CHI Conference on Human
  Factors in Computing Systems}}. \bibinfo{pages}{1--17}.
\newblock


\bibitem[Bottou(2014)]%
        {bottou2014machine}
\bibfield{author}{\bibinfo{person}{L{\'e}on Bottou}.}
  \bibinfo{year}{2014}\natexlab{}.
\newblock \showarticletitle{From machine learning to machine reasoning: An
  essay}.
\newblock \bibinfo{journal}{\emph{Machine learning}}  \bibinfo{volume}{94}
  (\bibinfo{year}{2014}), \bibinfo{pages}{133--149}.
\newblock


\bibitem[Brachman and Levesque(2004a)]%
        {brachman2004knowledge_ch13}
\bibfield{author}{\bibinfo{person}{Ronald~J. Brachman} {and}
  \bibinfo{person}{Hector~J. Levesque}.} \bibinfo{year}{2004}\natexlab{a}.
\newblock \showarticletitle{Chapter 13 - Explanation and Diagnosis}.
\newblock In \bibinfo{booktitle}{\emph{Knowledge Representation and
  Reasoning}}, \bibfield{editor}{\bibinfo{person}{Ronald~J. Brachman} {and}
  \bibinfo{person}{Hector~J. Levesque}} (Eds.). \bibinfo{publisher}{Morgan
  Kaufmann}, \bibinfo{address}{San Francisco}, \bibinfo{pages}{267--284}.
\newblock
\showISBNx{978-1-55860-932-7}


\bibitem[Brachman and Levesque(2004b)]%
        {brachman2004knowledge}
\bibfield{author}{\bibinfo{person}{Ronald~J Brachman} {and}
  \bibinfo{person}{Hector~J Levesque}.} \bibinfo{year}{2004}\natexlab{b}.
\newblock \bibinfo{booktitle}{\emph{Knowledge Representation and Reasoning}}.
\newblock \bibinfo{publisher}{Morgan Kaufman/Elsevier}.
\newblock


\bibitem[Bulkowski(2021)]%
        {bulkowski2021encyclopedia}
\bibfield{author}{\bibinfo{person}{Thomas~N Bulkowski}.}
  \bibinfo{year}{2021}\natexlab{}.
\newblock \bibinfo{booktitle}{\emph{Encyclopedia of chart patterns}}.
\newblock \bibinfo{publisher}{John Wiley \& Sons}.
\newblock


\bibitem[Cai et~al\mbox{.}(2019a)]%
        {cai2019effects}
\bibfield{author}{\bibinfo{person}{Carrie~J Cai}, \bibinfo{person}{Jonas
  Jongejan}, {and} \bibinfo{person}{Jess Holbrook}.}
  \bibinfo{year}{2019}\natexlab{a}.
\newblock \showarticletitle{The effects of example-based explanations in a
  machine learning interface}. In \bibinfo{booktitle}{\emph{Proceedings of the
  24th international conference on intelligent user interfaces}}.
  \bibinfo{pages}{258--262}.
\newblock


\bibitem[Cai et~al\mbox{.}(2019b)]%
        {cai2019human}
\bibfield{author}{\bibinfo{person}{Carrie~J Cai}, \bibinfo{person}{Emily Reif},
  \bibinfo{person}{Narayan Hegde}, \bibinfo{person}{Jason Hipp},
  \bibinfo{person}{Been Kim}, \bibinfo{person}{Daniel Smilkov},
  \bibinfo{person}{Martin Wattenberg}, \bibinfo{person}{Fernanda Viegas},
  \bibinfo{person}{Greg~S Corrado}, \bibinfo{person}{Martin~C Stumpe},
  {et~al\mbox{.}}} \bibinfo{year}{2019}\natexlab{b}.
\newblock \showarticletitle{Human-centered tools for coping with imperfect
  algorithms during medical decision-making}. In
  \bibinfo{booktitle}{\emph{Proceedings of the 2019 chi conference on human
  factors in computing systems}}. \bibinfo{pages}{1--14}.
\newblock


\bibitem[Cai et~al\mbox{.}(2019c)]%
        {cai2019hello}
\bibfield{author}{\bibinfo{person}{Carrie~J Cai}, \bibinfo{person}{Samantha
  Winter}, \bibinfo{person}{David Steiner}, \bibinfo{person}{Lauren Wilcox},
  {and} \bibinfo{person}{Michael Terry}.} \bibinfo{year}{2019}\natexlab{c}.
\newblock \showarticletitle{" Hello AI": uncovering the onboarding needs of
  medical practitioners for human-AI collaborative decision-making}.
\newblock \bibinfo{journal}{\emph{Proceedings of the ACM on Human-computer
  Interaction}} \bibinfo{volume}{3}, \bibinfo{number}{CSCW}
  (\bibinfo{year}{2019}), \bibinfo{pages}{1--24}.
\newblock


\bibitem[Card et~al\mbox{.}(1999)]%
        {card1999readings}
\bibfield{author}{\bibinfo{person}{Stuart~K Card}, \bibinfo{person}{Jock
  Mackinlay}, {and} \bibinfo{person}{Ben Shneiderman}.}
  \bibinfo{year}{1999}\natexlab{}.
\newblock \bibinfo{booktitle}{\emph{Readings in information visualization:
  using vision to think}}.
\newblock \bibinfo{publisher}{Morgan Kaufmann}.
\newblock


\bibitem[Caruana et~al\mbox{.}(2015)]%
        {caruana2015intelligible}
\bibfield{author}{\bibinfo{person}{Rich Caruana}, \bibinfo{person}{Yin Lou},
  \bibinfo{person}{Johannes Gehrke}, \bibinfo{person}{Paul Koch},
  \bibinfo{person}{Marc Sturm}, {and} \bibinfo{person}{Noemie Elhadad}.}
  \bibinfo{year}{2015}\natexlab{}.
\newblock \showarticletitle{Intelligible models for healthcare: Predicting
  pneumonia risk and hospital 30-day readmission}. In
  \bibinfo{booktitle}{\emph{Proceedings of the 21th ACM SIGKDD international
  conference on knowledge discovery and data mining}}.
  \bibinfo{pages}{1721--1730}.
\newblock


\bibitem[Cavallo and Demiralp(2018)]%
        {cavallo2018visual}
\bibfield{author}{\bibinfo{person}{Marco Cavallo} {and}
  \bibinfo{person}{{\c{C}}a{\u{g}}atay Demiralp}.}
  \bibinfo{year}{2018}\natexlab{}.
\newblock \showarticletitle{A visual interaction framework for dimensionality
  reduction based data exploration}. In \bibinfo{booktitle}{\emph{Proceedings
  of the 2018 CHI Conference on Human Factors in Computing Systems}}.
  \bibinfo{pages}{1--13}.
\newblock


\bibitem[Chandrasekaran(2005)]%
        {chandrasekaran2005makes}
\bibfield{author}{\bibinfo{person}{B Chandrasekaran}.}
  \bibinfo{year}{2005}\natexlab{}.
\newblock \showarticletitle{What makes a bunch of marks a diagrammatic
  representation, and another bunch a sentential representation?}. In
  \bibinfo{booktitle}{\emph{AAAI Spring Symposium: Reasoning with Mental and
  External Diagrams: Computational Modeling and Spatial Assistance}}.
  \bibinfo{pages}{83--89}.
\newblock


\bibitem[Corti et~al\mbox{.}(2024)]%
        {corti2024moving}
\bibfield{author}{\bibinfo{person}{Lorenzo Corti}, \bibinfo{person}{Rembrandt
  Oltmans}, \bibinfo{person}{Jiwon Jung}, \bibinfo{person}{Agathe Balayn},
  \bibinfo{person}{Marlies Wijsenbeek}, {and} \bibinfo{person}{Jie Yang}.}
  \bibinfo{year}{2024}\natexlab{}.
\newblock \showarticletitle{``It Is a Moving Process": Understanding the
  Evolution of Explainability Needs of Clinicians in Pulmonary Medicine}. In
  \bibinfo{booktitle}{\emph{Proceedings of the CHI Conference on Human Factors
  in Computing Systems}}. \bibinfo{pages}{1--21}.
\newblock


\bibitem[Dai et~al\mbox{.}(2019)]%
        {dai2019bridging}
\bibfield{author}{\bibinfo{person}{Wang-Zhou Dai}, \bibinfo{person}{Qiuling
  Xu}, \bibinfo{person}{Yang Yu}, {and} \bibinfo{person}{Zhi-Hua Zhou}.}
  \bibinfo{year}{2019}\natexlab{}.
\newblock \showarticletitle{Bridging machine learning and logical reasoning by
  abductive learning}.
\newblock \bibinfo{journal}{\emph{Advances in Neural Information Processing
  Systems}}  \bibinfo{volume}{32} (\bibinfo{year}{2019}).
\newblock


\bibitem[Dhanorkar et~al\mbox{.}(2021)]%
        {dhanorkar2021needs}
\bibfield{author}{\bibinfo{person}{Shipi Dhanorkar},
  \bibinfo{person}{Christine~T Wolf}, \bibinfo{person}{Kun Qian},
  \bibinfo{person}{Anbang Xu}, \bibinfo{person}{Lucian Popa}, {and}
  \bibinfo{person}{Yunyao Li}.} \bibinfo{year}{2021}\natexlab{}.
\newblock \showarticletitle{Who needs to know what, when?: Broadening the
  Explainable AI (XAI) Design Space by Looking at Explanations Across the AI
  Lifecycle}. In \bibinfo{booktitle}{\emph{Designing Interactive Systems
  Conference 2021}}. \bibinfo{pages}{1591--1602}.
\newblock


\bibitem[Dissanayake et~al\mbox{.}(2020)]%
        {dissanayake2020robust}
\bibfield{author}{\bibinfo{person}{Theekshana Dissanayake},
  \bibinfo{person}{Tharindu Fernando}, \bibinfo{person}{Simon Denman},
  \bibinfo{person}{Sridha Sridharan}, \bibinfo{person}{Houman Ghaemmaghami},
  {and} \bibinfo{person}{Clinton Fookes}.} \bibinfo{year}{2020}\natexlab{}.
\newblock \showarticletitle{A robust interpretable deep learning classifier for
  heart anomaly detection without segmentation}.
\newblock \bibinfo{journal}{\emph{IEEE Journal of Biomedical and Health
  Informatics}} \bibinfo{volume}{25}, \bibinfo{number}{6}
  (\bibinfo{year}{2020}), \bibinfo{pages}{2162--2171}.
\newblock


\bibitem[Du et~al\mbox{.}(2021)]%
        {du2021learning}
\bibfield{author}{\bibinfo{person}{Li Du}, \bibinfo{person}{Xiao Ding},
  \bibinfo{person}{Ting Liu}, {and} \bibinfo{person}{Bing Qin}.}
  \bibinfo{year}{2021}\natexlab{}.
\newblock \showarticletitle{Learning event graph knowledge for abductive
  reasoning}. In \bibinfo{booktitle}{\emph{Proceedings of the 59th Annual
  Meeting of the Association for Computational Linguistics and the 11th
  International Joint Conference on Natural Language Processing (Volume 1: Long
  Papers)}}. \bibinfo{pages}{5181--5190}.
\newblock


\bibitem[Dwivedi et~al\mbox{.}(2018)]%
        {dwivedi2018algorithms}
\bibfield{author}{\bibinfo{person}{Amit~Krishna Dwivedi},
  \bibinfo{person}{Syed~Anas Imtiaz}, {and} \bibinfo{person}{Esther
  Rodriguez-Villegas}.} \bibinfo{year}{2018}\natexlab{}.
\newblock \showarticletitle{Algorithms for automatic analysis and
  classification of heart sounds--a systematic review}.
\newblock \bibinfo{journal}{\emph{IEEE Access}}  \bibinfo{volume}{7}
  (\bibinfo{year}{2018}), \bibinfo{pages}{8316--8345}.
\newblock


\bibitem[Dzindolet et~al\mbox{.}(2003)]%
        {dzindolet2003role}
\bibfield{author}{\bibinfo{person}{Mary~T Dzindolet}, \bibinfo{person}{Scott~A
  Peterson}, \bibinfo{person}{Regina~A Pomranky}, \bibinfo{person}{Linda~G
  Pierce}, {and} \bibinfo{person}{Hall~P Beck}.}
  \bibinfo{year}{2003}\natexlab{}.
\newblock \showarticletitle{The role of trust in automation reliance}.
\newblock \bibinfo{journal}{\emph{International journal of human-computer
  studies}} \bibinfo{volume}{58}, \bibinfo{number}{6} (\bibinfo{year}{2003}),
  \bibinfo{pages}{697--718}.
\newblock


\bibitem[Ehsan et~al\mbox{.}(2018)]%
        {ehsan2018rationalization}
\bibfield{author}{\bibinfo{person}{Upol Ehsan}, \bibinfo{person}{Brent
  Harrison}, \bibinfo{person}{Larry Chan}, {and} \bibinfo{person}{Mark~O
  Riedl}.} \bibinfo{year}{2018}\natexlab{}.
\newblock \showarticletitle{Rationalization: A neural machine translation
  approach to generating natural language explanations}. In
  \bibinfo{booktitle}{\emph{Proceedings of the 2018 AAAI/ACM Conference on AI,
  Ethics, and Society}}. \bibinfo{pages}{81--87}.
\newblock


\bibitem[Ehsan et~al\mbox{.}(2021)]%
        {ehsan2021expanding}
\bibfield{author}{\bibinfo{person}{Upol Ehsan}, \bibinfo{person}{Q~Vera Liao},
  \bibinfo{person}{Michael Muller}, \bibinfo{person}{Mark~O Riedl}, {and}
  \bibinfo{person}{Justin~D Weisz}.} \bibinfo{year}{2021}\natexlab{}.
\newblock \showarticletitle{Expanding explainability: Towards social
  transparency in ai systems}. In \bibinfo{booktitle}{\emph{Proceedings of the
  2021 CHI Conference on Human Factors in Computing Systems}}.
\newblock


\bibitem[Ehsan et~al\mbox{.}(2019)]%
        {ehsan2019automated}
\bibfield{author}{\bibinfo{person}{Upol Ehsan}, \bibinfo{person}{Pradyumna
  Tambwekar}, \bibinfo{person}{Larry Chan}, \bibinfo{person}{Brent Harrison},
  {and} \bibinfo{person}{Mark~O Riedl}.} \bibinfo{year}{2019}\natexlab{}.
\newblock \showarticletitle{Automated rationale generation: a technique for
  explainable AI and its effects on human perceptions}. In
  \bibinfo{booktitle}{\emph{Proceedings of the 24th International Conference on
  Intelligent User Interfaces}}. \bibinfo{pages}{263--274}.
\newblock


\bibitem[Eiband et~al\mbox{.}(2018)]%
        {eiband2018bringing}
\bibfield{author}{\bibinfo{person}{Malin Eiband}, \bibinfo{person}{Hanna
  Schneider}, \bibinfo{person}{Mark Bilandzic}, \bibinfo{person}{Julian
  Fazekas-Con}, \bibinfo{person}{Mareike Haug}, {and} \bibinfo{person}{Heinrich
  Hussmann}.} \bibinfo{year}{2018}\natexlab{}.
\newblock \showarticletitle{Bringing transparency design into practice}. In
  \bibinfo{booktitle}{\emph{Proceedings of the 23rd International Conference on
  Intelligent User Interfaces}}.
\newblock


\bibitem[Erion et~al\mbox{.}(2021)]%
        {erion2021improving}
\bibfield{author}{\bibinfo{person}{Gabriel Erion}, \bibinfo{person}{Joseph~D
  Janizek}, \bibinfo{person}{Pascal Sturmfels}, \bibinfo{person}{Scott~M
  Lundberg}, {and} \bibinfo{person}{Su-In Lee}.}
  \bibinfo{year}{2021}\natexlab{}.
\newblock \showarticletitle{Improving performance of deep learning models with
  axiomatic attribution priors and expected gradients}.
\newblock \bibinfo{journal}{\emph{Nature machine intelligence}}
  \bibinfo{volume}{3}, \bibinfo{number}{7} (\bibinfo{year}{2021}).
\newblock


\bibitem[Farha and Gall(2019)]%
        {farha2019ms}
\bibfield{author}{\bibinfo{person}{Yazan~Abu Farha} {and}
  \bibinfo{person}{Jurgen Gall}.} \bibinfo{year}{2019}\natexlab{}.
\newblock \showarticletitle{Ms-tcn: Multi-stage temporal convolutional network
  for action segmentation}. In \bibinfo{booktitle}{\emph{Proceedings of the
  IEEE/CVF Conference on Computer Vision and Pattern Recognition}}.
  \bibinfo{pages}{3575--3584}.
\newblock


\bibitem[Fraser et~al\mbox{.}(1989)]%
        {fraser1989errors}
\bibfield{author}{\bibinfo{person}{Jane~M Fraser}, \bibinfo{person}{Patricia
  Strohm}, \bibinfo{person}{Jack~W Smith}, \bibinfo{person}{John~R Svirbely},
  \bibinfo{person}{Sally Rudmann}, \bibinfo{person}{TE Miller},
  \bibinfo{person}{Janice Blazina}, \bibinfo{person}{Melanie Kennedy}, {and}
  \bibinfo{person}{Philip~J Smith}.} \bibinfo{year}{1989}\natexlab{}.
\newblock \showarticletitle{Errors in abductive reasoning}. In
  \bibinfo{booktitle}{\emph{Conference Proceedings., IEEE International
  Conference on Systems, Man and Cybernetics}}. IEEE,
  \bibinfo{pages}{1136--1141}.
\newblock


\bibitem[Guidotti et~al\mbox{.}(2018)]%
        {guidotti2018survey}
\bibfield{author}{\bibinfo{person}{Riccardo Guidotti}, \bibinfo{person}{Anna
  Monreale}, \bibinfo{person}{Salvatore Ruggieri}, \bibinfo{person}{Franco
  Turini}, \bibinfo{person}{Fosca Giannotti}, {and} \bibinfo{person}{Dino
  Pedreschi}.} \bibinfo{year}{2018}\natexlab{}.
\newblock \showarticletitle{A survey of methods for explaining black box
  models}.
\newblock \bibinfo{journal}{\emph{ACM computing surveys (CSUR)}}
  \bibinfo{volume}{51}, \bibinfo{number}{5} (\bibinfo{year}{2018}),
  \bibinfo{pages}{1--42}.
\newblock


\bibitem[Harman(1965)]%
        {harman1965inference}
\bibfield{author}{\bibinfo{person}{Gilbert~H Harman}.}
  \bibinfo{year}{1965}\natexlab{}.
\newblock \showarticletitle{The inference to the best explanation}.
\newblock \bibinfo{journal}{\emph{The philosophical review}}
  \bibinfo{volume}{74}, \bibinfo{number}{1} (\bibinfo{year}{1965}),
  \bibinfo{pages}{88--95}.
\newblock


\bibitem[Hershey et~al\mbox{.}(2017)]%
        {hershey2017cnn}
\bibfield{author}{\bibinfo{person}{Shawn Hershey}, \bibinfo{person}{Sourish
  Chaudhuri}, \bibinfo{person}{Daniel~PW Ellis}, \bibinfo{person}{Jort~F
  Gemmeke}, \bibinfo{person}{Aren Jansen}, \bibinfo{person}{R~Channing Moore},
  \bibinfo{person}{Manoj Plakal}, \bibinfo{person}{Devin Platt},
  \bibinfo{person}{Rif~A Saurous}, \bibinfo{person}{Bryan Seybold},
  {et~al\mbox{.}}} \bibinfo{year}{2017}\natexlab{}.
\newblock \showarticletitle{CNN architectures for large-scale audio
  classification}. In \bibinfo{booktitle}{\emph{2017 ieee international
  conference on acoustics, speech and signal processing (icassp)}}. IEEE,
  \bibinfo{pages}{131--135}.
\newblock


\bibitem[Hoffman et~al\mbox{.}(2020)]%
        {hoffman2020explaining}
\bibfield{author}{\bibinfo{person}{Robert~R Hoffman},
  \bibinfo{person}{William~J Clancey}, {and} \bibinfo{person}{Shane~T
  Mueller}.} \bibinfo{year}{2020}\natexlab{}.
\newblock \showarticletitle{Explaining AI as an exploratory process: The
  peircean abduction model}.
\newblock \bibinfo{journal}{\emph{arXiv preprint arXiv:2009.14795}}
  (\bibinfo{year}{2020}).
\newblock


\bibitem[Hoffman and Klein(2017)]%
        {hoffman2017explaining}
\bibfield{author}{\bibinfo{person}{Robert~R Hoffman} {and}
  \bibinfo{person}{Gary Klein}.} \bibinfo{year}{2017}\natexlab{}.
\newblock \showarticletitle{Explaining explanation, part 1: theoretical
  foundations}.
\newblock \bibinfo{journal}{\emph{IEEE Intelligent Systems}}
  \bibinfo{volume}{32}, \bibinfo{number}{3} (\bibinfo{year}{2017}),
  \bibinfo{pages}{68--73}.
\newblock


\bibitem[Hoffman et~al\mbox{.}(2022)]%
        {hoffman2022psychology}
\bibfield{author}{\bibinfo{person}{Robert~R Hoffman}, \bibinfo{person}{Timothy
  Miller}, {and} \bibinfo{person}{William~J Clancey}.}
  \bibinfo{year}{2022}\natexlab{}.
\newblock \showarticletitle{Psychology and AI at a Crossroads: How Might
  Complex Systems Explain Themselves?}
\newblock \bibinfo{journal}{\emph{The American journal of psychology}}
  \bibinfo{volume}{135}, \bibinfo{number}{4} (\bibinfo{year}{2022}),
  \bibinfo{pages}{365--378}.
\newblock


\bibitem[Hoffman et~al\mbox{.}(2023)]%
        {hoffman2023measures}
\bibfield{author}{\bibinfo{person}{Robert~R Hoffman}, \bibinfo{person}{Shane~T
  Mueller}, \bibinfo{person}{Gary Klein}, {and} \bibinfo{person}{Jordan
  Litman}.} \bibinfo{year}{2023}\natexlab{}.
\newblock \showarticletitle{Measures for explainable AI: Explanation goodness,
  user satisfaction, mental models, curiosity, trust, and human-AI
  performance}.
\newblock \bibinfo{journal}{\emph{Frontiers in Computer Science}}
  \bibinfo{volume}{5} (\bibinfo{year}{2023}), \bibinfo{pages}{1096257}.
\newblock


\bibitem[Hoffmann(2010)]%
        {hoffmann2010diagrams}
\bibfield{author}{\bibinfo{person}{Michael Hans~Georg Hoffmann}.}
  \bibinfo{year}{2010}\natexlab{}.
\newblock \showarticletitle{Diagrams as scaffolds for abductive insights}. In
  \bibinfo{booktitle}{\emph{Workshops at the Twenty-Fourth AAAI Conference on
  Artificial Intelligence}}.
\newblock


\bibitem[Hohman et~al\mbox{.}(2018)]%
        {hohman2018visual}
\bibfield{author}{\bibinfo{person}{Fred Hohman}, \bibinfo{person}{Minsuk
  Kahng}, \bibinfo{person}{Robert Pienta}, {and} \bibinfo{person}{Duen~Horng
  Chau}.} \bibinfo{year}{2018}\natexlab{}.
\newblock \showarticletitle{Visual analytics in deep learning: An interrogative
  survey for the next frontiers}.
\newblock \bibinfo{journal}{\emph{IEEE transactions on visualization and
  computer graphics}} \bibinfo{volume}{25}, \bibinfo{number}{8}
  (\bibinfo{year}{2018}), \bibinfo{pages}{2674--2693}.
\newblock


\bibitem[Hohman et~al\mbox{.}(2019)]%
        {hohman2019s}
\bibfield{author}{\bibinfo{person}{Fred Hohman}, \bibinfo{person}{Haekyu Park},
  \bibinfo{person}{Caleb Robinson}, {and} \bibinfo{person}{Duen Horng~Polo
  Chau}.} \bibinfo{year}{2019}\natexlab{}.
\newblock \showarticletitle{Summit: Scaling deep learning interpretability by
  visualizing activation and attribution summarizations}.
\newblock \bibinfo{journal}{\emph{IEEE transactions on visualization and
  computer graphics}} \bibinfo{volume}{26}, \bibinfo{number}{1}
  (\bibinfo{year}{2019}), \bibinfo{pages}{1096--1106}.
\newblock


\bibitem[Holten and Van~Wijk(2009)]%
        {holten2009force}
\bibfield{author}{\bibinfo{person}{Danny Holten} {and} \bibinfo{person}{Jarke~J
  Van~Wijk}.} \bibinfo{year}{2009}\natexlab{}.
\newblock \showarticletitle{Force-directed edge bundling for graph
  visualization}. In \bibinfo{booktitle}{\emph{Computer graphics forum}},
  Vol.~\bibinfo{volume}{28}. Wiley Online Library.
\newblock


\bibitem[Hudson and Manning(2018)]%
        {hudson2018compositional}
\bibfield{author}{\bibinfo{person}{Drew~A Hudson} {and}
  \bibinfo{person}{Christopher~D Manning}.} \bibinfo{year}{2018}\natexlab{}.
\newblock \showarticletitle{Compositional attention networks for machine
  reasoning}.
\newblock \bibinfo{journal}{\emph{arXiv preprint arXiv:1803.03067}}
  (\bibinfo{year}{2018}).
\newblock


\bibitem[Hurter et~al\mbox{.}(2012)]%
        {hurter2012graph}
\bibfield{author}{\bibinfo{person}{Christophe Hurter}, \bibinfo{person}{Ozan
  Ersoy}, {and} \bibinfo{person}{Alexandru Telea}.}
  \bibinfo{year}{2012}\natexlab{}.
\newblock \showarticletitle{Graph bundling by kernel density estimation}. In
  \bibinfo{booktitle}{\emph{Computer graphics forum}},
  Vol.~\bibinfo{volume}{31}. Wiley Online Library, \bibinfo{pages}{865--874}.
\newblock


\bibitem[Izza et~al\mbox{.}(2023)]%
        {izza2023computing}
\bibfield{author}{\bibinfo{person}{Yacine Izza}, \bibinfo{person}{Xuanxiang
  Huang}, \bibinfo{person}{Alexey Ignatiev}, \bibinfo{person}{Nina Narodytska},
  \bibinfo{person}{Martin Cooper}, {and} \bibinfo{person}{Joao Marques-Silva}.}
  \bibinfo{year}{2023}\natexlab{}.
\newblock \showarticletitle{On computing probabilistic abductive explanations}.
\newblock \bibinfo{journal}{\emph{International Journal of Approximate
  Reasoning}}  \bibinfo{volume}{159} (\bibinfo{year}{2023}).
\newblock


\bibitem[Johnson et~al\mbox{.}(2019)]%
        {johnson2019simplicity}
\bibfield{author}{\bibinfo{person}{Samuel~GB Johnson}, \bibinfo{person}{JJ
  Valenti}, {and} \bibinfo{person}{Frank~C Keil}.}
  \bibinfo{year}{2019}\natexlab{}.
\newblock \showarticletitle{Simplicity and complexity preferences in causal
  explanation: An opponent heuristic account}.
\newblock \bibinfo{journal}{\emph{Cognitive psychology}}  \bibinfo{volume}{113}
  (\bibinfo{year}{2019}), \bibinfo{pages}{101222}.
\newblock


\bibitem[Josephson and Josephson(1996)]%
        {josephson1996abductive}
\bibfield{author}{\bibinfo{person}{John~R Josephson} {and}
  \bibinfo{person}{Susan~G Josephson}.} \bibinfo{year}{1996}\natexlab{}.
\newblock \bibinfo{booktitle}{\emph{Abductive inference: Computation,
  philosophy, technology}}.
\newblock \bibinfo{publisher}{Cambridge University Press}.
\newblock


\bibitem[Judge and Mangrulkar(2015)]%
        {judge2015heart}
\bibfield{author}{\bibinfo{person}{Richard Judge} {and} \bibinfo{person}{Rajesh
  Mangrulkar}.} \bibinfo{year}{2015}\natexlab{}.
\newblock \bibinfo{title}{Heart Sound and Murmur Library}.
\newblock
  \bibinfo{howpublished}{\url{https://www.med.umich.edu/lrc/psb_open/html/repo/primer_heartsound/primer_heartsound.html}}.
\newblock
\newblock
\shownote{[Online; accessed 14-Sep-2022]}.


\bibitem[Kahng et~al\mbox{.}(2017)]%
        {kahng2017cti}
\bibfield{author}{\bibinfo{person}{Minsuk Kahng}, \bibinfo{person}{Pierre~Y
  Andrews}, \bibinfo{person}{Aditya Kalro}, {and} \bibinfo{person}{Duen~Horng
  Chau}.} \bibinfo{year}{2017}\natexlab{}.
\newblock \showarticletitle{A cti v is: Visual exploration of industry-scale
  deep neural network models}.
\newblock \bibinfo{journal}{\emph{IEEE transactions on visualization and
  computer graphics}} \bibinfo{volume}{24}, \bibinfo{number}{1}
  (\bibinfo{year}{2017}), \bibinfo{pages}{88--97}.
\newblock


\bibitem[Kahng et~al\mbox{.}(2018)]%
        {kahng2018gan}
\bibfield{author}{\bibinfo{person}{Minsuk Kahng}, \bibinfo{person}{Nikhil
  Thorat}, \bibinfo{person}{Duen~Horng Chau}, \bibinfo{person}{Fernanda~B
  Vi{\'e}gas}, {and} \bibinfo{person}{Martin Wattenberg}.}
  \bibinfo{year}{2018}\natexlab{}.
\newblock \showarticletitle{Gan lab: Understanding complex deep generative
  models using interactive visual experimentation}.
\newblock \bibinfo{journal}{\emph{IEEE transactions on visualization and
  computer graphics}} \bibinfo{volume}{25}, \bibinfo{number}{1}
  (\bibinfo{year}{2018}), \bibinfo{pages}{310--320}.
\newblock


\bibitem[Kakas et~al\mbox{.}(1992)]%
        {kakas1992abductive}
\bibfield{author}{\bibinfo{person}{Antonis~C Kakas}, \bibinfo{person}{Robert~A.
  Kowalski}, {and} \bibinfo{person}{Francesca Toni}.}
  \bibinfo{year}{1992}\natexlab{}.
\newblock \showarticletitle{Abductive logic programming}.
\newblock \bibinfo{journal}{\emph{Journal of logic and computation}}
  \bibinfo{volume}{2}, \bibinfo{number}{6} (\bibinfo{year}{1992}),
  \bibinfo{pages}{719--770}.
\newblock


\bibitem[Kaur et~al\mbox{.}(2022)]%
        {kaur2022sensible}
\bibfield{author}{\bibinfo{person}{Harmanpreet Kaur}, \bibinfo{person}{Eytan
  Adar}, \bibinfo{person}{Eric Gilbert}, {and} \bibinfo{person}{Cliff Lampe}.}
  \bibinfo{year}{2022}\natexlab{}.
\newblock \showarticletitle{Sensible AI: Re-imagining Interpretability and
  Explainability using Sensemaking Theory}. In \bibinfo{booktitle}{\emph{2022
  ACM Conference on Fairness, Accountability, and Transparency}}.
\newblock


\bibitem[Kaur et~al\mbox{.}(2020)]%
        {kaur2020interpreting}
\bibfield{author}{\bibinfo{person}{Harmanpreet Kaur}, \bibinfo{person}{Harsha
  Nori}, \bibinfo{person}{Samuel Jenkins}, \bibinfo{person}{Rich Caruana},
  \bibinfo{person}{Hanna Wallach}, {and} \bibinfo{person}{Jennifer
  Wortman~Vaughan}.} \bibinfo{year}{2020}\natexlab{}.
\newblock \showarticletitle{Interpreting interpretability: understanding data
  scientists' use of interpretability tools for machine learning}. In
  \bibinfo{booktitle}{\emph{Proceedings of the 2020 CHI conference on human
  factors in computing systems}}.
\newblock


\bibitem[Keil(2006)]%
        {keil2006explanation}
\bibfield{author}{\bibinfo{person}{Frank~C Keil}.}
  \bibinfo{year}{2006}\natexlab{}.
\newblock \showarticletitle{Explanation and understanding}.
\newblock \bibinfo{journal}{\emph{Annu. Rev. Psychol.}} \bibinfo{volume}{57},
  \bibinfo{number}{1} (\bibinfo{year}{2006}), \bibinfo{pages}{227--254}.
\newblock


\bibitem[Kim et~al\mbox{.}(2016)]%
        {kim2016examples}
\bibfield{author}{\bibinfo{person}{Been Kim}, \bibinfo{person}{Rajiv Khanna},
  {and} \bibinfo{person}{Oluwasanmi~O Koyejo}.}
  \bibinfo{year}{2016}\natexlab{}.
\newblock \showarticletitle{Examples are not enough, learn to criticize!
  criticism for interpretability}.
\newblock \bibinfo{journal}{\emph{Advances in neural information processing
  systems}}  \bibinfo{volume}{29} (\bibinfo{year}{2016}).
\newblock


\bibitem[Kim et~al\mbox{.}(2018)]%
        {kim2018interpretability}
\bibfield{author}{\bibinfo{person}{Been Kim}, \bibinfo{person}{Martin
  Wattenberg}, \bibinfo{person}{Justin Gilmer}, \bibinfo{person}{Carrie Cai},
  \bibinfo{person}{James Wexler}, \bibinfo{person}{Fernanda Viegas},
  {et~al\mbox{.}}} \bibinfo{year}{2018}\natexlab{}.
\newblock \showarticletitle{Interpretability beyond feature attribution:
  Quantitative testing with concept activation vectors (TCAV)}. In
  \bibinfo{booktitle}{\emph{International conference on machine learning}}.
  PMLR, \bibinfo{pages}{2668--2677}.
\newblock


\bibitem[Koh and Liang(2017)]%
        {koh2017understanding}
\bibfield{author}{\bibinfo{person}{Pang~Wei Koh} {and} \bibinfo{person}{Percy
  Liang}.} \bibinfo{year}{2017}\natexlab{}.
\newblock \showarticletitle{Understanding black-box predictions via influence
  functions}. In \bibinfo{booktitle}{\emph{International Conference on Machine
  Learning}}. PMLR.
\newblock


\bibitem[Koh et~al\mbox{.}(2020)]%
        {koh2020concept}
\bibfield{author}{\bibinfo{person}{Pang~Wei Koh}, \bibinfo{person}{Thao
  Nguyen}, \bibinfo{person}{Yew~Siang Tang}, \bibinfo{person}{Stephen
  Mussmann}, \bibinfo{person}{Emma Pierson}, \bibinfo{person}{Been Kim}, {and}
  \bibinfo{person}{Percy Liang}.} \bibinfo{year}{2020}\natexlab{}.
\newblock \showarticletitle{Concept bottleneck models}. In
  \bibinfo{booktitle}{\emph{International Conference on Machine Learning}}.
  PMLR, \bibinfo{pages}{5338--5348}.
\newblock


\bibitem[Kohl et~al\mbox{.}(2018)]%
        {kohl2018probabilistic}
\bibfield{author}{\bibinfo{person}{Simon Kohl}, \bibinfo{person}{Bernardino
  Romera-Paredes}, \bibinfo{person}{Clemens Meyer}, \bibinfo{person}{Jeffrey
  De~Fauw}, \bibinfo{person}{Joseph~R Ledsam}, \bibinfo{person}{Klaus
  Maier-Hein}, \bibinfo{person}{SM Eslami}, \bibinfo{person}{Danilo
  Jimenez~Rezende}, {and} \bibinfo{person}{Olaf Ronneberger}.}
  \bibinfo{year}{2018}\natexlab{}.
\newblock \showarticletitle{A probabilistic u-net for segmentation of ambiguous
  images}.
\newblock \bibinfo{journal}{\emph{Advances in neural information processing
  systems}}  \bibinfo{volume}{31} (\bibinfo{year}{2018}).
\newblock


\bibitem[Koller(2009)]%
        {koller2009probabilistic}
\bibfield{author}{\bibinfo{person}{Daphane Koller}.}
  \bibinfo{year}{2009}\natexlab{}.
\newblock \bibinfo{title}{Probabilistic Graphical Models: Principles and
  Techniques}.
\newblock
\newblock


\bibitem[Krause et~al\mbox{.}(2016)]%
        {krause2016interacting}
\bibfield{author}{\bibinfo{person}{Josua Krause}, \bibinfo{person}{Adam Perer},
  {and} \bibinfo{person}{Kenney Ng}.} \bibinfo{year}{2016}\natexlab{}.
\newblock \showarticletitle{Interacting with predictions: Visual inspection of
  black-box machine learning models}. In \bibinfo{booktitle}{\emph{Proceedings
  of the 2016 CHI conference on human factors in computing systems}}.
  \bibinfo{pages}{5686--5697}.
\newblock


\bibitem[Kulesza et~al\mbox{.}(2009)]%
        {kulesza2009fixing}
\bibfield{author}{\bibinfo{person}{Todd Kulesza}, \bibinfo{person}{Weng-Keen
  Wong}, \bibinfo{person}{Simone Stumpf}, \bibinfo{person}{Stephen Perona},
  \bibinfo{person}{Rachel White}, \bibinfo{person}{Margaret~M Burnett},
  \bibinfo{person}{Ian Oberst}, {and} \bibinfo{person}{Amy~J Ko}.}
  \bibinfo{year}{2009}\natexlab{}.
\newblock \showarticletitle{Fixing the program my computer learned: Barriers
  for end users, challenges for the machine}. In
  \bibinfo{booktitle}{\emph{Proceedings of the 14th international conference on
  Intelligent user interfaces}}.
\newblock


\bibitem[Lage et~al\mbox{.}(2019)]%
        {lage2019evaluation}
\bibfield{author}{\bibinfo{person}{Isaac Lage}, \bibinfo{person}{Emily Chen},
  \bibinfo{person}{Jeffrey He}, \bibinfo{person}{Menaka Narayanan},
  \bibinfo{person}{Been Kim}, \bibinfo{person}{Sam Gershman}, {and}
  \bibinfo{person}{Finale Doshi-Velez}.} \bibinfo{year}{2019}\natexlab{}.
\newblock \showarticletitle{An evaluation of the human-interpretability of
  explanation}.
\newblock \bibinfo{journal}{\emph{arXiv preprint arXiv:1902.00006}}
  (\bibinfo{year}{2019}).
\newblock


\bibitem[Lage et~al\mbox{.}(2018)]%
        {lage2018human}
\bibfield{author}{\bibinfo{person}{Isaac Lage}, \bibinfo{person}{Andrew Ross},
  \bibinfo{person}{Samuel~J Gershman}, \bibinfo{person}{Been Kim}, {and}
  \bibinfo{person}{Finale Doshi-Velez}.} \bibinfo{year}{2018}\natexlab{}.
\newblock \showarticletitle{Human-in-the-loop interpretability prior}.
\newblock \bibinfo{journal}{\emph{Advances in neural information processing
  systems}}  \bibinfo{volume}{31} (\bibinfo{year}{2018}).
\newblock


\bibitem[Lakkaraju et~al\mbox{.}(2016)]%
        {lakkaraju2016interpretable}
\bibfield{author}{\bibinfo{person}{Himabindu Lakkaraju},
  \bibinfo{person}{Stephen~H Bach}, {and} \bibinfo{person}{Jure Leskovec}.}
  \bibinfo{year}{2016}\natexlab{}.
\newblock \showarticletitle{Interpretable decision sets: A joint framework for
  description and prediction}. In \bibinfo{booktitle}{\emph{Proceedings of the
  22nd ACM SIGKDD international conference on knowledge discovery and data
  mining}}. \bibinfo{pages}{1675--1684}.
\newblock


\bibitem[Langer et~al\mbox{.}(2021)]%
        {langer2021we}
\bibfield{author}{\bibinfo{person}{Markus Langer}, \bibinfo{person}{Daniel
  Oster}, \bibinfo{person}{Timo Speith}, \bibinfo{person}{Holger Hermanns},
  \bibinfo{person}{Lena K{\"a}stner}, \bibinfo{person}{Eva Schmidt},
  \bibinfo{person}{Andreas Sesing}, {and} \bibinfo{person}{Kevin Baum}.}
  \bibinfo{year}{2021}\natexlab{}.
\newblock \showarticletitle{What do we want from Explainable Artificial
  Intelligence (XAI)?--A stakeholder perspective on XAI and a conceptual model
  guiding interdisciplinary XAI research}.
\newblock \bibinfo{journal}{\emph{Artificial Intelligence}}
  \bibinfo{volume}{296} (\bibinfo{year}{2021}), \bibinfo{pages}{103473}.
\newblock


\bibitem[Larkin and Simon(1987)]%
        {larkin1987diagram}
\bibfield{author}{\bibinfo{person}{Jill~H Larkin} {and}
  \bibinfo{person}{Herbert~A Simon}.} \bibinfo{year}{1987}\natexlab{}.
\newblock \showarticletitle{Why a diagram is (sometimes) worth ten thousand
  words}.
\newblock \bibinfo{journal}{\emph{Cognitive science}} \bibinfo{volume}{11},
  \bibinfo{number}{1} (\bibinfo{year}{1987}), \bibinfo{pages}{65--100}.
\newblock


\bibitem[Leake(1995)]%
        {leake1995abduction}
\bibfield{author}{\bibinfo{person}{David~B Leake}.}
  \bibinfo{year}{1995}\natexlab{}.
\newblock \showarticletitle{Abduction, experience, and goals: A model of
  everyday abductive explanation}.
\newblock \bibinfo{journal}{\emph{Journal of Experimental \& Theoretical
  Artificial Intelligence}} \bibinfo{volume}{7}, \bibinfo{number}{4}
  (\bibinfo{year}{1995}), \bibinfo{pages}{407--428}.
\newblock


\bibitem[Lee et~al\mbox{.}(2022)]%
        {lee2022fully}
\bibfield{author}{\bibinfo{person}{Sung~Hoon Lee}, \bibinfo{person}{Yun-Soung
  Kim}, \bibinfo{person}{Min-Kyung Yeo}, \bibinfo{person}{Musa Mahmood},
  \bibinfo{person}{Nathan Zavanelli}, \bibinfo{person}{Chaeuk Chung},
  \bibinfo{person}{Jun~Young Heo}, \bibinfo{person}{Yoonjoo Kim},
  \bibinfo{person}{Sung-Soo Jung}, {and} \bibinfo{person}{Woon-Hong Yeo}.}
  \bibinfo{year}{2022}\natexlab{}.
\newblock \showarticletitle{Fully portable continuous real-time auscultation
  with a soft wearable stethoscope designed for automated disease diagnosis}.
\newblock \bibinfo{journal}{\emph{Science Advances}} \bibinfo{volume}{8},
  \bibinfo{number}{21} (\bibinfo{year}{2022}).
\newblock


\bibitem[Letham et~al\mbox{.}(2015)]%
        {letham2015interpretable}
\bibfield{author}{\bibinfo{person}{Benjamin Letham}, \bibinfo{person}{Cynthia
  Rudin}, \bibinfo{person}{Tyler~H McCormick}, {and} \bibinfo{person}{David
  Madigan}.} \bibinfo{year}{2015}\natexlab{}.
\newblock \showarticletitle{Interpretable classifiers using rules and bayesian
  analysis: Building a better stroke prediction model}.
\newblock \bibinfo{journal}{\emph{The Annals of Applied Statistics}}
  \bibinfo{volume}{9}, \bibinfo{number}{3} (\bibinfo{year}{2015}).
\newblock


\bibitem[Li et~al\mbox{.}(2023)]%
        {li2023multi}
\bibfield{author}{\bibinfo{person}{Mengze Li}, \bibinfo{person}{Tianbao Wang},
  \bibinfo{person}{Jiahe Xu}, \bibinfo{person}{Kairong Han},
  \bibinfo{person}{Shengyu Zhang}, \bibinfo{person}{Zhou Zhao},
  \bibinfo{person}{Jiaxu Miao}, \bibinfo{person}{Wenqiao Zhang},
  \bibinfo{person}{Shiliang Pu}, {and} \bibinfo{person}{Fei Wu}.}
  \bibinfo{year}{2023}\natexlab{}.
\newblock \showarticletitle{Multi-modal action chain abductive reasoning}. In
  \bibinfo{booktitle}{\emph{Proceedings of the 61st Annual Meeting of the
  Association for Computational Linguistics (Volume 1: Long Papers)}}.
  \bibinfo{pages}{4617--4628}.
\newblock


\bibitem[Liang et~al\mbox{.}(2022)]%
        {liang2022visual}
\bibfield{author}{\bibinfo{person}{Chen Liang}, \bibinfo{person}{Wenguan Wang},
  \bibinfo{person}{Tianfei Zhou}, {and} \bibinfo{person}{Yi Yang}.}
  \bibinfo{year}{2022}\natexlab{}.
\newblock \showarticletitle{Visual abductive reasoning}. In
  \bibinfo{booktitle}{\emph{Proceedings of the IEEE/CVF Conference on Computer
  Vision and Pattern Recognition}}. \bibinfo{pages}{15565--15575}.
\newblock


\bibitem[Liao et~al\mbox{.}(2020)]%
        {liao2020questioning}
\bibfield{author}{\bibinfo{person}{Q~Vera Liao}, \bibinfo{person}{Daniel
  Gruen}, {and} \bibinfo{person}{Sarah Miller}.}
  \bibinfo{year}{2020}\natexlab{}.
\newblock \showarticletitle{Questioning the AI: informing design practices for
  explainable AI user experiences}. In \bibinfo{booktitle}{\emph{Proceedings of
  the 2020 CHI Conference on Human Factors in Computing Systems}}.
  \bibinfo{pages}{1--15}.
\newblock


\bibitem[Lilly(2012)]%
        {lilly2012pathophysiology}
\bibfield{author}{\bibinfo{person}{Leonard~S Lilly}.}
  \bibinfo{year}{2012}\natexlab{}.
\newblock \bibinfo{booktitle}{\emph{Pathophysiology of heart disease: a
  collaborative project of medical students and faculty}}.
\newblock \bibinfo{publisher}{Lippincott Williams \& Wilkins}.
\newblock


\bibitem[Lim and Dey(2009)]%
        {lim2009assessing}
\bibfield{author}{\bibinfo{person}{Brian~Y Lim} {and} \bibinfo{person}{Anind~K
  Dey}.} \bibinfo{year}{2009}\natexlab{}.
\newblock \showarticletitle{Assessing demand for intelligibility in
  context-aware applications}. In \bibinfo{booktitle}{\emph{Proceedings of the
  11th international conference on Ubiquitous computing}}.
  \bibinfo{pages}{195--204}.
\newblock


\bibitem[Lim and Dey(2011a)]%
        {lim2011design}
\bibfield{author}{\bibinfo{person}{Brian~Y Lim} {and} \bibinfo{person}{Anind~K
  Dey}.} \bibinfo{year}{2011}\natexlab{a}.
\newblock \showarticletitle{Design of an intelligible mobile context-aware
  application}. In \bibinfo{booktitle}{\emph{Proceedings of the 13th
  international conference on human computer interaction with mobile devices
  and services}}. \bibinfo{pages}{157--166}.
\newblock


\bibitem[Lim and Dey(2011b)]%
        {lim2011investigating}
\bibfield{author}{\bibinfo{person}{Brian~Y Lim} {and} \bibinfo{person}{Anind~K
  Dey}.} \bibinfo{year}{2011}\natexlab{b}.
\newblock \showarticletitle{Investigating intelligibility for uncertain
  context-aware applications}. In \bibinfo{booktitle}{\emph{Proceedings of the
  13th international conference on Ubiquitous computing}}.
  \bibinfo{pages}{415--424}.
\newblock


\bibitem[Lim et~al\mbox{.}(2009)]%
        {lim2009and}
\bibfield{author}{\bibinfo{person}{Brian~Y Lim}, \bibinfo{person}{Anind~K Dey},
  {and} \bibinfo{person}{Daniel Avrahami}.} \bibinfo{year}{2009}\natexlab{}.
\newblock \showarticletitle{Why and why not explanations improve the
  intelligibility of context-aware intelligent systems}. In
  \bibinfo{booktitle}{\emph{Proceedings of the SIGCHI conference on human
  factors in computing systems}}. \bibinfo{pages}{2119--2128}.
\newblock


\bibitem[Lim et~al\mbox{.}(2019)]%
        {lim2019does}
\bibfield{author}{\bibinfo{person}{Brian~Y Lim}, \bibinfo{person}{Judy Kay},
  {and} \bibinfo{person}{Weilong Liu}.} \bibinfo{year}{2019}\natexlab{}.
\newblock \showarticletitle{How Does a Nation Walk? Interpreting Large-Scale
  Step Count Activity with Weekly Streak Patterns}.
\newblock \bibinfo{journal}{\emph{Proceedings of the ACM on Interactive,
  Mobile, Wearable and Ubiquitous Technologies}} \bibinfo{volume}{3},
  \bibinfo{number}{2} (\bibinfo{year}{2019}), \bibinfo{pages}{1--46}.
\newblock


\bibitem[Lundberg et~al\mbox{.}(2020)]%
        {lundberg2020local}
\bibfield{author}{\bibinfo{person}{Scott~M Lundberg}, \bibinfo{person}{Gabriel
  Erion}, \bibinfo{person}{Hugh Chen}, \bibinfo{person}{Alex DeGrave},
  \bibinfo{person}{Jordan~M Prutkin}, \bibinfo{person}{Bala Nair},
  \bibinfo{person}{Ronit Katz}, \bibinfo{person}{Jonathan Himmelfarb},
  \bibinfo{person}{Nisha Bansal}, {and} \bibinfo{person}{Su-In Lee}.}
  \bibinfo{year}{2020}\natexlab{}.
\newblock \showarticletitle{From local explanations to global understanding
  with explainable AI for trees}.
\newblock \bibinfo{journal}{\emph{Nature machine intelligence}}
  \bibinfo{volume}{2}, \bibinfo{number}{1} (\bibinfo{year}{2020}),
  \bibinfo{pages}{56--67}.
\newblock


\bibitem[Lundberg and Lee(2017)]%
        {lundberg2017unified}
\bibfield{author}{\bibinfo{person}{Scott~M Lundberg} {and}
  \bibinfo{person}{Su-In Lee}.} \bibinfo{year}{2017}\natexlab{}.
\newblock \showarticletitle{A unified approach to interpreting model
  predictions}.
\newblock \bibinfo{journal}{\emph{Advances in neural information processing
  systems}}  \bibinfo{volume}{30} (\bibinfo{year}{2017}).
\newblock


\bibitem[Lyell and Coiera(2017)]%
        {lyell2017automation}
\bibfield{author}{\bibinfo{person}{David Lyell} {and} \bibinfo{person}{Enrico
  Coiera}.} \bibinfo{year}{2017}\natexlab{}.
\newblock \showarticletitle{Automation bias and verification complexity: a
  systematic review}.
\newblock \bibinfo{journal}{\emph{Journal of the American Medical Informatics
  Association}} \bibinfo{volume}{24}, \bibinfo{number}{2}
  (\bibinfo{year}{2017}), \bibinfo{pages}{423--431}.
\newblock


\bibitem[Lyu et~al\mbox{.}(2019)]%
        {lyu2019od}
\bibfield{author}{\bibinfo{person}{Yan Lyu}, \bibinfo{person}{Xu Liu},
  \bibinfo{person}{Hanyi Chen}, \bibinfo{person}{Arpan Mangal},
  \bibinfo{person}{Kai Liu}, \bibinfo{person}{Chao Chen}, {and}
  \bibinfo{person}{Brian Lim}.} \bibinfo{year}{2019}\natexlab{}.
\newblock \showarticletitle{OD Morphing: Balancing simplicity with faithfulness
  for OD bundling}.
\newblock \bibinfo{journal}{\emph{IEEE Transactions on Visualization and
  Computer Graphics}} \bibinfo{volume}{26}, \bibinfo{number}{1}
  (\bibinfo{year}{2019}), \bibinfo{pages}{811--821}.
\newblock


\bibitem[Lyu et~al\mbox{.}(2023)]%
        {lyu2023if}
\bibfield{author}{\bibinfo{person}{Yan Lyu}, \bibinfo{person}{Hangxin Lu},
  \bibinfo{person}{Min~Kyung Lee}, \bibinfo{person}{Gerhard Schmitt}, {and}
  \bibinfo{person}{Brian~Y Lim}.} \bibinfo{year}{2023}\natexlab{}.
\newblock \showarticletitle{IF-City: Intelligible Fair City Planning to
  Measure, Explain and Mitigate Inequality}.
\newblock \bibinfo{journal}{\emph{IEEE Transactions on Visualization and
  Computer Graphics}} (\bibinfo{year}{2023}).
\newblock


\bibitem[Ma'ayan et~al\mbox{.}(2020)]%
        {ma2020domain}
\bibfield{author}{\bibinfo{person}{Dor Ma'ayan}, \bibinfo{person}{Wode Ni},
  \bibinfo{person}{Katherine Ye}, \bibinfo{person}{Chinmay Kulkarni}, {and}
  \bibinfo{person}{Joshua Sunshine}.} \bibinfo{year}{2020}\natexlab{}.
\newblock \showarticletitle{How domain experts create conceptual diagrams and
  implications for tool design}. In \bibinfo{booktitle}{\emph{Proceedings of
  the 2020 CHI Conference on Human Factors in Computing Systems}}.
  \bibinfo{pages}{1--14}.
\newblock


\bibitem[Magnani(2009)]%
        {magnani2009abductive}
\bibfield{author}{\bibinfo{person}{Lorenzo Magnani}.}
  \bibinfo{year}{2009}\natexlab{}.
\newblock \bibinfo{booktitle}{\emph{Abductive cognition: The epistemological
  and eco-cognitive dimensions of hypothetical reasoning}}.
  Vol.~\bibinfo{volume}{3}.
\newblock \bibinfo{publisher}{Springer}.
\newblock


\bibitem[Magnani(2011)]%
        {magnani2011abduction}
\bibfield{author}{\bibinfo{person}{Lorenzo Magnani}.}
  \bibinfo{year}{2011}\natexlab{}.
\newblock \bibinfo{booktitle}{\emph{Abduction, reason and science: Processes of
  discovery and explanation}}.
\newblock \bibinfo{publisher}{Springer Science \& Business Media}.
\newblock


\bibitem[Makatchev et~al\mbox{.}(2004b)]%
        {makatchev2004abductive}
\bibfield{author}{\bibinfo{person}{Maxim Makatchev}, \bibinfo{person}{Pamela~W
  Jordan}, \bibinfo{person}{Umarani Pappuswamy}, {and} \bibinfo{person}{Kurt
  VanLehn}.} \bibinfo{year}{2004}\natexlab{b}.
\newblock \showarticletitle{Abductive Proofs as Models of Students' Reasoning
  about Qualitative Physics.}. In \bibinfo{booktitle}{\emph{ICCM}}.
  \bibinfo{pages}{166--171}.
\newblock


\bibitem[Makatchev et~al\mbox{.}(2004a)]%
        {makatchev2004b_abductive}
\bibfield{author}{\bibinfo{person}{Maxim Makatchev}, \bibinfo{person}{Pamela~W
  Jordan}, {and} \bibinfo{person}{Kurt VanLehn}.}
  \bibinfo{year}{2004}\natexlab{a}.
\newblock \showarticletitle{Abductive theorem proving for analyzing student
  explanations to guide feedback in intelligent tutoring systems}.
\newblock \bibinfo{journal}{\emph{Journal of Automated Reasoning}}
  \bibinfo{volume}{32} (\bibinfo{year}{2004}), \bibinfo{pages}{187--226}.
\newblock


\bibitem[Matsuyama et~al\mbox{.}(2023)]%
        {matsuyama2023iris}
\bibfield{author}{\bibinfo{person}{Hitoshi Matsuyama}, \bibinfo{person}{Nobuo
  Kawaguchi}, {and} \bibinfo{person}{Brian~Y Lim}.}
  \bibinfo{year}{2023}\natexlab{}.
\newblock \showarticletitle{IRIS: Interpretable Rubric-Informed Segmentation
  for Action Quality Assessment}. In \bibinfo{booktitle}{\emph{Proceedings of
  the 28th International Conference on Intelligent User Interfaces}}.
\newblock


\bibitem[McInnes et~al\mbox{.}(2018)]%
        {mcinnes2018umap}
\bibfield{author}{\bibinfo{person}{Leland McInnes}, \bibinfo{person}{John
  Healy}, {and} \bibinfo{person}{James Melville}.}
  \bibinfo{year}{2018}\natexlab{}.
\newblock \showarticletitle{Umap: Uniform manifold approximation and projection
  for dimension reduction}.
\newblock \bibinfo{journal}{\emph{arXiv preprint arXiv:1802.03426}}
  (\bibinfo{year}{2018}).
\newblock


\bibitem[Medianovskyi and Pietarinen(2022)]%
        {medianovskyi2022explainable}
\bibfield{author}{\bibinfo{person}{Kyrylo Medianovskyi} {and}
  \bibinfo{person}{Ahti-Veikko Pietarinen}.} \bibinfo{year}{2022}\natexlab{}.
\newblock \showarticletitle{On Explainable AI and Abductive Inference}.
\newblock \bibinfo{journal}{\emph{Philosophies}} \bibinfo{volume}{7},
  \bibinfo{number}{2} (\bibinfo{year}{2022}), \bibinfo{pages}{35}.
\newblock


\bibitem[Miller(2019)]%
        {miller2019explanation}
\bibfield{author}{\bibinfo{person}{Tim Miller}.}
  \bibinfo{year}{2019}\natexlab{}.
\newblock \showarticletitle{Explanation in artificial intelligence: Insights
  from the social sciences}.
\newblock \bibinfo{journal}{\emph{Artificial intelligence}}
  \bibinfo{volume}{267} (\bibinfo{year}{2019}), \bibinfo{pages}{1--38}.
\newblock


\bibitem[Miller(2023)]%
        {miller2023explainable}
\bibfield{author}{\bibinfo{person}{Tim Miller}.}
  \bibinfo{year}{2023}\natexlab{}.
\newblock \showarticletitle{Explainable AI is Dead, Long Live Explainable AI!
  Hypothesis-Driven Decision Support Using Evaluative AI}. In
  \bibinfo{booktitle}{\emph{Proceedings of the 2023 ACM Conference on Fairness,
  Accountability, and Transparency}} (Chicago, IL, USA)
  \emph{(\bibinfo{series}{FAccT '23})}. \bibinfo{publisher}{Association for
  Computing Machinery}, \bibinfo{address}{New York, NY, USA},
  \bibinfo{pages}{333–342}.
\newblock
\showISBNx{9798400701924}
\urldef\tempurl%
\url{https://doi.org/10.1145/3593013.3594001}
\showDOI{\tempurl}


\bibitem[Mooney(2000)]%
        {mooney2000integrating}
\bibfield{author}{\bibinfo{person}{Raymond~J Mooney}.}
  \bibinfo{year}{2000}\natexlab{}.
\newblock \showarticletitle{Integrating abduction and induction in machine
  learning}.
\newblock \bibinfo{journal}{\emph{Abduction and Induction: essays on their
  relation and integration}} (\bibinfo{year}{2000}).
\newblock


\bibitem[Morvan et~al\mbox{.}(2008)]%
        {morvan2008simulation}
\bibfield{author}{\bibinfo{person}{Gildas Morvan}, \bibinfo{person}{Daniel
  Dupont}, {and} \bibinfo{person}{Philippe Kubiak}.}
  \bibinfo{year}{2008}\natexlab{}.
\newblock \showarticletitle{A simulation-based model of abduction}. In
  \bibinfo{booktitle}{\emph{ESM'2008}}. \bibinfo{pages}{183--187}.
\newblock


\bibitem[Munzner(2014)]%
        {munzner2014visualization}
\bibfield{author}{\bibinfo{person}{Tamara Munzner}.}
  \bibinfo{year}{2014}\natexlab{}.
\newblock \bibinfo{booktitle}{\emph{Visualization analysis and design}}.
\newblock \bibinfo{publisher}{CRC press}.
\newblock


\bibitem[Nguyen et~al\mbox{.}(2016)]%
        {nguyen2016synthesizing}
\bibfield{author}{\bibinfo{person}{Anh Nguyen}, \bibinfo{person}{Alexey
  Dosovitskiy}, \bibinfo{person}{Jason Yosinski}, \bibinfo{person}{Thomas
  Brox}, {and} \bibinfo{person}{Jeff Clune}.} \bibinfo{year}{2016}\natexlab{}.
\newblock \showarticletitle{Synthesizing the preferred inputs for neurons in
  neural networks via deep generator networks}.
\newblock \bibinfo{journal}{\emph{Advances in neural information processing
  systems}}  \bibinfo{volume}{29} (\bibinfo{year}{2016}).
\newblock


\bibitem[Nocedal and Wright(2006)]%
        {nocedal2006numerical}
\bibfield{author}{\bibinfo{person}{Jorge Nocedal} {and}
  \bibinfo{person}{Stephen~J Wright}.} \bibinfo{year}{2006}\natexlab{}.
\newblock \bibinfo{booktitle}{\emph{Numerical optimization}}.
\newblock \bibinfo{publisher}{Springer}.
\newblock


\bibitem[Olah et~al\mbox{.}(2017)]%
        {olah2017feature}
\bibfield{author}{\bibinfo{person}{Chris Olah}, \bibinfo{person}{Alexander
  Mordvintsev}, {and} \bibinfo{person}{Ludwig Schubert}.}
  \bibinfo{year}{2017}\natexlab{}.
\newblock \showarticletitle{Feature visualization}.
\newblock \bibinfo{journal}{\emph{Distill}} \bibinfo{volume}{2},
  \bibinfo{number}{11} (\bibinfo{year}{2017}), \bibinfo{pages}{e7}.
\newblock


\bibitem[Oswald and Grosjean(2004)]%
        {oswald2004confirmation}
\bibfield{author}{\bibinfo{person}{Margit~E Oswald} {and}
  \bibinfo{person}{Stefan Grosjean}.} \bibinfo{year}{2004}\natexlab{}.
\newblock \showarticletitle{Confirmation bias}.
\newblock \bibinfo{journal}{\emph{Cognitive illusions: A handbook on fallacies
  and biases in thinking, judgement and memory}}  \bibinfo{volume}{79}
  (\bibinfo{year}{2004}), \bibinfo{pages}{83}.
\newblock


\bibitem[Patel et~al\mbox{.}(2005)]%
        {patel2005thinking}
\bibfield{author}{\bibinfo{person}{Vimla~L Patel}, \bibinfo{person}{Jos{\'e}~F
  Arocha}, {and} \bibinfo{person}{Jiajie Zhang}.}
  \bibinfo{year}{2005}\natexlab{}.
\newblock \showarticletitle{Thinking and reasoning in medicine}.
\newblock \bibinfo{journal}{\emph{The Cambridge handbook of thinking and
  reasoning}}  \bibinfo{volume}{14} (\bibinfo{year}{2005}).
\newblock


\bibitem[Pearl(2009)]%
        {pearl2009causality}
\bibfield{author}{\bibinfo{person}{Judea Pearl}.}
  \bibinfo{year}{2009}\natexlab{}.
\newblock \bibinfo{booktitle}{\emph{Causality}}.
\newblock \bibinfo{publisher}{Cambridge university press}.
\newblock


\bibitem[Peirce(1903)]%
        {peirce1903harvard}
\bibfield{author}{\bibinfo{person}{Charles~S Peirce}.}
  \bibinfo{year}{1903}\natexlab{}.
\newblock \showarticletitle{Harvard lectures on pragmatism}.
\newblock \bibinfo{journal}{\emph{Collected Papers}}  \bibinfo{volume}{5}
  (\bibinfo{year}{1903}), \bibinfo{pages}{188--189}.
\newblock


\bibitem[Peirce and Eisele(1976)]%
        {peirce1976new}
\bibfield{author}{\bibinfo{person}{Charles~Sanders Peirce} {and}
  \bibinfo{person}{Carolyn Eisele}.} \bibinfo{year}{1976}\natexlab{}.
\newblock \bibinfo{booktitle}{\emph{The new elements of mathematics}}.
  Vol.~\bibinfo{volume}{4}.
\newblock \bibinfo{publisher}{Mouton The Hague}.
\newblock


\bibitem[Pople(1973)]%
        {pople1973mechanization}
\bibfield{author}{\bibinfo{person}{Harry~E Pople}.}
  \bibinfo{year}{1973}\natexlab{}.
\newblock \showarticletitle{On the mechanization of abductive logic.}. In
  \bibinfo{booktitle}{\emph{IJCAI}}, Vol.~\bibinfo{volume}{73}. Citeseer,
  \bibinfo{pages}{147--152}.
\newblock


\bibitem[Popper(2014)]%
        {popper2014conjectures}
\bibfield{author}{\bibinfo{person}{Karl Popper}.}
  \bibinfo{year}{2014}\natexlab{}.
\newblock \bibinfo{booktitle}{\emph{Conjectures and refutations: The growth of
  scientific knowledge}}.
\newblock \bibinfo{publisher}{routledge}.
\newblock


\bibitem[Poursabzi-Sangdeh et~al\mbox{.}(2021)]%
        {poursabzi2021manipulating}
\bibfield{author}{\bibinfo{person}{Forough Poursabzi-Sangdeh},
  \bibinfo{person}{Daniel~G Goldstein}, \bibinfo{person}{Jake~M Hofman},
  \bibinfo{person}{Jennifer~Wortman Wortman~Vaughan}, {and}
  \bibinfo{person}{Hanna Wallach}.} \bibinfo{year}{2021}\natexlab{}.
\newblock \showarticletitle{Manipulating and measuring model interpretability}.
  In \bibinfo{booktitle}{\emph{Proceedings of the 2021 CHI conference on human
  factors in computing systems}}. \bibinfo{pages}{1--52}.
\newblock


\bibitem[Radford et~al\mbox{.}(2021)]%
        {radford2021learning}
\bibfield{author}{\bibinfo{person}{Alec Radford}, \bibinfo{person}{Jong~Wook
  Kim}, \bibinfo{person}{Chris Hallacy}, \bibinfo{person}{Aditya Ramesh},
  \bibinfo{person}{Gabriel Goh}, \bibinfo{person}{Sandhini Agarwal},
  \bibinfo{person}{Girish Sastry}, \bibinfo{person}{Amanda Askell},
  \bibinfo{person}{Pamela Mishkin}, \bibinfo{person}{Jack Clark},
  {et~al\mbox{.}}} \bibinfo{year}{2021}\natexlab{}.
\newblock \showarticletitle{Learning transferable visual models from natural
  language supervision}. In \bibinfo{booktitle}{\emph{International Conference
  on Machine Learning}}. PMLR, \bibinfo{pages}{8748--8763}.
\newblock


\bibitem[Rajani et~al\mbox{.}(2019)]%
        {rajani2019explain}
\bibfield{author}{\bibinfo{person}{Nazneen~Fatema Rajani},
  \bibinfo{person}{Bryan McCann}, \bibinfo{person}{Caiming Xiong}, {and}
  \bibinfo{person}{Richard Socher}.} \bibinfo{year}{2019}\natexlab{}.
\newblock \showarticletitle{Explain yourself! leveraging language models for
  commonsense reasoning}.
\newblock \bibinfo{journal}{\emph{arXiv preprint arXiv:1906.02361}}
  (\bibinfo{year}{2019}).
\newblock


\bibitem[Raza et~al\mbox{.}(2022)]%
        {raza2022designing}
\bibfield{author}{\bibinfo{person}{Ali Raza}, \bibinfo{person}{Kim~Phuc Tran},
  \bibinfo{person}{Ludovic Koehl}, {and} \bibinfo{person}{Shujun Li}.}
  \bibinfo{year}{2022}\natexlab{}.
\newblock \showarticletitle{Designing ecg monitoring healthcare system with
  federated transfer learning and explainable ai}.
\newblock \bibinfo{journal}{\emph{Knowledge-Based Systems}}
  \bibinfo{volume}{236} (\bibinfo{year}{2022}), \bibinfo{pages}{107763}.
\newblock


\bibitem[Ren et~al\mbox{.}(2022)]%
        {ren2022deep}
\bibfield{author}{\bibinfo{person}{Zhao Ren}, \bibinfo{person}{Kun Qian},
  \bibinfo{person}{Fengquan Dong}, \bibinfo{person}{Zhenyu Dai},
  \bibinfo{person}{Wolfgang Nejdl}, \bibinfo{person}{Yoshiharu Yamamoto}, {and}
  \bibinfo{person}{Bj{\"o}rn~W Schuller}.} \bibinfo{year}{2022}\natexlab{}.
\newblock \showarticletitle{Deep attention-based neural networks for
  explainable heart sound classification}.
\newblock \bibinfo{journal}{\emph{Machine Learning with Applications}}
  (\bibinfo{year}{2022}), \bibinfo{pages}{100322}.
\newblock


\bibitem[Renna et~al\mbox{.}(2019)]%
        {renna2019deep}
\bibfield{author}{\bibinfo{person}{Francesco Renna}, \bibinfo{person}{Jorge
  Oliveira}, {and} \bibinfo{person}{Miguel~T Coimbra}.}
  \bibinfo{year}{2019}\natexlab{}.
\newblock \showarticletitle{Deep convolutional neural networks for heart sound
  segmentation}.
\newblock \bibinfo{journal}{\emph{IEEE journal of biomedical and health
  informatics}} \bibinfo{volume}{23}, \bibinfo{number}{6}
  (\bibinfo{year}{2019}), \bibinfo{pages}{2435--2445}.
\newblock


\bibitem[Ribeiro et~al\mbox{.}(2016)]%
        {ribeiro2016should}
\bibfield{author}{\bibinfo{person}{Marco~Tulio Ribeiro},
  \bibinfo{person}{Sameer Singh}, {and} \bibinfo{person}{Carlos Guestrin}.}
  \bibinfo{year}{2016}\natexlab{}.
\newblock \showarticletitle{" Why should i trust you?" Explaining the
  predictions of any classifier}. In \bibinfo{booktitle}{\emph{Proceedings of
  the 22nd ACM SIGKDD international conference on knowledge discovery and data
  mining}}. \bibinfo{pages}{1135--1144}.
\newblock


\bibitem[Ribeiro et~al\mbox{.}(2018)]%
        {ribeiro2018anchors}
\bibfield{author}{\bibinfo{person}{Marco~Tulio Ribeiro},
  \bibinfo{person}{Sameer Singh}, {and} \bibinfo{person}{Carlos Guestrin}.}
  \bibinfo{year}{2018}\natexlab{}.
\newblock \showarticletitle{Anchors: High-precision model-agnostic
  explanations}. In \bibinfo{booktitle}{\emph{Proceedings of the AAAI
  conference on artificial intelligence}}, Vol.~\bibinfo{volume}{32}.
\newblock


\bibitem[Riegel et~al\mbox{.}(2020)]%
        {riegel2020logical}
\bibfield{author}{\bibinfo{person}{Ryan Riegel}, \bibinfo{person}{Alexander
  Gray}, \bibinfo{person}{Francois Luus}, \bibinfo{person}{Naweed Khan},
  \bibinfo{person}{Ndivhuwo Makondo}, \bibinfo{person}{Ismail~Yunus Akhalwaya},
  \bibinfo{person}{Haifeng Qian}, \bibinfo{person}{Ronald Fagin},
  \bibinfo{person}{Francisco Barahona}, \bibinfo{person}{Udit Sharma},
  {et~al\mbox{.}}} \bibinfo{year}{2020}\natexlab{}.
\newblock \showarticletitle{Logical neural networks}.
\newblock \bibinfo{journal}{\emph{arXiv preprint arXiv:2006.13155}}
  (\bibinfo{year}{2020}).
\newblock


\bibitem[Ronneberger et~al\mbox{.}(2015)]%
        {ronneberger2015unet}
\bibfield{author}{\bibinfo{person}{Olaf Ronneberger}, \bibinfo{person}{Philipp
  Fischer}, {and} \bibinfo{person}{Thomas Brox}.}
  \bibinfo{year}{2015}\natexlab{}.
\newblock \showarticletitle{U-Net: Convolutional networks for biomedical image
  segmentation}. In \bibinfo{booktitle}{\emph{International Conference on
  Medical image computing and computer-assisted intervention}}. Springer,
  \bibinfo{pages}{234--241}.
\newblock


\bibitem[Rosenthal et~al\mbox{.}(2016)]%
        {rosenthal2016verbalization}
\bibfield{author}{\bibinfo{person}{Stephanie Rosenthal}, \bibinfo{person}{Sai~P
  Selvaraj}, {and} \bibinfo{person}{Manuela~M Veloso}.}
  \bibinfo{year}{2016}\natexlab{}.
\newblock \showarticletitle{Verbalization: Narration of Autonomous Robot
  Experience.}. In \bibinfo{booktitle}{\emph{IJCAI}},
  Vol.~\bibinfo{volume}{16}. \bibinfo{pages}{862--868}.
\newblock


\bibitem[Ross et~al\mbox{.}(2017)]%
        {ross2017right}
\bibfield{author}{\bibinfo{person}{Andrew~Slavin Ross},
  \bibinfo{person}{Michael~C Hughes}, {and} \bibinfo{person}{Finale
  Doshi-Velez}.} \bibinfo{year}{2017}\natexlab{}.
\newblock \showarticletitle{Right for the right reasons: training
  differentiable models by constraining their explanations}. In
  \bibinfo{booktitle}{\emph{Proceedings of the 26th International Joint
  Conference on Artificial Intelligence}}. \bibinfo{pages}{2662--2670}.
\newblock


\bibitem[Rubin et~al\mbox{.}(2017)]%
        {rubin2017recognizing}
\bibfield{author}{\bibinfo{person}{Jonathan Rubin}, \bibinfo{person}{Rui
  Abreu}, \bibinfo{person}{Anurag Ganguli}, \bibinfo{person}{Saigopal
  Nelaturi}, \bibinfo{person}{Ion Matei}, {and} \bibinfo{person}{Kumar
  Sricharan}.} \bibinfo{year}{2017}\natexlab{}.
\newblock \showarticletitle{Recognizing abnormal heart sounds using deep
  learning}.
\newblock \bibinfo{journal}{\emph{arXiv preprint arXiv:1707.04642}}
  (\bibinfo{year}{2017}).
\newblock


\bibitem[Rudin(2019)]%
        {rudin2019stop}
\bibfield{author}{\bibinfo{person}{Cynthia Rudin}.}
  \bibinfo{year}{2019}\natexlab{}.
\newblock \showarticletitle{Stop explaining black box machine learning models
  for high stakes decisions and use interpretable models instead}.
\newblock \bibinfo{journal}{\emph{Nature Machine Intelligence}}
  \bibinfo{volume}{1}, \bibinfo{number}{5} (\bibinfo{year}{2019}),
  \bibinfo{pages}{206--215}.
\newblock


\bibitem[Selvaraju et~al\mbox{.}(2017)]%
        {selvaraju2017grad}
\bibfield{author}{\bibinfo{person}{Ramprasaath~R Selvaraju},
  \bibinfo{person}{Michael Cogswell}, \bibinfo{person}{Abhishek Das},
  \bibinfo{person}{Ramakrishna Vedantam}, \bibinfo{person}{Devi Parikh}, {and}
  \bibinfo{person}{Dhruv Batra}.} \bibinfo{year}{2017}\natexlab{}.
\newblock \showarticletitle{Grad-cam: Visual explanations from deep networks
  via gradient-based localization}. In \bibinfo{booktitle}{\emph{Proceedings of
  the IEEE international conference on computer vision}}.
  \bibinfo{pages}{618--626}.
\newblock


\bibitem[Shimojima(1999)]%
        {shimojima1999graphic}
\bibfield{author}{\bibinfo{person}{Atsushi Shimojima}.}
  \bibinfo{year}{1999}\natexlab{}.
\newblock \showarticletitle{The graphic-linguistic distinction exploring
  alternatives}.
\newblock \bibinfo{journal}{\emph{Artificial Intelligence Review}}
  \bibinfo{volume}{13}, \bibinfo{number}{4} (\bibinfo{year}{1999}),
  \bibinfo{pages}{313--335}.
\newblock


\bibitem[Simonyan et~al\mbox{.}(2014)]%
        {simonyan2014deep}
\bibfield{author}{\bibinfo{person}{Karen Simonyan}, \bibinfo{person}{Andrea
  Vedaldi}, {and} \bibinfo{person}{Andrew Zisserman}.}
  \bibinfo{year}{2014}\natexlab{}.
\newblock \showarticletitle{Deep inside convolutional networks: Visualising
  image classification models and saliency maps}.
\newblock  (\bibinfo{year}{2014}).
\newblock


\bibitem[Siontis et~al\mbox{.}(2021)]%
        {siontis2021artificial}
\bibfield{author}{\bibinfo{person}{Konstantinos~C Siontis},
  \bibinfo{person}{Peter~A Noseworthy}, \bibinfo{person}{Zachi~I Attia}, {and}
  \bibinfo{person}{Paul~A Friedman}.} \bibinfo{year}{2021}\natexlab{}.
\newblock \showarticletitle{Artificial intelligence-enhanced
  electrocardiography in cardiovascular disease management}.
\newblock \bibinfo{journal}{\emph{Nature Reviews Cardiology}}
  \bibinfo{volume}{18}, \bibinfo{number}{7} (\bibinfo{year}{2021}),
  \bibinfo{pages}{465--478}.
\newblock


\bibitem[Teijeiro et~al\mbox{.}(2016)]%
        {teijeiro2016heartbeat}
\bibfield{author}{\bibinfo{person}{Tom{\'a}s Teijeiro}, \bibinfo{person}{Paulo
  F{\'e}lix}, \bibinfo{person}{Jes{\'u}s Presedo}, {and}
  \bibinfo{person}{Daniel Castro}.} \bibinfo{year}{2016}\natexlab{}.
\newblock \showarticletitle{Heartbeat classification using abstract features
  from the abductive interpretation of the ECG}.
\newblock \bibinfo{journal}{\emph{IEEE journal of biomedical and health
  informatics}} \bibinfo{volume}{22}, \bibinfo{number}{2}
  (\bibinfo{year}{2016}), \bibinfo{pages}{409--420}.
\newblock


\bibitem[Teijeiro et~al\mbox{.}(2018)]%
        {teijeiro2018abductive}
\bibfield{author}{\bibinfo{person}{Tom{\'a}s Teijeiro},
  \bibinfo{person}{Constantino~A Garc{\'\i}a}, \bibinfo{person}{Daniel Castro},
  {and} \bibinfo{person}{Paulo F{\'e}lix}.} \bibinfo{year}{2018}\natexlab{}.
\newblock \showarticletitle{Abductive reasoning as a basis to reproduce expert
  criteria in ECG atrial fibrillation identification}.
\newblock \bibinfo{journal}{\emph{Physiological measurement}}
  \bibinfo{volume}{39}, \bibinfo{number}{8} (\bibinfo{year}{2018}),
  \bibinfo{pages}{084006}.
\newblock


\bibitem[Tjoa and Guan(2020)]%
        {tjoa2020survey}
\bibfield{author}{\bibinfo{person}{Erico Tjoa} {and} \bibinfo{person}{Cuntai
  Guan}.} \bibinfo{year}{2020}\natexlab{}.
\newblock \showarticletitle{A survey on explainable artificial intelligence
  (xai): Toward medical xai}.
\newblock \bibinfo{journal}{\emph{IEEE transactions on neural networks and
  learning systems}} \bibinfo{volume}{32}, \bibinfo{number}{11}
  (\bibinfo{year}{2020}), \bibinfo{pages}{4793--4813}.
\newblock


\bibitem[Van~der Maaten and Hinton(2008)]%
        {van2008visualizing}
\bibfield{author}{\bibinfo{person}{Laurens Van~der Maaten} {and}
  \bibinfo{person}{Geoffrey Hinton}.} \bibinfo{year}{2008}\natexlab{}.
\newblock \showarticletitle{Visualizing data using t-SNE.}
\newblock \bibinfo{journal}{\emph{Journal of machine learning research}}
  \bibinfo{volume}{9}, \bibinfo{number}{11} (\bibinfo{year}{2008}).
\newblock


\bibitem[Veale et~al\mbox{.}(2018)]%
        {veale2018fairness}
\bibfield{author}{\bibinfo{person}{Michael Veale}, \bibinfo{person}{Max
  Van~Kleek}, {and} \bibinfo{person}{Reuben Binns}.}
  \bibinfo{year}{2018}\natexlab{}.
\newblock \showarticletitle{Fairness and accountability design needs for
  algorithmic support in high-stakes public sector decision-making}. In
  \bibinfo{booktitle}{\emph{Proceedings of the 2018 chi conference on human
  factors in computing systems}}. \bibinfo{pages}{1--14}.
\newblock


\bibitem[Vellido(2020)]%
        {vellido2020importance}
\bibfield{author}{\bibinfo{person}{Alfredo Vellido}.}
  \bibinfo{year}{2020}\natexlab{}.
\newblock \showarticletitle{The importance of interpretability and
  visualization in machine learning for applications in medicine and health
  care}.
\newblock \bibinfo{journal}{\emph{Neural computing and applications}}
  \bibinfo{volume}{32}, \bibinfo{number}{24} (\bibinfo{year}{2020}),
  \bibinfo{pages}{18069--18083}.
\newblock


\bibitem[Wachter et~al\mbox{.}(2017)]%
        {wachter2017counterfactual}
\bibfield{author}{\bibinfo{person}{Sandra Wachter}, \bibinfo{person}{Brent
  Mittelstadt}, {and} \bibinfo{person}{Chris Russell}.}
  \bibinfo{year}{2017}\natexlab{}.
\newblock \showarticletitle{Counterfactual explanations without opening the
  black box: Automated decisions and the GDPR}.
\newblock \bibinfo{journal}{\emph{Harv. JL \& Tech.}}  \bibinfo{volume}{31}
  (\bibinfo{year}{2017}), \bibinfo{pages}{841}.
\newblock


\bibitem[Wang et~al\mbox{.}(2019)]%
        {wang2019designing}
\bibfield{author}{\bibinfo{person}{Danding Wang}, \bibinfo{person}{Qian Yang},
  \bibinfo{person}{Ashraf Abdul}, {and} \bibinfo{person}{Brian~Y Lim}.}
  \bibinfo{year}{2019}\natexlab{}.
\newblock \showarticletitle{Designing theory-driven user-centric explainable
  AI}. In \bibinfo{booktitle}{\emph{Proceedings of the 2019 CHI conference on
  human factors in computing systems}}. \bibinfo{pages}{1--15}.
\newblock


\bibitem[Wang et~al\mbox{.}(2021)]%
        {wang2021show}
\bibfield{author}{\bibinfo{person}{Danding Wang}, \bibinfo{person}{Wencan
  Zhang}, {and} \bibinfo{person}{Brian~Y Lim}.}
  \bibinfo{year}{2021}\natexlab{}.
\newblock \showarticletitle{Show or suppress? Managing input uncertainty in
  machine learning model explanations}.
\newblock \bibinfo{journal}{\emph{Artificial Intelligence}}
  \bibinfo{volume}{294} (\bibinfo{year}{2021}), \bibinfo{pages}{103456}.
\newblock


\bibitem[Wang et~al\mbox{.}(2022)]%
        {wang2022interpretable}
\bibfield{author}{\bibinfo{person}{Yunlong Wang},
  \bibinfo{person}{Priyadarshini Venkatesh}, {and} \bibinfo{person}{Brian~Y
  Lim}.} \bibinfo{year}{2022}\natexlab{}.
\newblock \showarticletitle{Interpretable Directed Diversity: Leveraging Model
  Explanations for Iterative Crowd Ideation}. In \bibinfo{booktitle}{\emph{CHI
  Conference on Human Factors in Computing Systems}}. \bibinfo{pages}{1--28}.
\newblock


\bibitem[Williamson(2016)]%
        {williamson2016abductive}
\bibfield{author}{\bibinfo{person}{Timothy Williamson}.}
  \bibinfo{year}{2016}\natexlab{}.
\newblock \showarticletitle{Abductive philosophy}. In
  \bibinfo{booktitle}{\emph{The Philosophical Forum}},
  Vol.~\bibinfo{volume}{47}. Wiley Online Library, \bibinfo{pages}{263--280}.
\newblock


\bibitem[Woods et~al\mbox{.}(2019)]%
        {woods2019adversarial}
\bibfield{author}{\bibinfo{person}{Walt Woods}, \bibinfo{person}{Jack Chen},
  {and} \bibinfo{person}{Christof Teuscher}.} \bibinfo{year}{2019}\natexlab{}.
\newblock \showarticletitle{Adversarial explanations for understanding image
  classification decisions and improved neural network robustness}.
\newblock \bibinfo{journal}{\emph{Nature Machine Intelligence}}
  \bibinfo{volume}{1}, \bibinfo{number}{11} (\bibinfo{year}{2019}),
  \bibinfo{pages}{508--516}.
\newblock


\bibitem[{World Health Organization}(2021)]%
        {who2021cardiovascular}
\bibfield{author}{\bibinfo{person}{{World Health Organization}}.}
  \bibinfo{year}{2021}\natexlab{}.
\newblock \bibinfo{title}{Cardiovascular diseases (CVDs) fact sheet}.
\newblock
  \bibinfo{howpublished}{\url{https://www.who.int/news-room/fact-sheets/detail/cardiovascular-diseases-(cvds)}}.
\newblock
\newblock
\shownote{[Online; accessed 14-Sep-2024]}.


\bibitem[Wu et~al\mbox{.}(2018)]%
        {wu2018beyond}
\bibfield{author}{\bibinfo{person}{Mike Wu}, \bibinfo{person}{Michael Hughes},
  \bibinfo{person}{Sonali Parbhoo}, \bibinfo{person}{Maurizio Zazzi},
  \bibinfo{person}{Volker Roth}, {and} \bibinfo{person}{Finale Doshi-Velez}.}
  \bibinfo{year}{2018}\natexlab{}.
\newblock \showarticletitle{Beyond sparsity: Tree regularization of deep models
  for interpretability}. In \bibinfo{booktitle}{\emph{Proceedings of the AAAI
  conference on artificial intelligence}}, Vol.~\bibinfo{volume}{32}.
\newblock


\bibitem[Yaseen et~al\mbox{.}(2018)]%
        {yaseen2018classification}
\bibfield{author}{\bibinfo{person}{Yaseen}, \bibinfo{person}{Gui-Young Son},
  {and} \bibinfo{person}{Soonil Kwon}.} \bibinfo{year}{2018}\natexlab{}.
\newblock \showarticletitle{Classification of heart sound signal using multiple
  features}.
\newblock \bibinfo{journal}{\emph{Applied Sciences}} \bibinfo{volume}{8},
  \bibinfo{number}{12} (\bibinfo{year}{2018}), \bibinfo{pages}{2344}.
\newblock


\bibitem[Ye et~al\mbox{.}(2020)]%
        {ye2020user}
\bibfield{author}{\bibinfo{person}{Yucong Ye}, \bibinfo{person}{Franz Sauer},
  \bibinfo{person}{Kwan-Liu Ma}, \bibinfo{person}{Konduri Aditya}, {and}
  \bibinfo{person}{Jacqueline Chen}.} \bibinfo{year}{2020}\natexlab{}.
\newblock \showarticletitle{A user-centered design study in scientific
  visualization targeting domain experts}.
\newblock \bibinfo{journal}{\emph{IEEE Transactions on Visualization and
  Computer Graphics}} \bibinfo{volume}{26}, \bibinfo{number}{6}
  (\bibinfo{year}{2020}).
\newblock


\bibitem[Zhang and Banovic(2021)]%
        {zhang2021method}
\bibfield{author}{\bibinfo{person}{Enhao Zhang} {and} \bibinfo{person}{Nikola
  Banovic}.} \bibinfo{year}{2021}\natexlab{}.
\newblock \showarticletitle{Method for Exploring Generative Adversarial
  Networks (GANs) via Automatically Generated Image Galleries}. In
  \bibinfo{booktitle}{\emph{Proceedings of the 2021 CHI Conference on Human
  Factors in Computing Systems}}.
\newblock


\bibitem[Zhang and Lim(2022)]%
        {zhang2022towards}
\bibfield{author}{\bibinfo{person}{Wencan Zhang} {and} \bibinfo{person}{Brian~Y
  Lim}.} \bibinfo{year}{2022}\natexlab{}.
\newblock \showarticletitle{Towards Relatable Explainable AI with the
  Perceptual Process}. In \bibinfo{booktitle}{\emph{CHI Conference on Human
  Factors in Computing Systems}}. \bibinfo{pages}{1--24}.
\newblock


\bibitem[Zhou et~al\mbox{.}(2016)]%
        {zhou2016learning}
\bibfield{author}{\bibinfo{person}{Bolei Zhou}, \bibinfo{person}{Aditya
  Khosla}, \bibinfo{person}{Agata Lapedriza}, \bibinfo{person}{Aude Oliva},
  {and} \bibinfo{person}{Antonio Torralba}.} \bibinfo{year}{2016}\natexlab{}.
\newblock \showarticletitle{Learning deep features for discriminative
  localization}. In \bibinfo{booktitle}{\emph{Proceedings of the IEEE
  conference on computer vision and pattern recognition}}.
  \bibinfo{pages}{2921--2929}.
\newblock


\bibitem[Ziou and Tabbone(1998)]%
        {ziou1998edge}
\bibfield{author}{\bibinfo{person}{Djemel Ziou} {and}
  \bibinfo{person}{Salvatore Tabbone}.} \bibinfo{year}{1998}\natexlab{}.
\newblock \showarticletitle{Edge detection techniques-an overview}.
\newblock \bibinfo{journal}{\emph{Pattern Recognition and Image Analysis:
  Advances in Mathematical Theory and Applications}} \bibinfo{volume}{8},
  \bibinfo{number}{4} (\bibinfo{year}{1998}), \bibinfo{pages}{537--559}.
\newblock


\bibitem[Zytek et~al\mbox{.}(2022)]%
        {zytek2022need}
\bibfield{author}{\bibinfo{person}{Alexandra Zytek}, \bibinfo{person}{Ignacio
  Arnaldo}, \bibinfo{person}{Dongyu Liu}, \bibinfo{person}{Laure
  Berti-Equille}, {and} \bibinfo{person}{Kalyan Veeramachaneni}.}
  \bibinfo{year}{2022}\natexlab{}.
\newblock \showarticletitle{The need for interpretable features: Motivation and
  taxonomy}.
\newblock \bibinfo{journal}{\emph{ACM SIGKDD Explorations Newsletter}}
  \bibinfo{volume}{24}, \bibinfo{number}{1} (\bibinfo{year}{2022}),
  \bibinfo{pages}{1--13}.
\newblock


\end{thebibliography}

\clearpage

\appendix
\onecolumn


\section{Appendix}

\subsection{Demonstration study}

\subsubsection{Contrastive explanation details}
\phantom{...}

\begin{table}[h!]
\small
\centering
\caption{
    Predicted parameter values ($\tau$'s, $\pi$'s), murmur shape fits (MSE $d$), and labels ($y$'s) for each diagnosis, for the example MVP instance in Fig. \ref{fig:demo-mvp}.
    {\color[HTML]{C0C0C0}Grey} text indicate redundant parameters.
    {\color[HTML]{C0C0C0}$\varnothing$} indicates no parameter for that specific diagnosis.
    \textbf{Bold} numbers indicate evidence to predict towards the row's diagnosis.
    $\tau_4$ for the MS diagnosis is redundant, since $\tau_4 = \tau_L$; this indicates that the MS shape overfits to the data, both MVP and MS have the same MSE, and the MVP shape is sufficient. 
    Thus, the model predicts MVP as the sufficient hypothesis for this instance.
}
\label{table:demo-mvp-predictions}
\begin{tabular}{rlccccccccccccccc}
\hline
\multicolumn{1}{l}{} &  & \multicolumn{5}{c}{Time Parameters (sec)}                                                                                                                                                        &  & \multicolumn{3}{c}{Slope Parameters}                                                                               &  & Shape MSE          &  & \multicolumn{3}{c}{Predictions}                    \\ \cline{3-7} \cline{9-11} \cline{13-13} \cline{15-17} 
Diagnosis            &  & $\tau_1$                             & $\tau_2$                             & $\tau_3$                             & $\tau_4$                             & $\tau_L$                             &  & $\pi_0$                              & $\pi_1$                              & $\pi_2$                              &  & $d$ (1e$-$4) &  & $y_0$           & $y_r$          & $y$             \\ \cline{1-1} \cline{3-7} \cline{9-11} \cline{13-13} \cline{15-17} 
N                    &  & {\color[HTML]{C0C0C0} 0.36} & {\color[HTML]{C0C0C0} $\varnothing$} & {\color[HTML]{C0C0C0} $\varnothing$} & {\color[HTML]{C0C0C0} $\varnothing$} & {\color[HTML]{C0C0C0} 0.56} &  & 0 & {\color[HTML]{C0C0C0} $\varnothing$} & {\color[HTML]{C0C0C0} $\varnothing$} &  & 75          &  & 0.0000          & 0.000          & 0.000           \\
AS                   &  & 0.36                                 & 0.38                                 & {\color[HTML]{C0C0C0} $\varnothing$} & {\color[HTML]{C0C0C0} $\varnothing$} & 0.56                                 &  & $-$0.03                              & 27.0                                 & 1.8                                  &  & 25          &  & 0.0000          & 0.000          & 0.000           \\
MR                   &  & 0.36                                 & {\color[HTML]{C0C0C0} $\varnothing$} & {\color[HTML]{C0C0C0} $\varnothing$} & {\color[HTML]{C0C0C0} $\varnothing$} & 0.56                                 &  & {\color[HTML]{FFFFFF} $+$}0.13                                 & {\color[HTML]{C0C0C0} $\varnothing$} & {\color[HTML]{C0C0C0} $\varnothing$} &  & 43          &  & 0.0000          & 0.000          & 0.000           \\
\textbf{MVP}                  &  & 0.36                                 & 0.39                                 & 0.42                                 & {\color[HTML]{C0C0C0} $\varnothing$} & 0.56                                 &  & {\color[HTML]{FFFFFF} $+$}0.03                                 & 20.5                                 & {\color[HTML]{C0C0C0} $\varnothing$} &  & \textbf{3.3} &  & \textbf{0.9998} & \textbf{0.952} & \textbf{1.000} \\
MS                   &  & 0.36                                 & 0.39                                 & 0.42                                 & {\color[HTML]{C0C0C0} 0.56}          & 0.56                                 &  & {\color[HTML]{FFFFFF} $+$}0.03                                 & 20.5                                 & 0.9                                  &  & \textbf{3.3} &  & 0.0001          & 0.048          & 0.000           \\ \hline
\end{tabular}
\end{table}

\subsubsection{Counterfactual explanations}
\phantom{...}

\begin{figure}[h]
    \centering
    \includegraphics[width=12.2cm]{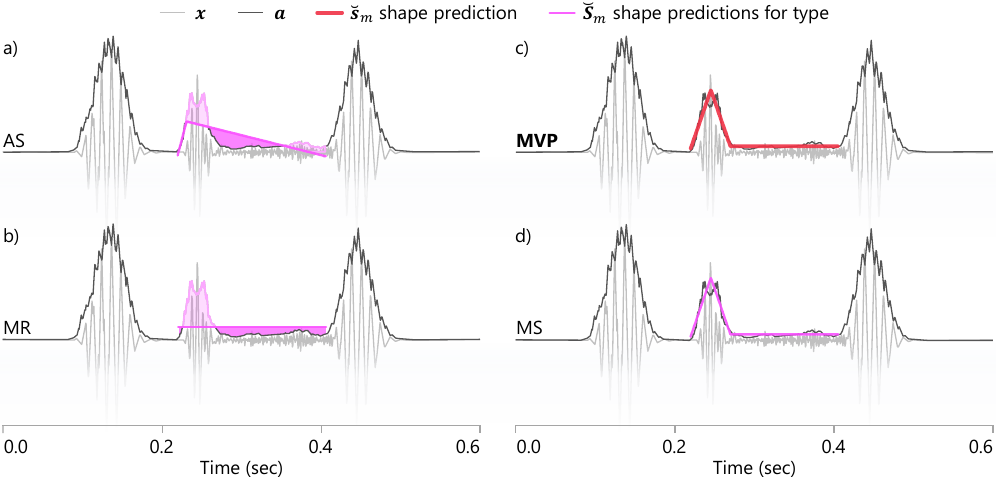}
    \caption{
    Counterfactual explanations for alternative diagnoses of an actual MVP case.
    This augments the contrastive explanation by showing which parts of the murmur amplitude should be higher or lower to match the target diagnosis.
    There are no counterfactual differences for MVP (the prediction), and MS (which overfits to MVP).
    }
    \Description{
    Figure shows four line graphs identical to Figure 7, except that additional annotations show areas of the murmur lines above and below the phonocardiogram amplitude, emphasizing discrepancies between the observed phonocardiogram and murmur shape.
    }
    \label{fig:demo-mvp-counterfactual}
\end{figure}


\subsection{Modeling study}

\subsubsection{Comparison model architectures} \label{subsubsection:comparison-models}
We describe the models compared against our proposed DiagramNet model.

\begin{figure}[h]
    \centering
    \vspace*{-0.05cm}
    \includegraphics[width=5.4cm]{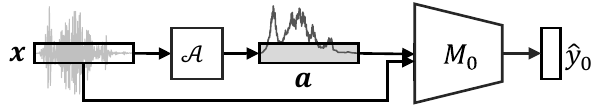}
    \vspace*{-0.05cm}
    \caption{
    Base CNN model that inputs the displacement $\bm{x}$ and amplitude $\bm{a} = \mathscr{A}(\bm{x})$ to predict cardiac diagnosis $\hat{\bm{y}}_0 = M_0(\bm{x},\bm{a})$.
    }
    \Description{
    The figure presents the modular architecture of Base CNN. The architecture (shown as a flowchart) shows how input x goes into the module A to compute amplitude a, and both x and a are input to model M_0 to output baseline diagnosis y_0.
    }
    \label{fig:architecture-base-cnn}
    \vspace{-0.1cm}
\end{figure}

\paragraph{Base prediction model}
We treat each audio time series like a 1D image, since all instances are fixed-length and single-channel. 
We further concatenate displacement $\bm{x}$ and amplitude $\bm{a}$ into a 2-channel "image".
To compare with our full proposed model, we trained a base convolutional neural network (CNN) \cite{hershey2017cnn} as model $M_0$ on $(\bm{x},\bm{a})$ to predict diagnosis $\hat{\bm{y}}_0$ (see Fig. \ref{fig:architecture-base-cnn}).
Although $M_0$ can indicate the probability-like disease risk, this could be spurious, since it does not consider the murmur shapes of each diagnosis.
Instead, a more reliable model should explicitly encode constraints that are domain relevant.

\begin{figure}[h]
    \centering
    \vspace{-0.0cm}
    \includegraphics[width=7.5cm]{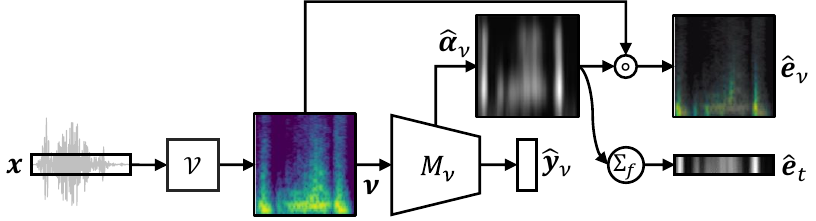}
    \vspace{-0.2cm}
    \caption{
    Alternative CNN model that inputs the mel specrogram $\bm{\nu} = \mathscr{V}(\bm{x})$ of the audio to predict diagnosis $\hat{y}_0 = M_\nu(\bm{\nu})$.
    It generates a saliency map explanation $\hat{\bm{\alpha}}_\nu$ overlaid on $\bm{\nu}$ as a Spectrogram-saliency explanation $\hat{\bm{e}}_\nu$ or aggregated by time as Time-saliency $\hat{\bm{e}}_t$.
    }
    \Description{
    The figure presents the modular architecture of Base CNN (Spectrogram). The architecture (shown as a flowchart) shows how input x goes into the module V to compute spectrogram v, and both x and v are input to model M_v to output baseline diagnosis y_0. Another output from M_v on the side is the saliency map explanation a_v. a_v is processed into two types of explanations: 1) combined with v to produce the saliency-masked spectrogram explanation e_v, and 2) aggregated by frequency to output e_t.
    }
    \label{fig:model-spectrogram}
    \vspace{-0.2cm}
\end{figure}

\paragraph{Alternative prediction model with spectrogram input and saliency map explanation}
To compare our proposed method with current XAI, we implemented a spectrogram-based CNN classifier (Fig. \ref{fig:model-spectrogram}).
Spectrograms are popular to extract features from high-frequency time series data, and have been used to extract features from heart auscultation~\cite{dissanayake2020robust, ren2022deep}.
They show how the frequency (pitch) of the signal (y-axis) changes over time (x-axis), by indicating the magnitude of specific frequency components at each pixel.
We represented each spectrogram as a 3D tensor for (frequency, time, magnitude) with raw magnitude numeric values, rather than as a colored 2D image that would be biased by the color map used.
To capture the temporal motifs of frequency patterns in spectrograms, we trained a CNN to learn convolutional filters that activate if specific spatial patterns are detected.
Specifically, we used the mel spectrogram $\bm{\nu} = \mathscr{V}(\bm{x})$, since it is more sensitive to variations in lower frequencies.

Saliency maps are popular to explain which pixels were important for image-based predictions with CNNs~\cite{simonyan2014deep,selvaraju2017grad,zhou2016learning}. 
For spectrograms, this indicates the frequencies at specific times that the model focused on.
We implemented Grad-CAM~\cite{selvaraju2017grad} to generate saliency explanation $\hat{\bm{e}}_\nu$. 
Despite their popularity, saliency map explanations neglect the interpretability needs of domain experts. 
Specifically, clinicians are not trained on spectrograms, thus we hypothesize that this saliency map, spectrogram-saliency, is less appropriate than the murmur-shape diagrammatic explanation.
Also, similar to \cite{zhang2022towards}, we provide simplified saliency map explanations to show importance by time, time-saliency $\hat{\bm{e}}_t$,
by aggregating all saliency across frequencies $\Sigma_{f}$. 
This is simpler and does not require the user to understand spectrograms or note frequencies.
These explanations were used as baseline comparisons in our qualitative user study.

Despite their use for AI heart auscultation~\cite{dissanayake2020robust, ren2022deep}, we do not advocate explaining with saliency maps on spectrograms. 
Indeed, they are overly technical, non-relatable~\cite{zhang2022towards}, less interpretable, and less trustworthy (as evidenced in our user study). 
Instead, clinical AI explanations need to be abductive and diagrammatic to be more trustworthy.

\paragraph{Other comparison models}
We examined various model architectures as soft classifiers to predict the murmur shapes. These were all evaluated to perform worse than our proposed DiagramNet method.

\begin{figure}[h!]
    \centering
    \vspace{-0.2cm}
    \includegraphics[width=14.0cm]{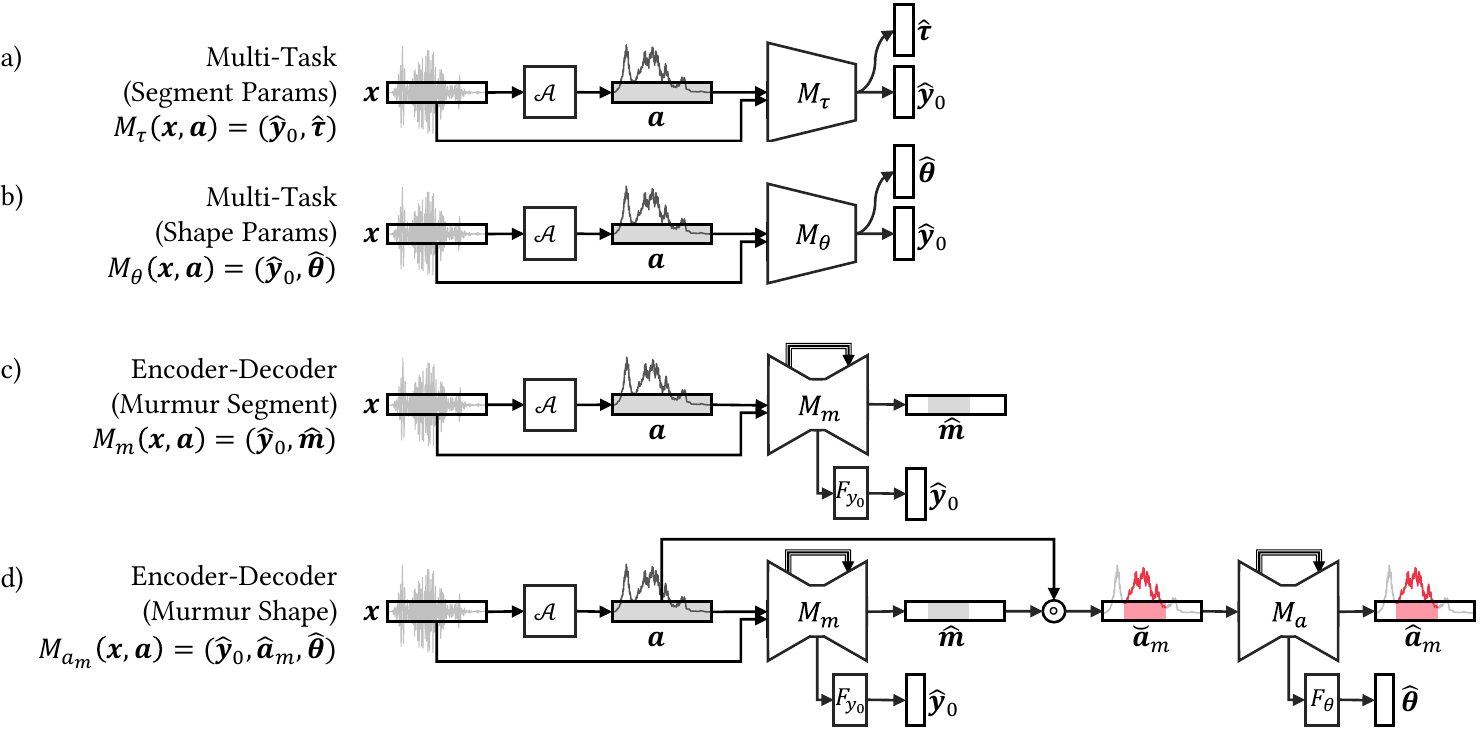}
    \vspace{-0.2cm}
    \caption{
    Architectures of various alternative models that take displacement $\bm{x}$ and amplitude $\bm{a}$ as inputs to 
    predict diagnosis $\hat{\bm{y}}_0$, murmur time parameters $\hat{\bm{\tau}}$, murmur shape parameters $\hat{\bm{\theta}}$, 
    murmur time segment $\hat{\bm{m}}$, and murmur amplitude $\hat{\bm{a}}_m$.    
    }
    \Description{
    The figure consists of four architectures shown as flow charts for a) Multi-Task (Segment Params), b) Multi-Task (Shape Params), c) Encoder-Decoder (Murmur Segment), and d) Encoder-Decoder (Murmur Shape). All architectures start with x input to A to output amplitude a. (a) outputs tau and y_0, (b) outputs theta and y_0. (c) outputs murmur segment m through a U-Net model, and y_0. (d) extends (c) by predicting the murmur amplitude reconstruction a_m and parameters theta using a second U-Net.
    }
    \label{fig:model-others}
\end{figure}

\clearpage

\subsubsection{Performance results} \label{sec:appendix-performance-layout}
\phantom{...}

\begin{figure*}[h!]
    \centering
    \vspace{-0.1cm}
    \includegraphics[width=3.6cm]{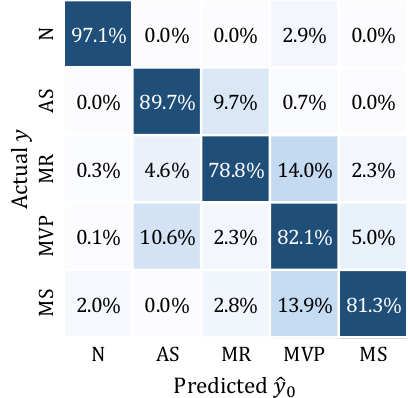}
    \vspace{-0.25cm}
    \caption{
    Confusion matrices of the base CNN showing lower accuracy than predictions from DiagramNet (see Fig. \ref{fig:results-confusion-matrices}).
    }
    \Description{
    Confusion Matrix for the base CNN model. The confusion matrix is a 5x5 grid with rows labeled as "Actual y" and columns labeled as "Predicted y hat subscript 0". The rows and columns represent five classes: N, AS, MR, MVP, and MS. Each cell contains a percentage value indicating the proportion of predictions for each actual class. The diagonal cells, representing correct predictions, have the highest values: 97.1\% for N, 89.7\% for AS, 78.8\% for MR, 82.1\% for MVP, and 81.3\% for MS. Off-diagonal cells show the misclassification rates, including AS as MR: 9.7\%, MR as MVP: 14.0\%, MVP as AS: 10.6\%, MS as MVP: 13.9\%.
    }
    \label{fig:results-confusion-matrix-cnn}
    \vspace{-0.30cm}
\end{figure*}
    
\begin{figure*}[h!]
    \centering
    \includegraphics[width=9.65cm]{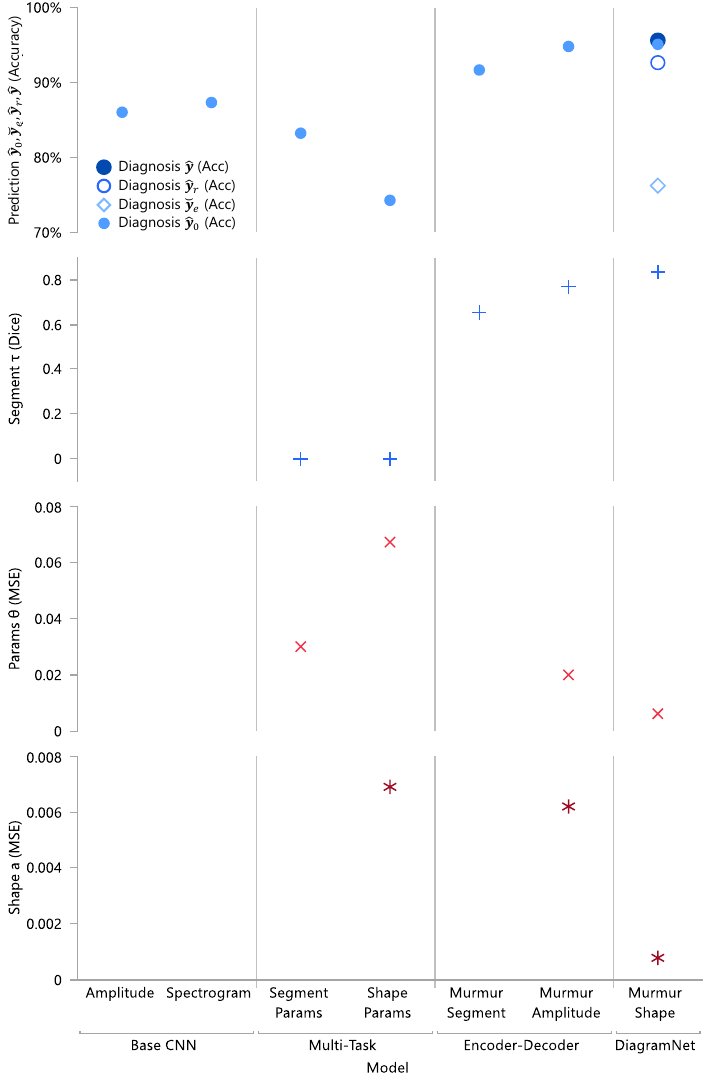}
    \vspace{-0.40cm}
    \caption{
    Same content as Fig. \ref{fig:results-model-performances}. Expanded layout of results from the modeling study comparing DiagramNet with baseline and alternative models.
    Model performance is measured with perceptual $\hat{\bm{y}}_0$, rule-like evaluation-based $\Breve{\bm{y}}_e$, parsimoniously-resolved abductive $\hat{\bm{y}}_r$, and ensemble $\hat{\bm{y}}$ diagnosis accuracy.
    Explanation faithfulness by murmur segment $\hat{\bm{\tau}}$ Dice coefficient, murmur shape parameters $\hat{\bm{\theta}}$, $\tilde{\bm{\theta}}$ MSE, murmur shape $\Breve{\bm{s}}_m$ fit MSE, and reconstructed murmur shape amplitude $\hat{\bm{a}}_m$ MSE.
    For blue metrics, higher is better; for red metrics, lower is better.
    DiagramNet has highest prediction and segmentation accuracy, and lowest estimation error for parameter values and shape fits (all good).
    See Table \ref{table:results-performance} for numeric details.
    }
    \Description{
    The figure of a dot plot comparing the performance of different models on various measures. The figure displays performance metrics for models including Base CNN, Multi-Task, Encoder-Decoder, and DiagramNet on diagnosis accuracy, segment Dice score, parameter mean squared error (MSE), and shape MSE. Each model's performance is represented by symbols and colors: blue symbols for diagnosis accuracy of different models (solid dark blue for DiagramNet combined prediction, outline blue circle for DiagramNet parsimonously-resolved abductive prediction, outline blue diamond for DiagramNet evaluation-based prediction, and small light blue circle for base perceptual prediction), blue plus signs for segment Dice score, red crosses for parameter MSE, and red asterisks for shape MSE. The y-axis represents prediction accuracy and segment Dice score, while the secondary y-axis on the right represents parameter and shape MSE. The x-axis lists models: Base CNN with Amplitude and Spectrogram variants, Multi-Task model with Segment Parameters and Shape Parameters variants, Encoder-Decoder models with Murmur Segment and Murmur Amplitude segments, and DiagramNet with Murmur Shape.
    }
    \label{fig:results-model-performances-expanded}
    \vspace*{-0.35cm}
\end{figure*}

\clearpage

\begin{table}[h!]
\footnotesize
\centering
\caption{
Numeric details of performance results shown in Fig. \ref{fig:results-model-performances}.
\textbf{Bold} numbers indicates the best performance.
}
\vspace{-0.2cm}
\label{table:results-performance}
\renewcommand{\arraystretch}{1.28}
\begin{tabular}{lllcccclccc}
\hline
\multicolumn{2}{l}{} &
   &
  \multicolumn{4}{c}{Prediction Accuracy $\uparrow$} &
   &
  \multicolumn{3}{c}{Explanation Faithfulness} \\ \cline{4-7} \cline{9-11}
\multicolumn{2}{l}{Model} &
   &
  $\hat{\bm{y}}_0$ &
  $\Breve{\bm{y}}_e$ &
  $\hat{\bm{y}}_r$ &
  $\hat{\bm{y}}$ &
   &
  \begin{tabular}[c]{@{}c@{}}Segment $\hat{\bm{\tau}}$\\ (Dice) $\uparrow$\end{tabular} &
  \begin{tabular}[c]{@{}c@{}}Params $\hat{\bm{\theta}}$\\ (MSE) $\downarrow$\end{tabular} &
  \begin{tabular}[c]{@{}c@{}}Shape $\hat{\bm{a}}_m$, $\Breve{\bm{s}}_m$\\ (MSE) $\downarrow$\end{tabular} \\ \cline{1-2} \cline{4-7} \cline{9-11}
Base CNN &
  $M_0(\bm{x},\bm{a})=\hat{\bm{y}}_0$ &
   &
  86.0\% &
  {\color[HTML]{C0C0C0} -} &
  {\color[HTML]{C0C0C0} -} &
  {\color[HTML]{C0C0C0} -} &
   &
  {\color[HTML]{C0C0C0} -} &
  {\color[HTML]{C0C0C0} -} &
  {\color[HTML]{C0C0C0} -} \\
Base CNN &
  $M_\nu(\bm{\nu})=\hat{\bm{y}}_0$ &
   &
  87.3\% &
  {\color[HTML]{C0C0C0} -} &
  {\color[HTML]{C0C0C0} -} &
  {\color[HTML]{C0C0C0} -} &
   &
  {\color[HTML]{C0C0C0} -} &
  {\color[HTML]{C0C0C0} -} &
  {\color[HTML]{C0C0C0} -} \\ \cline{1-2} \cline{4-7} \cline{9-11}
Multi-task &
  $M_\tau(\bm{x},\bm{a})=(\hat{\bm{y}}_0,\hat{\bm{\tau}})$ &
   &
  83.2\% &
  {\color[HTML]{C0C0C0} -} &
  {\color[HTML]{C0C0C0} -} &
  {\color[HTML]{C0C0C0} -} &
   &
  0.000 &
  0.030 &
  {\color[HTML]{C0C0C0} -} \\
Multi-task &
  $M_\theta(\bm{x},\bm{a})=(\hat{\bm{y}}_0,\hat{\bm{\theta}})$ &
   &
  74.3\% &
  {\color[HTML]{C0C0C0} -} &
  {\color[HTML]{C0C0C0} -} &
  {\color[HTML]{C0C0C0} -} &
   &
  0.000 &
  0.067 &
  0.0069 \\ \cline{1-2} \cline{4-7} \cline{9-11}
Encoder-decoder &
  $M_m(\bm{x},\bm{a})=(\hat{\bm{y}}_0,\hat{\bm{m}})$ &
   &
  91.7\% &
  {\color[HTML]{C0C0C0} -} &
  {\color[HTML]{C0C0C0} -} &
  {\color[HTML]{C0C0C0} -} &
   &
  0.654 &
  {\color[HTML]{C0C0C0} -} &
  {\color[HTML]{C0C0C0} -} \\
Encoder-decoder &
  $M_{a_m}(\bm{x},\bm{a})=(\hat{\bm{y}}_0,\hat{\bm{a}}_m,\hat{\bm{\theta}})$ &
   &
  94.8\% &
  {\color[HTML]{C0C0C0} -} &
  {\color[HTML]{C0C0C0} -} &
  {\color[HTML]{C0C0C0} -} &
   &
  0.769 &
  0.020 &
  0.0062 \\ \cline{1-2} \cline{4-7} \cline{9-11}
DiagramNet &
  $M(\bm{x},\bm{a})=(\hat{\bm{y}}_0, \hat{\bm{\phi}},\Breve{\bm{s}}_m, \Breve{\bm{y}}_e,\hat{\bm{y}}_r,\hat{\bm{y}})$ &
   &
  95.1\% &
  76.2\% &
  92.6\% &
  \textbf{95.7\%} &
   &
  \textbf{0.835} &
  \textbf{0.006} &
  \textbf{0.0008} \\ \hline
\end{tabular}
\end{table}


\subsection{User study}

\subsubsection{Other explanation cases}
\phantom{...}

\begin{figure}[h!]
    \centering
    \includegraphics[width=13.5cm]{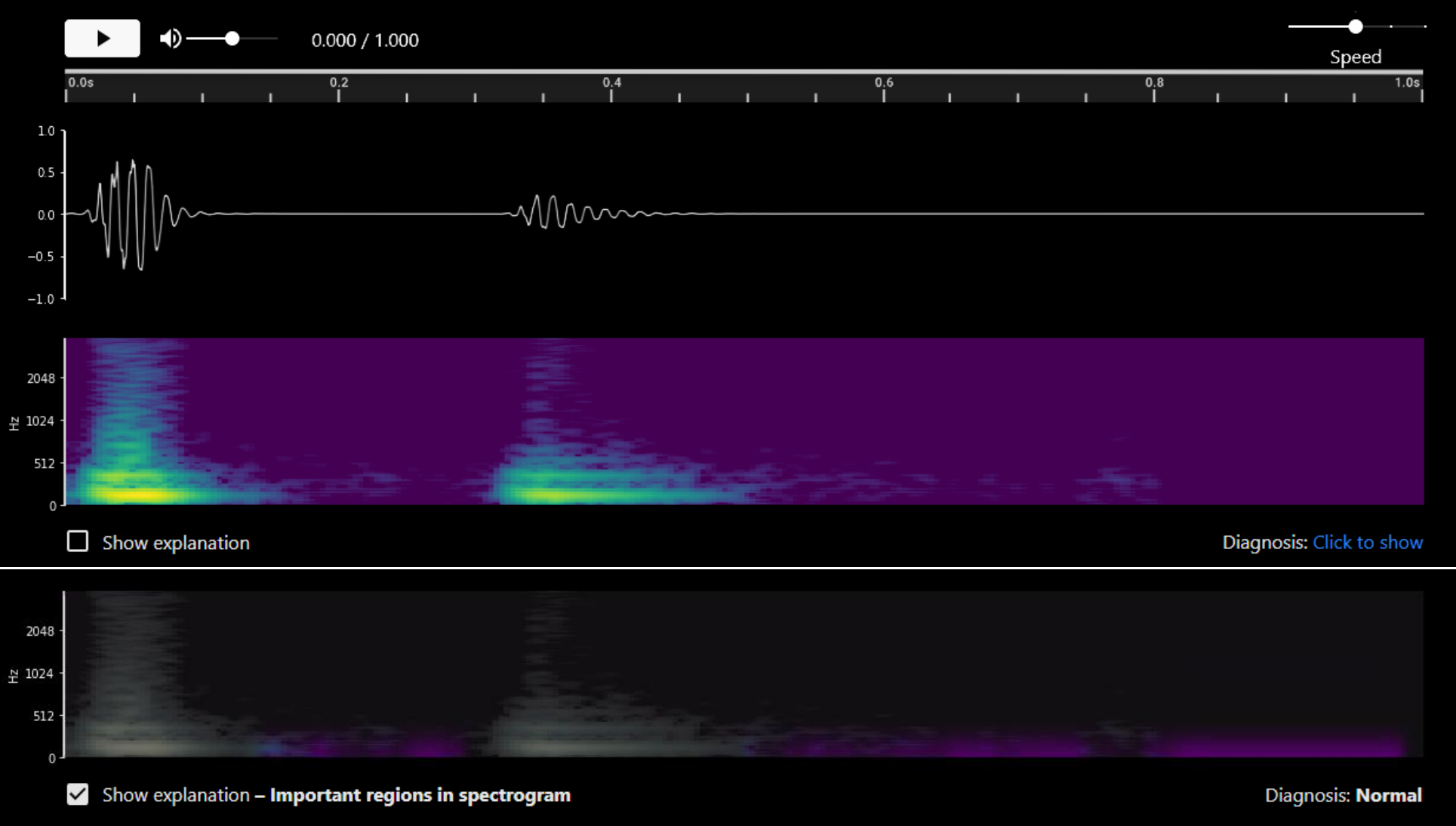}
    \caption{
    User interface with Spectrogram-saliency XAI for the user study.
    Case of Normal diagnosis. Notice how the saliency map identifies low amplitudes of the low frequencies outside of S1 and S2, suggesting the lack of murmurs as the explanation for normal sound.
    }
    \Description{
    Figure shows two screenshots of the user interface with identical layout to Figure 13. Main difference is that the phonocardiogram only shows audio at S1 lub and S2 dub times with zero amplitude elsewhere, spectrogram only showing amplitudes at the same time for low frequencies, and saliency map only highlighting the times outside of S1 and S2.
    }
    \label{fig:userstudy-spectrogram-N}
\end{figure}

\begin{figure}[h!]
    \centering
    \includegraphics[width=13.5cm]{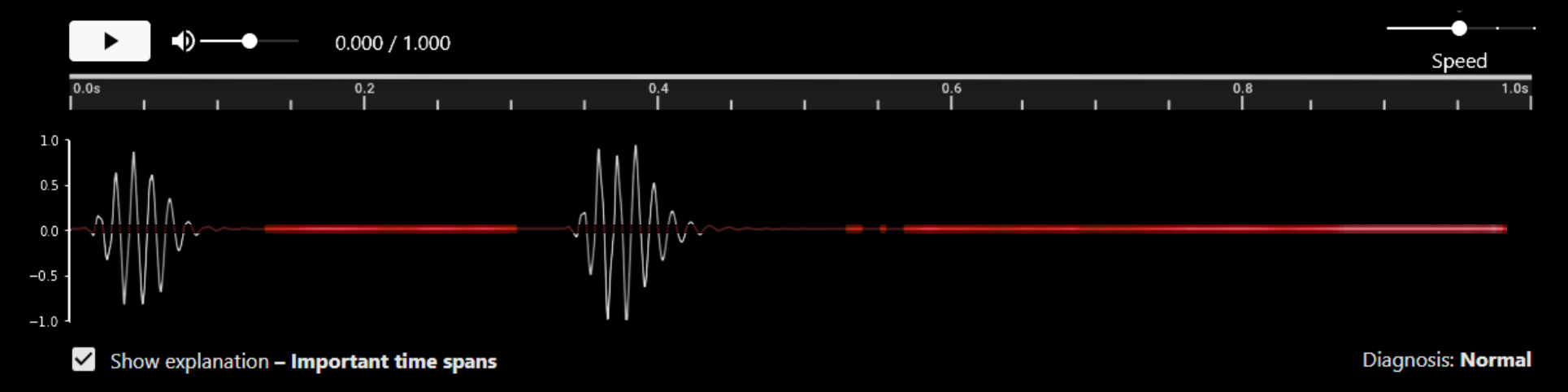}
    \caption{
    User interface with Time-saliency XAI for the user study.
    Case of Normal diagnosis. Notice how the saliency map identifies the time regions outside of S1 and S2, suggesting the lack of murmurs as the explanation for normal sound.
    }
    \Description{
    Figure shows a screenshot of the user interface with identical layout to Figure 14. Main difference is that the phonocardiogram only shows audio at S1 lub and S2 dub times with zero amplitude elsewhere, and Time-saliency map only highlighting the times outside of S1 and S2.
    }
    \label{fig:userstudy-time-N}
\end{figure}


\end{document}